\makeatletter
\newcommand{\dontusepackage}[2][]{%
  \@namedef{ver@#2.sty}{9999/12/31}%
  \@namedef{opt@#2.sty}{#1}}
\makeatother
\dontusepackage{subfigure}

\documentclass[]{article}

\usepackage{lmodern}
\usepackage{amssymb,amsmath}
\usepackage{ifxetex,ifluatex}
\usepackage[usenames,dvipsnames]{color}
\usepackage{fixltx2e} 
\ifnum 0\ifxetex 1\fi\ifluatex 1\fi=0 
  \usepackage[T1]{fontenc}
  \usepackage[utf8]{inputenc}
\else 
  \ifxetex
    \usepackage{mathspec}
    \usepackage{xltxtra,xunicode}
  \else
    \usepackage{fontspec}
  \fi
  \defaultfontfeatures{Mapping=tex-text,Scale=MatchLowercase}
  
\fi
\IfFileExists{upquote.sty}{\usepackage{upquote}}{}
\IfFileExists{microtype.sty}{%
\usepackage{microtype}
\UseMicrotypeSet[protrusion]{basicmath} 
}{}
\usepackage[margin=1.0in,bottom=1.5in]{geometry}
\usepackage[square,numbers,sort&compress]{natbib}
\usepackage{listings}
\usepackage{graphicx}
\makeatletter
\def\maxwidth{\ifdim\Gin@nat@width>\linewidth\linewidth\else\Gin@nat@width\fi}
\def\maxheight{\ifdim\Gin@nat@height>\textheight\textheight\else\Gin@nat@height\fi}
\makeatother
\setkeys{Gin}{width=\maxwidth,height=\maxheight,keepaspectratio}
\usepackage{caption}
\usepackage{float}
\usepackage{subfig}
\ifxetex
  \usepackage[setpagesize=false, 
              unicode=false, 
              xetex]{hyperref}
\else
  \usepackage[unicode=true]{hyperref}
\fi
\hypersetup{breaklinks=true,
            bookmarks=true,
            pdfauthor={},
            pdftitle={Reliable amortized variational inference with physics-based latent distribution correction},
            colorlinks=true,
            citecolor=black,
            urlcolor=blue,
            linkcolor=black,
            pdfborder={0 0 0}}
\usepackage[preprint]{neurips_2019}
\usepackage{url}
\usepackage{booktabs}
\usepackage{amsfonts}
\usepackage{nicefrac}
\usepackage{microtype}

\newcommand{\B}{\mathbf}
\newcommand{\Bs}{\boldsymbol}
\newcommand{\KL}{\mathbb{KL}}

\DeclareMathOperator*{\argmin}{arg\,min}

\title{Reliable amortized variational inference with physics-based latent
distribution correction}
\author{Ali Siahkoohi\\School of Computational Science and Engineering,\\Georgia
Institute of Technology\\\texttt{alisk@gatech.edu}\\\And
Gabrio Rizzuti\\Department of Mathematics\\Utrecht
University\\\texttt{g.rizzuti@uu.nl}\\\And
Rafael Orozco\\School of Computational Science and Engineering,\\Georgia
Institute of Technology\\\texttt{rorozco@gatech.edu} \And
Felix J. Herrmann\\School of Computational Science and
Engineering,\\Georgia Institute of
Technology\\\texttt{felix.herrmann@gatech.edu}}
\date{}

\begin{document}
\maketitle
\begin{abstract}
Bayesian inference for high-dimensional inverse problems is
computationally costly and requires selecting a suitable prior
distribution. Amortized variational inference addresses these challenges
by pretraining a neural network that approximates the posterior
distribution not only for one instance of observed data, but a
distribution of data pertaining to a specific inverse problem. When fed
previously unseen data, the neural network---in our case a conditional
normalizing flow---provides posterior samples at virtually no cost.
However, the accuracy of amortized variational inference relies on the
availability of high-fidelity training data, which seldom exists in
geophysical inverse problems because of the Earth's heterogeneous
subsurface. In addition, the network is prone to errors if evaluated
over data that is not drawn from the training data distribution. As
such, we propose to increase the resilience of amortized variational
inference in the presence of moderate data distribution shifts. We
achieve this via a correction to the conditional normalizing flow's
latent distribution that improves the approximation to the posterior
distribution for the data at hand. The correction involves relaxing the
standard Gaussian assumption on the latent distribution and
parameterizing it via a Gaussian distribution with an unknown mean and
(diagonal) covariance. These unknowns are then estimated by minimizing
the Kullback-Leibler divergence between the corrected and the
(physics-based) true posterior distributions. While generic and
applicable to other inverse problems, by means of a linearized seismic
imaging example, we show that our correction step improves the
robustness of amortized variational inference with respect to changes in
the number of seismic sources, noise variance, and shifts in the prior
distribution. This approach, given noisy seismic data simulated via
linearized Born modeling, provides a seismic image with limited
artifacts and an assessment of its uncertainty at approximately the same
cost as five reverse-time migrations.
\end{abstract}

\section{Introduction}\label{introduction}

Inverse problems involve the estimation of an unknown quantity based on
noisy indirect observations. The problem is typically solved by
minimizing the difference between observed and predicted data, where
predicted data can be computed by modeling the underlying data
generation process through a forward operator. Due to the presence of
noise in the data, forward modeling errors, and the inherent nullspace
of the forward operator, minimization of the data misfit alone
negatively impacts the quality of the obtained solution
\citep{aster2018parameter}. Casting inverse problems into a
probabilistic Bayesian framework allows for a more comprehensive
description of their solution, where instead of finding one single
solution, a distribution of solutions to the inverse problem---known as
the posterior distribution---is obtained whose samples are consistent
with the observed data \citep{tarantola2005inverse}. The posterior
distribution can be sampled to extract statistical information that
allows for quantification of uncertainty, i.e., assessing the
variability among the possible solutions to the inverse problem.

Uncertainty qualification and Bayesian inference in inverse problems
often require high-dimensional posterior distribution sampling, for
instance through the use of Markov chain Monte Carlo
\citep[MCMC,][]{robert2004monte, MartinMcMC2012, fang2018uqfip, siahkoohiGEOdbif}.
Because of their sequential nature, MCMC sampling methods require a
large number of sampling steps to perform accurate Bayesian inference
\citep{gelman2013bayesian}, which reduces their applicability to
large-scale problems due to the high-dimensionality of the unknown and
costs associated with the forward operator
\citep{curtis2001prior, welling2011bayesian, MartinMcMC2012, fang2018uqfip, herrmann2019NIPSliwcuc, zhao2019gradient, kotsi2020, siahkoohi2020EAGEdlb, siahkoohi2020SEGuqi}.
As an alternative, variational inference methods
\citep{jordan1999introduction, wainwright2008graphical, rezende2015variational, liu2016stein, rizzuti2020SEGpub, zhang2020seismic, tolle2021mean, li2021multiparameter, li2021traversing}
approximate the posterior distribution with a surrogate and
easy-to-sample distribution. By means of this approximation, sampling is
turned into an optimization problem, in which the parameters of the
surrogate distribution are tuned in order to minimize the divergence
between the surrogate and posterior distributions. This surrogate
distribution is then used for conducting Bayesian inference. While
variational inference methods may have computational advantages over
MCMC methods in high-dimensional inverse problems
\citep{blei2017variational, zhang2021introduction}, the resulting
approximation to the posterior distribution is typically non-amortized,
i.e., it is specific to the observed data used in solving the
variational inference optimization problem. Thus, the variational
inference optimization problem must be solved again for every new set of
observations. Solving this optimization problem may require numerous
iterations \citep{rizzuti2020SEGpub, zhang2020seismic}, which may not be
feasible in inverse problems with computationally costly forward
operators, such as seismic imaging.

On the other hand, amortized variational inference
\citep{pmlrv80kim18e, baptista2020adaptive, kruse2021hint, kovachki2021conditional, radev2020bayesflow, siahkoohi2020ABIpto, ren2021uq, siahkoohi2021Seglbe, khorashadizadeh2022conditional, orozco2021photoacoustic, siahkoohi2022EAGEweb, taghvaei2022optimal}
reduces Bayesian inference computational costs by incurring an up-front
optimization cost for finding a surrogate conditional distribution,
typically parameterized by deep neural networks \citep{kruse2021hint},
that approximate the posterior distribution across a family of observed
data instead of being specific to a single observed dataset. This
supervised learning problem involves maximization of the probability
density function (PDF) of the surrogate conditional distribution over
existing pairs model and data \citep{radev2020bayesflow}. Following
optimization, samples from the posterior distribution for previously
unseen data may be obtained by sampling the surrogate conditional
distribution, which does not require further optimization or MCMC
sampling. While drastically reducing the cost of Bayesian inference,
amortized variational inference can only be used for inverse problems
where a dataset of model and data pairs is available that sufficiently
captures the underlying joint distribution. In reality, such an
assumption is rarely true in geophysical applications due to the Earth's
strong heterogeneity across geological scenarios and our lack of access
to its interior
\citep{siahkoohi2020ABIpto, sun2021physics, jin2022unsupervised}.
Additionally, the accuracy of Bayesian inference with data-driven
amortized variational inference methods degrades as the distribution of
the data shifts with respect to pretraining data
\citep{schmitt2021bayesflow}. Among these shifts are changes in the
distribution of noise, the number of observed data in multi-source
inverse problems, and the distribution of unknowns, in other words, the
prior distribution.

\begin{figure*}
\centering
\includegraphics[width=1.000\hsize]{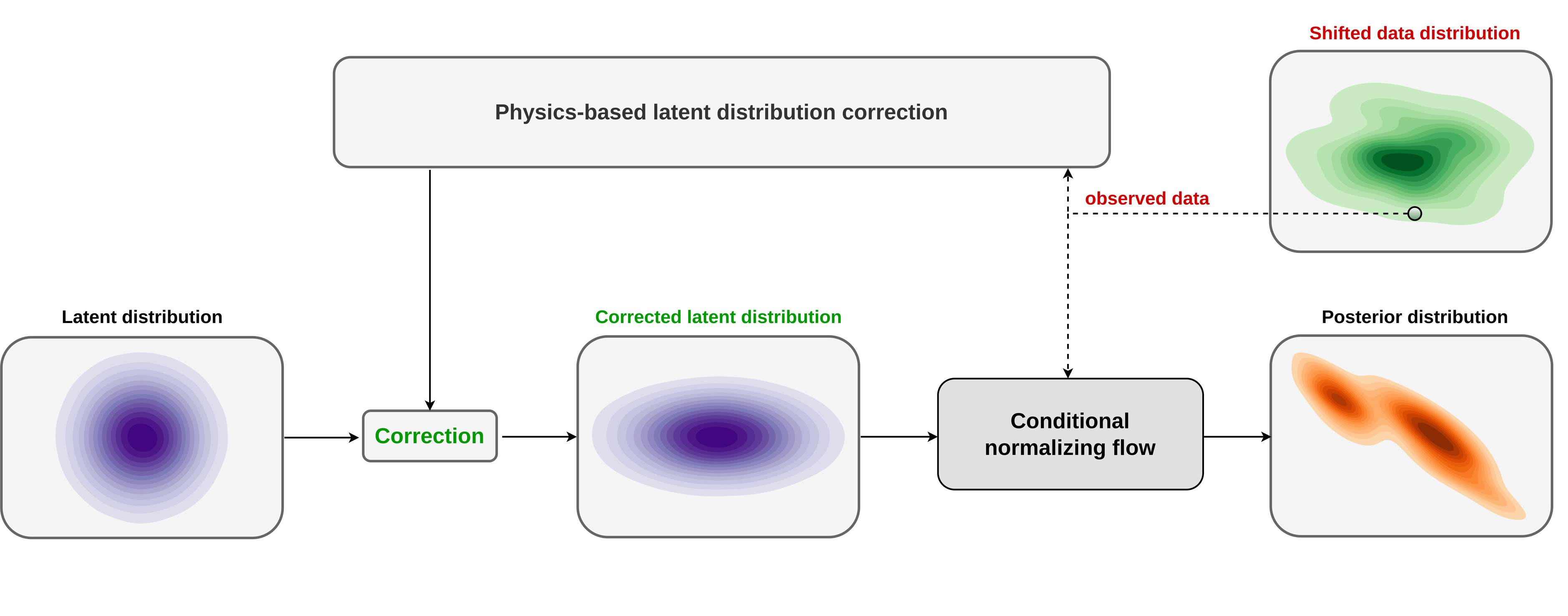}
\caption{Schematic representation of our proposed method. We modify the
standard Gaussian latent distribution of a pretrained conditional
normalizing flow through a computationally cheap diagonal physics-based
correction procedure to mitigate the errors due to data distribution
shifts. Upon correction, the new latent samples result in corrected
posterior samples when fed into the pretrained conditional normalizing
flow.}\label{schematic_correction_simple}
\end{figure*}

In this work, we leverage amortized variational inference to accelerate
Bayesian inference while building resilience against data distribution
shifts through an unsupervised, data-specific, a physics-based latent
distribution correction method. During this process, the latent
distribution of a normalizing-flow-based surrogate conditional
distribution \citep{kruse2021hint} is corrected to minimize the
Kullback-Leibler (KL) divergence between the predicted and true
posterior distributions. The invertibility of the conditional
normalizing flow---a family of invertible neural networks
\citep{dinh2016density}---guarantees the existence of a corrected latent
distribution \citep{pmlr-v119-asim20a} that when ``pushed forward'' by
the conditional normalization flow matches the posterior distribution.
During pretraining, the conditional normalizing flow learns to
Gaussianize the input model and data joint samples
\citep{kruse2021hint}, resulting in a standard Gaussian latent
distribution. As a result, for slightly shifted data distributions, the
conditional normalization flow can provide samples from the posterior
distribution given an ``approximately Gaussian'' latent distribution as
input \citep{marzuk2018, marzouk2018multifidelity}. Motivated by this,
and to limit the costs of the latent distribution correction step, we
learn a simple diagonal (elementwise) scaling and shift to the latent
distribution through a physic-based objective that minimizes the KL
divergence between the predicted and true posterior distributions. As
with amortized variational inference, after latent distribution
correction, we gain cheap access to corrected posterior samples. Besides
offering computational advantages, our proposed method implicitly learns
the prior distribution during conditional normalizing flow pretraining.
As advocated in the literature
\citep{pmlr-v70-bora17a, pmlr-v119-asim20a} learned priors have the
potential to better describe the prior information when compared to
generic handcrafted priors that are chosen purely for their simplicity
and applicability. A schematic representation of our proposed method is
shown in Figure~\ref{schematic_correction_simple}.

\subsection{Related work}\label{related-work}

In the context of variational inference for inverse problems,
\citet{rizzuti2020SEGpub}, \citet{andrle2021invertible},
\citet{zhao2021bayesian}, \citet{zhang2021bayesian2}, and
\citet{zhao2022interrogating} proposed a non-amortized variational
inference approach to approximate posterior distributions through the
use of normalizing flows. These methods do not require training data,
however they require choosing a prior distribution and repeated
computationally expensive evaluation of the forward operator and the
adjoint of its Jacobian. Therefore, the proposed methods may prove
computationally expensive when applied to inverse problems involving
computationally expensive forward operators. To speed up the convergence
of non-amortized variational inference, \citet{siahkoohi2020ABIpto}
introduces a normalizing-flow-based nonlinear preconditioning scheme. In
this approach, a pretrained conditional normalizing flow capable of
providing a low-fidelity approximation to the posterior distribution is
used to warm-start the variational inference optimization procedure. In
a related work, \citet{kothari2021trumpets} partially address challenges
associated with non-amortized variational inference by learning a
normalizing-flow-based prior distribution in a learned low-dimensional
space via an injective network. Additionally to learning a prior, this
approach also allowed non-amortized variational inference in a lower
dimensional space, which could potentially have computational benefits.

Alternatively, amortized variational inference was applied by
\citet{adler2018deep}, \citet{kruse2021hint},
\citet{kovachki2021conditional}, \citet{siahkoohi2021Seglbe}, and
\citet{khorashadizadeh2022conditional} to further reduce the
computational costs associated with Bayesian inference. These supervised
methods learn an implicit prior distribution from training data and
provide posterior samples for previously unseen data for a negligible
cost due to the low cost of forward evaluation of neural networks. The
success of such techniques hinges on having access to high-quality
training data, including pairs of model and data that sufficiently
capture the underlying model and data joint distribution. To address
this limitation, \citet{siahkoohi2020ABIpto} take amortized variational
inference a step further by proposing a two-stage multifidelity approach
where during the first stage a conditional normalizing flow is trained
in the context of amortized variational inference. To account for any
potential shift in data distribution, the weights of this pretrained
conditional normalizing flow are then further finetuned during an
optional second stage of physics-based variational inference, which is
customized for the specific imaging problem at hand. While limiting the
risk of errors caused by shifts in the distribution of data, the second
physics-based stage can be computationally expensive due to the high
dimensionality of the weight space of conditional normalizing flows. Our
work differs from the proposed method in \citet{siahkoohi2020ABIpto} in
that we learn to correct the latent distribution of the conditional
normalizing flow, which typically has a much smaller dimensionality
(approximately $\times 90$ in our case) than the dimension of the
conditional normalizing flow weight space.

The work we present is principally motivated by
\citet{pmlr-v119-asim20a}, which demonstrates that normalizing
flows---due to their invertibility---can mitigate biases caused by
shifts in the data distribution. This is achieved by reparameterizing
the unknown by a pretrained normalizing flow with fixed weights while
optimizing over the latent variable in order to fit the data. The
reparameterization together with a Gaussian prior on the latent variable
act as a regularization while the invertibility ensures the existence of
a latent variable that fits the data. \citet{pmlr-v119-asim20a} exploit
this property using a normalizing flow that is pretrained to capture the
prior distribution associated with an inverse problem. By computing the
maximum-a-posterior estimate in the latent space,
\citet{pmlr-v119-asim20a}, as well as \citet{li2021traversing} and
\citet{orozco2021photoacoustic}, limit biases originating from data
distribution shifts while utilizing the prior knowledge of the
normalizing flow. We extend this method by obtaining an approximation to
the full posterior distribution of an inverse problem instead of a point
estimate, e.g., maximum-a-posteriori.

Our work is also closely related to the non-amortized variational
inference techniques presented by \citet{whang2021composing} and
\citet{kothari2021trumpets}, in which the latent distribution of a
normalizing flow is altered in an unsupervised way in order to perform
Bayesian inference. In contrast to our approach, these methods employ a
pretrained normalizing flow that approximates the prior distribution. As
a result, it is necessary to significantly alter the latent distribution
in order to correct the pretrained normalizing flow to sample from the
posterior distribution. In response, \citet{whang2021composing} and
\citet{kothari2021trumpets} train a second normalizing flow aimed at
learning a latent distribution that approximates the posterior
distribution after passing through the pretrained normalizing flow. Our
study, however, utilizes a conditional normalizing flow, which, before
any corrections are applied, already approximates the posterior
distribution. We argue that our approach requires a simpler correction
in the latent space to mitigate biases caused by shifts in the data
distribution. This is crucial when dealing with large-scale inverse
problems with computationally expensive forward operators.

\subsection{Main contributions}\label{main-contributions}

The main contribution of our work involves a variational inference
formulation for solving probabilistic Bayesian inverse problems that
leverages the benefits of data-driven learned posteriors whilst being
informed by physics and data. The advantages of this formulation include

\begin{itemize}
\item
  Enhancing the solution quality of inverse problems by implicitly
  learning the prior distribution from the data;
\item
  Reliably reducing the cost of uncertainty quantification and Bayesian
  inference; and
\item
  Providing safeguards against data distribution shifts.
\end{itemize}

\subsection{Outline}\label{outline}

In the sections below, we first formulate multi-source inverse problems
mathematically and cast them within a Bayesian framework. We then
describe variational inference and examine how existing model and data
pairs can be used to obtain an approximation to the posterior
distribution that is amortized, i.e., the approximation holds over a
distribution of data rather than a specific set of observations. We
showcase amortized variational inference on a high-dimensional seismic
imaging example in a controlled setting where we assume observed data
during inference is drawn from the same distribution as training seismic
data. As means to mitigate potential errors due to data distribution
shifts, we introduce our proposed correction approach to amortized
variational inference, which exploits the advantages of learned
posteriors while reducing potential errors induced by certain data
distribution shifts. Two linearized seismic imaging examples are
presented, in which the distribution of the data (simulated via
linearized Born modeling) is shifted by altering the forward model and
the prior distribution. These numerical experiments are intended to
demonstrate the ability of the proposed latent distribution correction
method to correct for errors caused by shifts in the distribution of
data. Finally, we verify our proposed Bayesian inference method by
conducting posterior contraction experiments.

\section{Theory}\label{theory}

Our purpose is to present a technique for using deep neural networks to
accelerate Bayesian inference for ill-posed inverse problems while
ensuring that the inference is robust with respect to data distribution
shifts through the use of physics. We begin with an introduction to
Bayesian inverse problems and discuss variational inference
\citep{jordan1999introduction} as a probabilistic framework for solving
Bayesian inverse problems.

\subsection{Inverse problems}\label{inverse-problems}

We are concerned with estimating an unknown multidimensional quantity
$\B{x}^{\ast} \in \mathcal{X}$, often referred to as the unknown model,
given $N$ noisy and indirect observed data (e.g, shot records in seismic
imaging) $\B{y} = \{\B{y}_i\}_{i=1}^N$ with $\B{y}_i \in \mathcal{Y}$.
Here $\mathcal{X}$ and $\mathcal{Y}$ denote the space of unknown models
and data, respectively. The physical underlying data generation process
is assumed to be encoded in forward modeling operators,
$\mathcal{F}_i:\mathcal{X} \rightarrow \mathcal{Y}$, which relates the
unknown model to the observed data via the forward model
\begin{equation}
\B{y}_i = \mathcal{F}_i(\B{x}^{\ast}) + \boldsymbol{\epsilon}_i,
\quad i=1,\dots,N.
\label{forward_model}
\end{equation}
 In the above expression, $\boldsymbol{\epsilon}_i$ is a vector of
measurement noise, which might also include errors in the forward
modeling operator. Solving ill-posed inverse problems is challenged by
noise in the observed data, potential errors in the forward modeling
operator, and the intrinsic nontrivial nullspace of the forward operator
\citep{aster2018parameter}. These challenges can lead to non-unique
solutions where different estimates of the unknown model may fit the
observed data equally well. Under such conditions, the use of a single
model estimate ignores the intrinsic variability within inverse problem
solutions, which increases the risk of overfitting the data. Therefore,
not only the process of estimating $\B{x}^{\ast}$ from $\B{y}$ requires
regularization, but it also calls for a statistical inference framework
that allows us to characterize the variability among the solutions by
quantifying the solution uncertainty \citep{tarantola2005inverse}.

\subsection{Bayesian inference for solving inverse
problems}\label{bayesian-inference-for-solving-inverse-problems}

To systematically quantify the uncertainty, we cast the inverse problem
into a Bayesian framework \citep{tarantola2005inverse}. In this
framework, instead of having a single estimate of the unknown, the
solution is characterized by a probability distribution over the
solution space $\mathcal{X}$ that is conditioned on data, namely the
posterior distribution. This conditional distribution, denoted by
$p_{\text{post}} (\B{x} \mid \B{y})$, can according to the Bayes' rule
be written as follows:
\begin{equation}
p_{\text{post}} (\B{x} \mid \B{y}) =
\frac{p_{\text{like}} (\B{y} \mid \B{x})\,
p_{\text{prior}} (\B{x})}{p_{\text{data}} (\B{y})}.
\label{bayes_rule}
\end{equation}
 which equivalently can be expressed as
\begin{equation}
\begin{aligned}
- \log p_{\text{post}} (\B{x} \mid \B{y}) & =  - \sum_{i=1}^{N}
\log p_{\text{like}} (\B{y}_i \mid \B{x} )
- \log p_{\text{prior}} (\B{x}) + \log p_{\text{data}} (\B{y}) \\
& = \frac{1}{2 \sigma^2} \sum_{i=1}^{N}
\big  \| \B{y}_i-\mathcal{F}_i(\B{x}) \big\|_2^2
- \log p_{\text{prior}} (\B{x}) + \text{const},
\end{aligned}
\label{bayes_rule_log}
\end{equation}
 in case the observed data ($\B{y}_i$) are independent conditioned on
the unknown model $\B{x}$. In equations~\ref{bayes_rule}
and~\ref{bayes_rule_log}, the likelihood function
$p_{\text{like}} (\B{y} \mid \B{x})$ quantifies how well the predicted
data fits the observed data given the PDF of the noise distribution. For
simplicity, we assume the distribution of the noise is a zero-mean
Gaussian distribution with covariance $\sigma^2 \B{I}$ but other choices
can be incorporated. The prior distribution $p_{\text{prior}} (\B{x})$
encodes prior beliefs on the unknown quantity, which can also be
interpreted as a regularizer for the inverse problem. Finally,
$p_{\text{data}} (\B{y})$ denotes the data PDF, which is a normalization
constant that is independent of $\B{x}$.

Acquiring statistical information regarding the posterior distribution
requires access to samples from the posterior distribution. Sampling the
posterior distribution, commonly achieved via MCMC
\citep{robert2004monte} or variational inference techniques
\citep{jordan1999introduction}, is computationally costly in
high-dimensional inverse problems due to the costs associated with many
needed evaluations of the forward operator
\citep{malinverno2004expanded, tarantola2005inverse, malinverno2006two, MartinMcMC2012, blei2017variational, chevron2017, fang2018uqfip, stuart2019two, zhao2019gradient, kotsi2020, zhang2021introduction}.
For multi-source inverse problem the costs are especially high as
evaluating the likelihood function involves $N$ forward operator
evaluations (equation~\ref{bayes_rule_log}). Stochastic gradient
Langevin dynamics
\citep[SGLD;][]{welling2011bayesian, li2016preconditioned, siahkoohiGEOdbif}
alleviates the need to evaluate the likelihood for all the $N$ forward
operators by allowing for stochastic approximations to the likelihood,
i.e., evaluating the likelihood over randomly selected indices
$i \in \{1, \ldots, N \}$. While SGLD can provably provide accurate
posterior samples with more favorable computational costs
\citep{welling2011bayesian}, due to the sequential nature of MCMC
methods, SGLD still requires numerous iterations to fully traverse the
probability space \citep{gelman2013bayesian}, which is computationally
challenging in large-scale multi-source inverse problems. In the next
section, we introduce variational inference as an alternative Bayesian
inference method that has the potential to scale better than
MCMC-methods in inverse problems with costly forward operators
\citep{blei2017variational, zhang2021introduction}.

\subsection{Variational inference}\label{variational-inference}

As an alternative to MCMC-based methods, variational inference methods
\citep{jordan1999introduction} reduce the problem of sampling from the
posterior distribution $p_{\text{post}} (\B{x} \mid \B{y})$ to an
optimization problem. The optimization problem involves approximating
the posterior PDF via the PDF of a tractable surrogate distribution
$p_{\phi} (\B{x})$ with parameters $\Bs{\phi}$ by minimizing a
divergence (read ``distance'') between $p_{\phi} (\B{x})$ and
$p_{\text{post}}(\B{x} \mid \B{y})$ with respect to surrogate
distribution parameters $\Bs{\phi}$. This optimization problem can be
solved approximately, which allows for trading off computational cost
for accuracy \citep{jordan1999introduction}. After optimization, we gain
access to samples from the posterior distribution by sampling
$p_{\phi} (\B{x})$ instead, which does not involve forward operator
evaluations.

Due to its simplicity and connections to the maximum likelihood
principle \citep{bishop2006pattern}, we formulate variational inference
via the Kullback-Leibler (KL) divergence. The KL divergence can be
explained as the cross-entropy of $p_{\text{post}}(\B{x} \mid \B{y})$
relative to $p_{\phi}(\B{x})$ minus the entropy of $p_{\phi}(\B{x})$.
This definition describes the reverse KL divergence, denoted by
$\KL\,\big( p_{\phi}(\B{x}) \mid\mid p_{\text{post}}(\B{x} \mid \B{y})\big)$,
which is not equal to the forward KL divergence,
$\KL\,\big( p_{\text{post}}(\B{x} \mid \B{y}) \mid\mid p_{\phi}(\B{x}) \big)$.
This non-symmetry in KL divergence leads to different computational and
approximation properties during variational inference, which we describe
in detail in the following sections. We will first describe the reverse
KL divergence, followed by the forward KL divergence. Finally, we will
describe normalizing flows as a way of parameterized surrogate
distributions to facilitate variational inference.

\subsubsection{Non-amortized variational
inference}\label{non-amortized-variational-inference}

The reverse KL divergence is the common choice for formulating
variational inference
\citep{rizzuti2020SEGpub, andrle2021invertible, zhao2021bayesian, zhang2021bayesian2, zhao2022interrogating}
in which the physically-informed posterior density guides the
optimization over $\Bs{\phi}$. The reverse KL divergence can be
mathematically stated as
\begin{equation}
\KL\,\big( p_{\phi}(\B{x}) \mid\mid
p_{\text{post}}(\B{x} \mid \B{y}_{\text{obs}})\big)
= \mathbb{E}_{\B{x} \sim p_{\phi}(\B{x})}
\big[ - \log p_{\text{post}} (\B{x} \mid \B{y}_{\text{obs}})
+ \log  p_{\phi}(\B{x}) \big],
\label{reverse_kl_divergence}
\end{equation}
 where $\B{y}_{\text{obs}} \sim p_{\text{data}} (\B{y})$ refers to a
specific single observed data. $\B{x}$ in the right hand side of the
expression in equation~\ref{reverse_kl_divergence} is a random variable
obtained by sampling the surrogate distribution $p_{\phi}(\B{x})$, over
which we evaluate the expectation. Variational inference using the
reverse KL divergence involves minimizing
equation~\ref{reverse_kl_divergence} with respect to $\Bs{\phi}$ during
which the logarithm of the posterior PDF is approximated by the
logarithm of the surrogate PDF, when evaluated over samples from the
surrogate distribution. By expanding the negative-log posterior density
via Bayes' rule (equation~\ref{bayes_rule_log}), we write the
non-amortized variational inference optimization problem as
\begin{equation}
\Bs{\phi}^{\ast} = \argmin_{\phi} \mathbb{E}_{\B{x} \sim
p_{\phi}(\B{x})} \left [ \frac{1}{2 \sigma^2}
\sum_{i=1}^{N}\big  \| \B{y}_{\text{obs}, i}-\mathcal{F}_i (\B{x}) \big\|_2^2
- \log p_{\text{prior}} \left(\B{x} \right) + \log  p_{\phi}(\B{x})
\right ].
\label{vi_reverse_kl}
\end{equation}
 The expectation in the above equation is approximated with a sample
mean over samples drawn from $p_{\phi}(\B{x})$. The optimization problem
in equation~\ref{vi_reverse_kl} can be solved using stochastic gradient
descent and its variants
\citep{robbins1951stochastic, nemirovski2009robust, rmsprop, kingma2014adam}
where at each iteration the objective function is evaluated over a batch
of samples drawn from $p_{\phi}(\B{x})$ and randomly selected (without
replacement) indices $i \in \{1, \ldots, N \}$. To solve this
optimization problem, there are two considerations to take into account.
First consideration involves the tractable computation of the surrogate
PDF and its gradient with respect to $\Bs{\phi}$. As described in the
following sections, normalizing flows \citep{rezende2015variational},
which are a family of specially designed invertible neural networks
\citep{dinh2016density}, facilitate the computation of these quantities
via the change-of-variable formula in probability distributions
\citep{villani2009optimal}. The second consideration involves
differentiating (with respect to $\Bs{\phi}$) the expectation (sample
mean) operation in equation~\ref{vi_reverse_kl}. Evaluating this
expectation requires sampling from the surrogate distribution
$p_{\phi}(\B{x})$, which depends $\Bs{\phi}$. Differentiating through
the sampling procedure from the surrogate distribution $p_{\phi}(\B{x})$
can be facilitated through the reparameterization trick
\citep{Kingma2014}. In this approach sampling from $p_{\phi}(\B{x})$ is
interpreted as passing latent samples $\B{z} \in \mathcal{Z}$ from a
simple base distribution, such as standard Gaussian distribution,
through a parametric function parameterized by $\Bs{\phi}$
\citep{Kingma2014}. With this interpretation, the expectation over
$p_{\phi}(\B{x})$ can be computed over the latent distribution instead,
which does not depend on $\Bs{\phi}$, followed by a mapping of latent
samples through the parametric function. This process enables computing
the gradient of the expression in equation~\ref{vi_reverse_kl} with
respect to $\Bs{\phi}$ \citep{Kingma2014}.

Following optimization, $p_{\phi^{\ast}}(\B{x})$ provides unlimited
samples from the posterior distribution---virtually for free. While
there are indications that this approach can be computationally
favorable compared to MCMC sampling methods
\citep{blei2017variational, zhang2021introduction}, each iteration
during optimization problem~\ref{vi_reverse_kl} involves evaluating the
forward operator and the adjoint of its Jacobian, which can be
computationally costly depending on $N$ and the number of iterations
required to solve~\ref{vi_reverse_kl}. In addition, and more
importantly, this approach is non-amortized---i.e., the resulting
surrogate distribution $p_{\phi^{\ast}}(\B{x})$ approximates the
posterior distribution for the specific data $\B{y}_{\text{obs}}$ that
is used to solve optimization problem~\ref{vi_reverse_kl}. This
necessitates the optimization problem to be solved again for a new
instance of the inverse problem with different data. In the next
section, we introduce an amortized variational inference approach that
addresses these limitations.

\subsubsection{Amortized variational
inference}\label{amortized-variational-inference}

Similarly to reverse KL divergence, forward KL divergence involves
calculating the difference between the logarithms of the surrogate PDF
and the posterior PDF. In contrast to reverse KL divergence, however, to
compute the forward KL divergence the PDFs are evaluated over samples
from the posterior distribution rather than the surrogate distribution
samples (see equation~\ref{reverse_kl_divergence}). The forward KL
divergence can be written as follows
\begin{equation}
\KL\,\big(p_{\text{post}}(\B{x} \mid \B{y}) \mid\mid
p_{\phi}(\B{x})\big) = \mathbb{E}_{\B{x} \sim  p_{\text{post}}(\B{x} \mid \B{y})}
\big[ -\log p_{\phi} (\B{x}) + \log  p_{\text{post}}(\B{x} \mid \B{y}) \big].
\label{forward_kl_divergence}
\end{equation}
 Following the expression above, it is infeasible to evaluate the
forward KL divergence in inverse problems as it requires access to
samples from the posterior distribution---the samples that we are
ultimately after and do not have access to. However, the average (over
data) forward KL divergence can be computed using available model and
data pairs in the form of samples from the joint distribution
$p(\B{x}, \B{y})$. This involves integrating (marginalizing) the forward
KL divergence over existing data $\B{y} \sim p_{\text{data}}(\B{y})$:
\begin{equation}
\begin{aligned}
& \mathbb{E}_{\B{y} \sim
p_{\text{data}}(\B{y})} \Big[ \KL\,\big(p_{\text{post}}(\B{x} \mid \B{y}) \mid\mid
p_{\phi}(\B{x})\big) \Big] \\
& =\mathbb{E}_{\B{y} \sim p_{\text{data}}(\B{y})}
\mathbb{E}_{\B{x} \sim  p_{\text{post}}(\B{x} \mid \B{y})}
\Big[  -\log p_{\phi} (\B{x} \mid \B{y})  +
\underbrace{ \log  p_{\text{post}}(\B{x} \mid \B{y})}_{
  \text{constant w.r.t. } \Bs{\phi}} \Big] \\
& = \iint \underbrace{p_{\text{data}}(\B{y}) p_{\text{post}}(\B{x} \mid \B{y})}_{= p(\B{x}, \B{y})}
\Big[  -\log p_{\phi} (\B{x} \mid \B{y})\Big] \mathrm{d}\B{x}\,\mathrm{d}\B{y} + \text{const} \\
& =\mathbb{E}_{(\B{x}, \B{y}) \sim
 p(\B{x}, \B{y})} \big[ -\log p_{\phi} (\B{x} \mid \B{y}) \big] + \text{const}.
\end{aligned}
\label{average_forward_kl}
\end{equation}
 In the above expression $p_{\phi}(\B{x} \mid \B{y})$ represents a
surrogate conditional distribution that approximates the posterior
distribution for any data $\B{y} \sim p_{\text{data}}(\B{y})$. The third
line in equation~\ref{average_forward_kl} is the result of applying the
chain rule of PDFs\footnote{$p(x, y) = p(x \mid y)\, p(y), \ \forall\, x \in \mathcal{X},\, y \in \mathcal{Y}$.}.
By minimizing the average KL divergence we obtain the following
amortized variational inference objective:

\begin{equation}
\begin{aligned}
\Bs{\phi}^{\ast} & = \argmin_{\phi}\,\mathbb{E}_{\B{y} \sim
p_{\text{data}}(\B{y})} \big[ \KL\,(p_{\text{post}}(\B{x} \mid \B{y})
\mid\mid p_{\phi}(\B{x} \mid \B{y})) \big] \\
& = \argmin_{\phi}\, \mathbb{E}_{(\B{x}, \B{y}) \sim
 p(\B{x}, \B{y})} \big[ -\log p_{\phi} (\B{x} \mid \B{y}) \big].
\end{aligned}
\label{vi_forward_kl}
\end{equation}

The above optimization problem represent a supervised learning framework
for obtaining fully-learned posteriors using existing pairs of model and
data. The expectation is approximated with a sample mean over available
model and data joint samples. Note that this method does not impose any
explicit assumption on the noise distribution (see
equation~\ref{bayes_rule_log}), and the information about the forward
model is implicitly encoded in the model and data pairs. As a result,
this formulation is an instance of likelihood-free simulation-based
inference methods \citep{cranmer2020frontier, lavin2021simulation} that
allows us to approximate the posterior distribution for previously
unseen data as,
\begin{equation}
p_{\phi^{\ast}}(\B{x} \mid \B{y}_{\text{obs}}) \approx p_{\text{post}}
(\B{x} \mid \B{y}_{\text{obs}}),\quad \forall\, \B{y}_{\text{obs}}
\sim p_{\text{data}} (\B{y}).
\label{amortization}
\end{equation}
 Equation~\ref{amortization} holds for previously unseen data drawn from
$p_{\text{data}} (\B{y})$ provided that the optimization
problem~\ref{vi_forward_kl} is solved accurately
\citep{kruse2021hint, schmitt2021bayesflow}, i.e.,
$\mathbb{E}_{\B{y} \sim p_{\text{data}}(\B{y})} \big[ \KL\,(p_{\text{post}}(\B{x} \mid \B{y}) \mid\mid p_{\phi^{\ast}}(\B{x} \mid \B{y})) \big] =0$.
Following an one-time upfront cost of training,
equation~\ref{amortization} can be used to sample the posterior
distribution with no additional forward operator evaluations. While
computationally cheap, the accuracy of the amortized variational
inference approach in equation~\ref{vi_forward_kl} is directly linked to
the quality and quantity of model and data pairs used during
optimization \citep{cranmer2020frontier}. This raises questions
regarding the reliability of this approach in domains that sufficiently
capturing the underlying joint model and data distribution is
challenging, e.g., in geophysical applications due to the Earth's strong
heterogeneity across geological scenarios and our lack of access to its
interior
\citep{siahkoohi2020ABIpto, sun2021physics, jin2022unsupervised}. To
increases the resilience of amortized variational inference when faced
with data distribution shifts, e.g., changes in the forward model or
prior distribution, we propose a latent distribution correction to
physically inform the inference. Before describing our proposed
physics-based latent distribution correction approach, we introduce
conditional normalizing flows \citep{kruse2021hint} to parameterize the
surrogate conditional distribution for amortized variational inference.

\subsection{Conditional normalizing flows for amortized variational
inference}\label{conditional-normalizing-flows-for-amortized-variational-inference}

To limit the computational cost of amortized variational inference, both
during optimization and inference, it is imperative that the surrogate
conditional distribution be able to: (1) approximate complex
distributions, i.e., it should have a high representation power, which
is required to represent possibly multi-modal distributions; (2) support
cheap density estimation, which involves computing the density
$p_{\phi}(\B{x} \mid \B{y})$ for given $\B{x}$ and $\B{y}$; and (3)
permit fast sampling from $p_{\phi}(\B{x} \mid \B{y})$ for cheap
posterior sampling during inference. These characteristics are provided
by conditional normalizing flows \citep{kruse2021hint}, which are a
family of invertible neural networks \citep{dinh2016density} that are
capable of approximating complex conditional distributions
\citep{NEURIPS2020_2290a738, ishikawa2022universal}.

\begin{figure*}
\centering
\includegraphics[width=1.000\hsize]{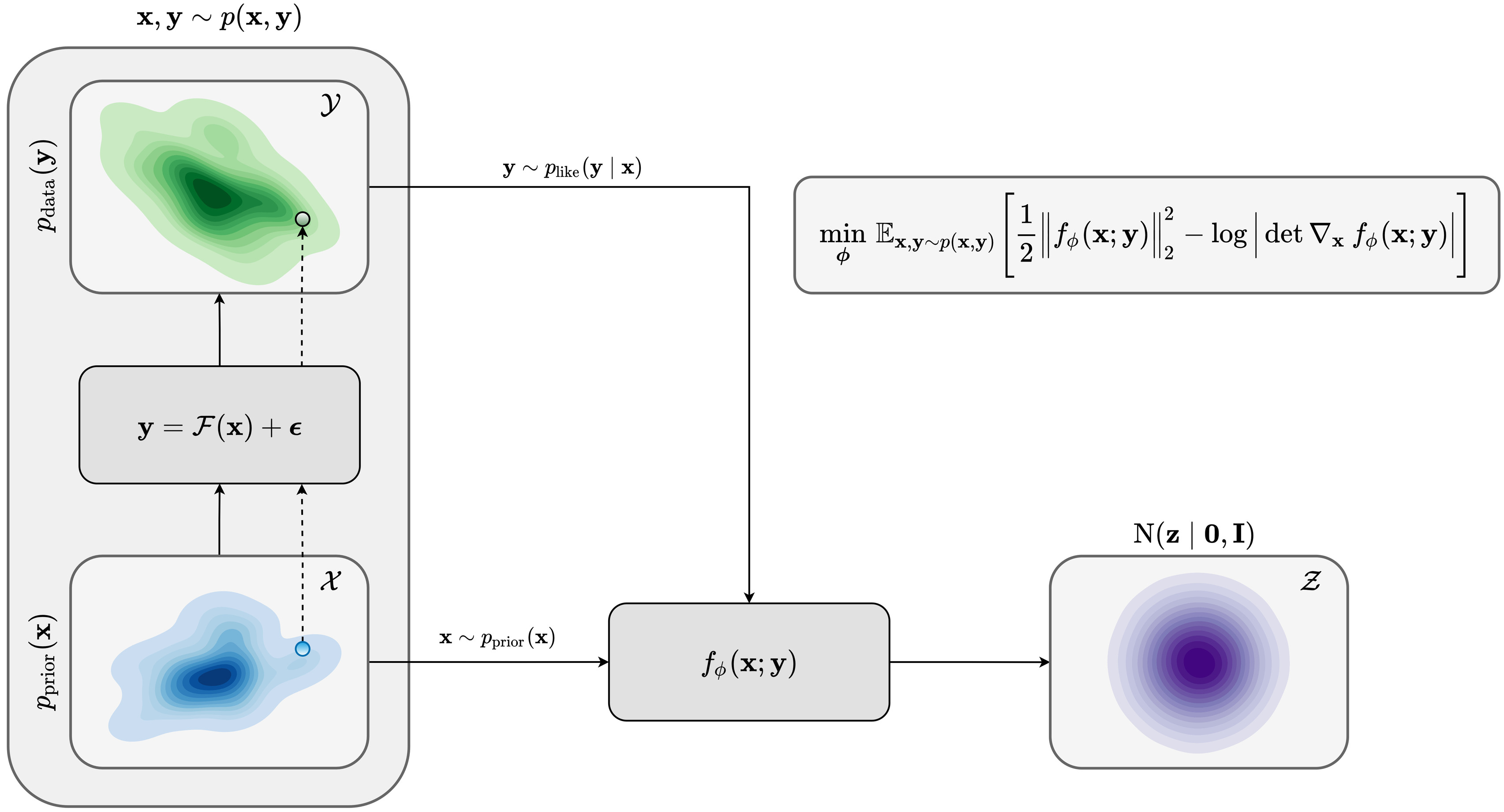}
\caption{A schematic representation of pretraining conditional
normalizing flows in the context of amortized variational inference.
During pretraining, joint model and data joint samples
$\B{x}, \B{y} \sim p(\B{x}, \B{y})$ from the training dataset and are
fed to the conditional normalizing flow. The training objective
(equation~\ref{vi_forward_kl_nf}) enforces the conditional normalizing
flow to Gaussianize its input.}\label{schematic_pretraining}
\end{figure*}

A conditional normalizing flows---in the context of amortized
variational inference---aims to map input samples $\B{z}$ from a latent
standard multivariate Gaussian distribution
$\mathrm{N}(\B{z} \mid \B{0}, \B{I})$ to samples from the posterior
distribution given the observed data $\B{y} \sim p_{\text{data}}(\B{y})$
as an additional input. This nonlinear mapping can formally be stated as
$f_{\phi}^{-1}(\,\cdot\,; \B{y}): \mathcal{Z} \to \mathcal{X}$, with
$f_{\phi}^{-1}(\B{z}; \B{y})$ being the inverse of the conditional
normalizing flow with respect to its first argument. Due to the low
computational cost of evaluating invertible neural networks in reverse
\citep{dinh2016density}, using conditional normalizing flows as a
surrogate conditional distribution $p_{\phi}(\B{x} \mid \B{y})$ allows
for extremely fast sampling from $p_{\phi}(\B{x} \mid \B{y})$. In
addition to low-cost sampling, the invertibility of conditional
normalizing flows permits straightforward and cheap estimation of the
density $p_{\phi}(\B{x} \mid \B{y})$. This allows for tractable
amortized variational inference via equation~\ref{vi_forward_kl} through
the following change-of-variable formula in probability distributions
\citep{villani2009optimal},
\begin{equation}
p_{\phi}(\B{x} \mid \B{y}) =
\mathrm{N}\left(f_{\phi}(\B{x}; \B{y}) \,\bigl\vert\, \B{0},
\B{I}\right)\, \Big | \det \nabla_{\B{x}} f_{\phi}(\B{x}; \B{y}) \Big |,
\quad \forall\, \B{x}, \B{y} \sim p(\B{x}, \B{y}).
\label{density_estimation}
\end{equation}
 In the above formula,
$\mathrm{N}\left(f_{\phi}(\B{x}; \B{y}) \,\bigl\vert\, \B{0}, \B{I}\right)$
represents the PDF for a multivariate standard Gaussian distribution
evaluated at $f_{\phi}(\B{x}; \B{y})$. Thanks to the special design of
invertible neural networks \citep{dinh2016density}, density estimation
via equation~\ref{density_estimation} is cheap since evaluating the
conditional normalizing flow and the determinant of its Jacobian
$\det \nabla_{\B{x}} f_{\phi}(\B{x}; \B{y})$ are almost free of cost.
Given the expression for $p_{\phi}(\B{x} \mid \B{y})$ in
equation~\ref{density_estimation}, we derive the following training
objective for amortized conditional normalizing flows:
\begin{equation}
\begin{aligned}
\Bs{\phi}^{\ast} & = \argmin_{\phi}\, \mathbb{E}_{(\B{x}, \B{y}) \sim
p(\B{x}, \B{y})} \big[ -\log p_{\phi} (\B{x} \mid \B{y}) \big] \\
& = \argmin_{\phi}\, \mathbb{E}_{(\B{x}, \B{y}) \sim
p(\B{x}, \B{y})} \Big[ \frac{1}{2}
\left \| f_{\phi}(\B{x}; \B{y}) \right \|^2_2
- \log \Big | \det \nabla_{\B{x}}
f_{\phi}(\B{x}; \B{y}) \Big | \Big].
\end{aligned}
\label{vi_forward_kl_nf}
\end{equation}
 In the above objective, the $\ell_2$-norm follows from a standard
Gaussian distribution assumption on the latent variable, i.e., the
output of the normalizing flow. The second term quantifies the relative
change of density volume {[}papamakarios2021{]} and can be interpreted
as an entropy regularization of $p_{\phi}(\B{x} \mid \B{y})$, which
prevents the conditional normalizing flow from converging to solutions,
e.g., $f_{\phi}(\B{x}; \B{y}) := \B{0}$. Due to the particular design of
invertible networks \citep{dinh2016density, kruse2021hint}, computing
the gradient of $\det \nabla_{\B{x}} f_{\phi}(\B{x}; \B{y})$ has a
negligible extra cost. Figure~\ref{schematic_pretraining} illustrates
the pretraining phase as a schematic.

\begin{figure*}
\centering
\includegraphics[width=0.850\hsize]{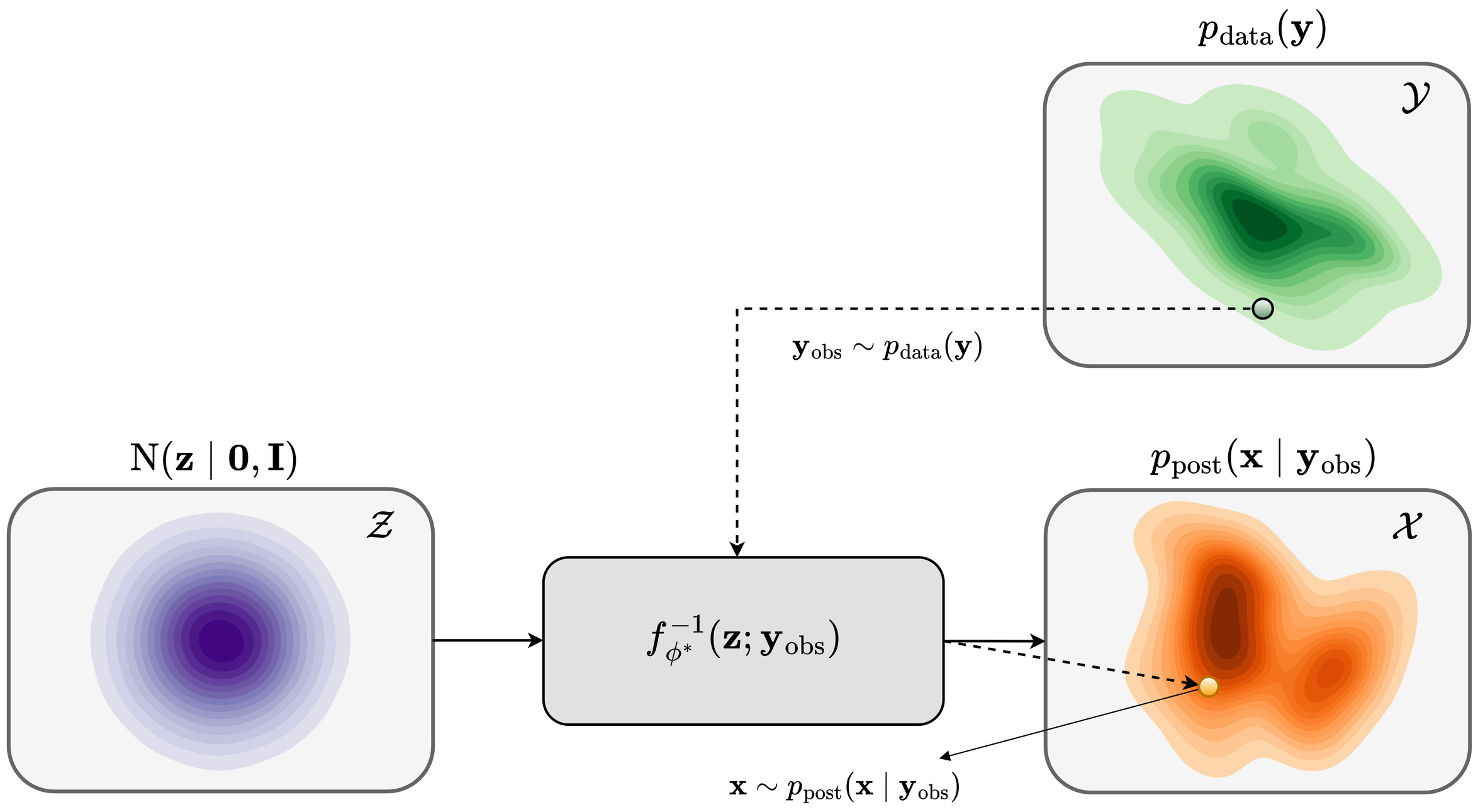}
\caption{A schematic representation of posterior sampling with
pretrained conditional normalizing flows. To sample from
$p_{\text{post}} (\B{x} \mid \B{y}_{\text{obs}})$, the observed data and
latent samples are fed to the conditional normalizing flow's inverse
$f_{\phi}^{-1}(\B{z}; \B{y}_{\text{obs}})$. Each latent sample
realization results in a realization of the posterior
distribution.}\label{schematic_sampling}
\end{figure*}

After training, given a previously unseen observed data
$\B{y}_{\text{obs}} \sim p_{\text{data}}(\B{y})$ we sample from the
posterior distribution using the inverse of the conditional normalizing
flow. We achieve this by feeding latent samples
$\B{z} \sim \mathrm{N} (\B{z} \mid \B{0}, \B{I})$ to the conditional
normalizing flow's inverse $f_{\phi}^{-1}(\B{z}; \B{y}_{\text{obs}})$
while conditioning on the observed data $\B{y}_{\text{obs}}$,
\begin{equation}
f_{\phi}^{-1}(\B{z}; \B{y}_{\text{obs}}) \sim p_{\text{post}} (\B{x}
\mid \B{y}_{\text{obs}}), \quad \B{z} \sim \mathrm{N} (\B{z} \mid \B{0}, \B{I}).
\label{sampling_amortized}
\end{equation}
 This step is illustrated in Figure~\ref{schematic_sampling}. As the
process above does not involve forward operator evaluations, sampling
with pretrained conditional normalizing flows is fast once an upfront
cost of amortized variational inference is incurred. In the next
section, we apply the above amortized variational inference to a seismic
imaging example in a controlled setting in which we assume no data
distribution shifts during inference.

\section{Validating amortized variational
inference}\label{validating-amortized-variational-inference}

The objective of this example is to apply amortized variational
inference to the high-dimensional seismic imaging problem. We show that
a relatively good pretrained conditional normalizing flow within the
context of amortized variational inference can be used to provide
approximate posterior samples for previously unseen seismic data that is
drawn from the same distribution as training seismic data. We begin by
introducing seismic imaging and describe challenges with Bayesian
inference in this problem.

\subsection{Seismic imaging}\label{seismic-imaging}

We are concerned with constructing an image of the Earth's subsurface
using indirect surface measurements that record the Earth's response to
synthetic sources being fired on the surface. The nonlinear relationship
between these measurements, known as shot records, and the
squared-slowness model of the Earth's subsurface is governed by the wave
equation. By linearizing this nonlinear relation, seismic imaging aims
to estimates the short-wavelength component of the Earth's subsurface
squared-slowness model. In its simplest acoustic form, the linearization
with respect to the slowness model---around a known, smooth background
squared slowness model $\B{m}_0$---leads to the following linear forward
problem:
\begin{equation}
\B{d}_i = \B{J}(\B{m}_0, \B{q}_i)
    \delta \B{m}^{\ast} + \boldsymbol{\epsilon}_i, \quad i = 1, \dots, N.
\label{linear-fwd-op}
\end{equation}
 We invert the above forward model to estimate the ground truth seismic
image $\delta \B{m}^{\ast}$ from $N$ processed (linearized) shot records
$\left \{\B{d}_{i}\right \}_{i=1}^{N}$ where $\B{J}(\B{m}_0, \B{q}_i)$
represents the linearized Born scattering operator \citep{gubernatis77}.
This operator is parameterized by the source signature $\B{q}_{i}$ and
the smooth background squared-slowness model $\B{m}_0$. Noise is denoted
by $\boldsymbol{\epsilon}_i$, and represents measurement noise and
linearization errors. While amortized variational inference does not
require knowing the closed from expression of the noise density to
simulate pairs of data and model (e.g., it is sufficient to be able to
simulate noise instances), for simplicity we assume the noise
distribution is a zero-centered Gaussian distribution with known
covariance $\sigma^2 \B{I}$. Due to the presence of shadow zones and
noisy finite-aperture shot data, wave-equation based linearized seismic
imaging (in short seismic imaging for the purposes of this paper)
corresponds to solving an inconsistent and ill-conditioned linear
inverse problem
\citep{lambare1992iterative, schuster1993least, nemeth1999least}. To
avoid the risk of overfitting the data and to quantify uncertainty, we
cast the seismic imaging problem into a Bayesian inverse problem
\citep{tarantola2005inverse}.

To address the challenge of Bayesian inference in this high-dimensional
inverse problem, we adhere to our amortized variational inference
framework. Within this approach, for an one-time upfront cost of
training a conditional normalizing flow, we get access to posterior
samples for previously unseen observed data that are drawn from the same
distribution as the distribution of training seismic data. This includes
data acquired in areas of the Earth with similar geologies, e.g., in
neighboring surveys. In addition, in our framework no explicit prior
density function needs to be chosen as the conditional normalizing flow
learns the prior distribution during pretraining from the collection of
seismic images in the training dataset. The implicitly learned prior
distribution by the conditional normalizing flow minimizes the risk of
negatively biasing the outcome of Bayesian inference by using overly
simplistic priors. In the next section, we describe the setup for our
amortized variational inference for seismic imaging.

\subsubsection{Acquisition geometry}\label{acquisition-geometry}

To mimic the complexity of real seismic images, we propose a
``quasi''-real data example in which we generate synthetic data by
applying the linearized Born scattering operator to $4750$ 2D sections
with size $3075\, \mathrm{m} \times 5120\, \mathrm{m}$ extracted from
the shallow section of the Kirchhoff migrated
\href{https://wiki.seg.org/wiki/Parihaka-3D}{Parihaka-3D} field dataset
\citep{Veritas2005, WesternGeco2012}. We consider a $12.5\, \mathrm{m}$
vertical and $20\, \mathrm{m}$ horizontal grid spacing, and we augment
an artificial $125\, \mathrm{m}$ water column on top of these images. We
parameterize the linearized Born scattering operator via a fictitious
background squared-slowness model, derived from the Kirchhoff migrated
images. To ensure good coverage, we simulate $102$ shot records with a
source spacing of $50\, \mathrm{m}$. Each shot is recorded for two
seconds with $204$ fixed receivers sampled at $25\, \mathrm{m}$ spread
on top of the model. The source is a Ricker wavelet with a central
frequency of $30\,\mathrm{Hz}$. To mimic a more realistic imaging
scenario, we add band-limited noise to the shot records, where the noise
is obtained by filtering white noise with the source wavelet
(Figure~\ref{d-obs}).

\subsubsection{Training configuration}\label{training-configuration}

Casting seismic imaging into amortized variational inference, as
described in this paper, is hampered by the high-dimensionality of the
data due to the multi-source nature of this inverse problem. To avoid
computational complexities associated with directly using $N$ shot
records as input to the conditional normalizing flow, we choose to
condition the conditional normalizing flow on the reverse-time migrated
image, which can be estimated by applying the adjoint of the linearized
Born scattering operator to the shot records,
\begin{equation}
\delta \B{m}_{\text{RTM}} = \sum_{i=1}^{N} \B{J}(\B{m}_0,
  \B{q}_i)^{\top} \B{d}_i.
\label{rtm}
\end{equation}
 While $\B{d}_i$ in the above expression is defined according to the
linearized forward model in equation~\ref{linear-fwd-op}, which does not
involve linearization errors, our method can handle observed data
simulated from wave-equation based nonlinear forward modeling.
Conditioning on the reverse-time migrated image and not on the shot
records directly may result in learning an approximation to the true
posterior distribution \citep{Adler_2022}. While technique from
statistics involving learned summary functions
\citep{radev2020bayesflow, schmitt2021bayesflow} can reduce the
dimensionality of the observed data, we propose to limit the Bayesian
inference bias induced by conditioning on the reverse-time migrated
image via our physics-based latent variable correction approach. We
leave utilizing summary functions in the context of seismic imaging
Bayesian inference to future work.

\begin{figure}
\centering
\includegraphics[width=0.600\hsize]{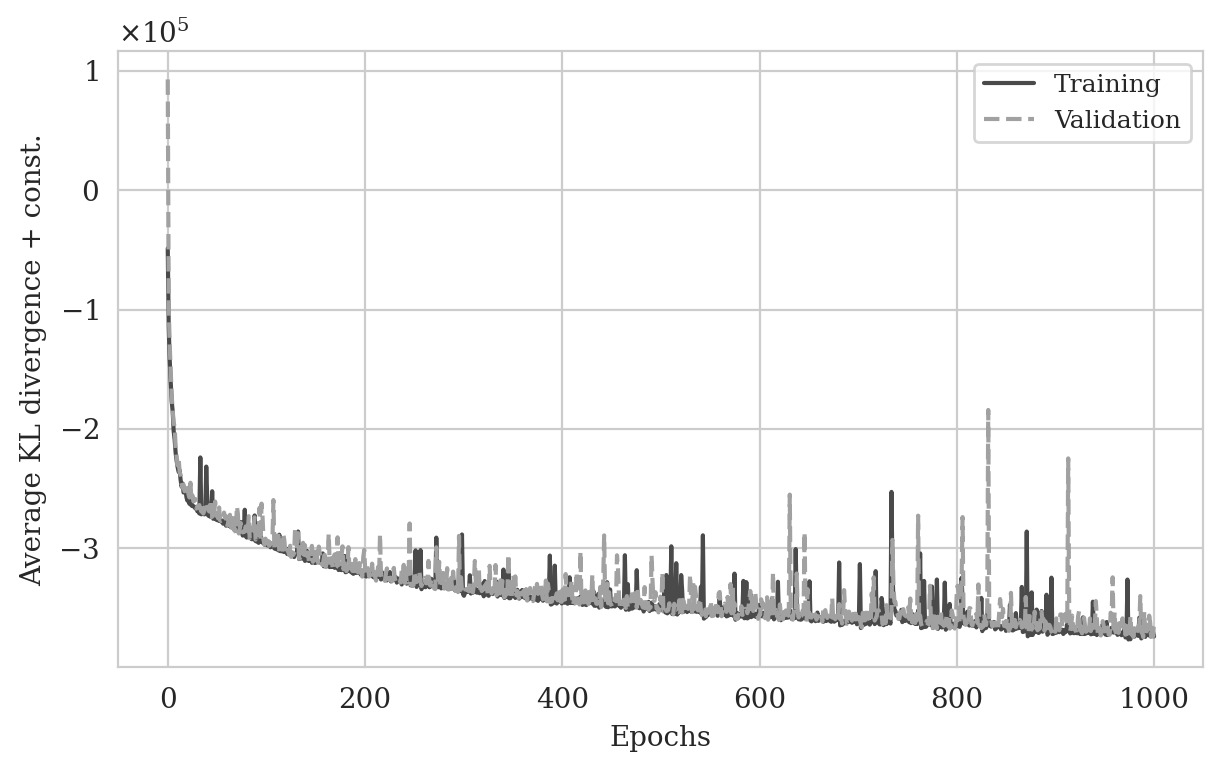}
\caption{Training and validation objective values as a function of
epochs. The validation objective value is computed over randomly
selected batched of the validation set at the end of each
epoch.}\label{seismic_avi_loss}
\end{figure}

To create training pairs,
$(\delta \B{m}^{(i)}, \delta \B{m}_{\text{RTM}}^{(i)}), \ i=1,\ldots,4750$,
we first simulate (see Figure~\ref{schematic_pretraining}) noisy seismic
data according to the above-mentioned acquisition design for all
extracted seismic images $\delta \B{m}^{(i)}$ from shallow sections of
the imaged Parihaka dataset. Next, we compute
$\delta \B{m}_{\text{RTM}}^{(i)}$ by applying reverse-time-migration to
the observed data for each image $\delta \B{m}^{(i)}$. As for the
conditional normalizing flow architecture, we follow
\citet{kruse2021hint} and use hierarchical normalizing flows due to
their increased expressiveness when compared to conventional invertible
architectures \citep{dinh2016density}. The expressive power of
hierarchical normalizing flows is a result of applying a series of
conventional invertible layers \citep{dinh2016density} to different
scales of the input in a hierarchical manner (refer to
\citet{kruse2021hint} for a schematic representation of the
architecture). This leads to a invertible architecture with a dense
Jacobian \citep{kruse2021hint} that is capable of representing
complicated bijective transformations. We train this conditional
normalizing flow on the pairs
$(\delta \B{m}^{(i)}, \delta \B{m}_{\text{RTM}}^{(i)}), \ i=1,\ldots,4750$
according to the objective function in equation~\ref{vi_forward_kl_nf}
with the Adam stochastic optimization method \citep{kingma2014adam} with
a batchsize of $16$ for one thousand passes over the training dataset
(epochs). We use an initial stepsize of $10^{-4}$ and decrease it after
each epoch until reaching the final stepsize of $10^{-6}$. To monitor
overfitting, we evaluate the objective function at the end of every
epoch over random subsets of the validation set, consisting of $530$
seismic images extracted from the shallow sections of the imaged
Parihaka dataset and the associated reverse-time migrated images. As
illustrated in Figure~\ref{seismic_avi_loss}, the training and
validation objective values exhibit a decreasing trend, which suggests
no overfitting. We stopped the training after one thousand epochs due to
a slowdown in the decrease of the training and validation objective
values.

\subsection{Results and observations}\label{results-and-observations}

\begin{figure*}
\centering
\subfloat[\label{d-noise-free}]{\includegraphics[width=0.350\hsize]{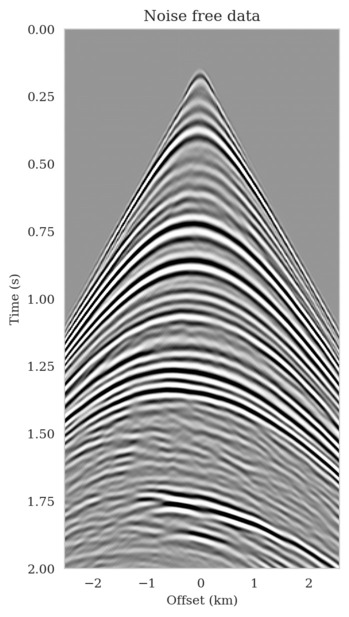}}
\subfloat[\label{d-obs}]{\includegraphics[width=0.350\hsize]{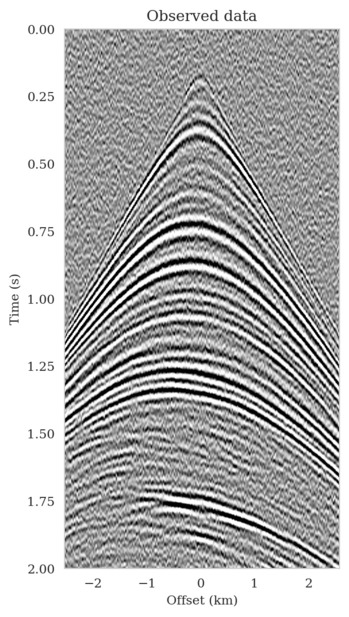}}
\caption{A shot record generated from an image extracted from the
Parihaka dataset. (a) Noise-free linearized data. (b) Linearized data
with bandwidth-limited noise.}\label{data}
\end{figure*}

Following training, the pretrained conditional normalizing flow is able
to produce samples from the posterior distribution for seismic data not
used in training. These samples resemble different regularized (via the
learned prior) least-squares migration images that explain the observed
data. To demonstrate this, we simulate seismic data for a previously
unseen perturbation model using the forward model~\ref{linear-fwd-op}
with the same noise variance. Figure~\ref{data} shows an example of a
single noise-free (Figure~\ref{d-noise-free}) and noisy
(Figure~\ref{d-obs}) shot record for one of $102$ sources.

\begin{figure}
\centering
\subfloat[\label{true_model}]{\includegraphics[width=0.650\hsize]{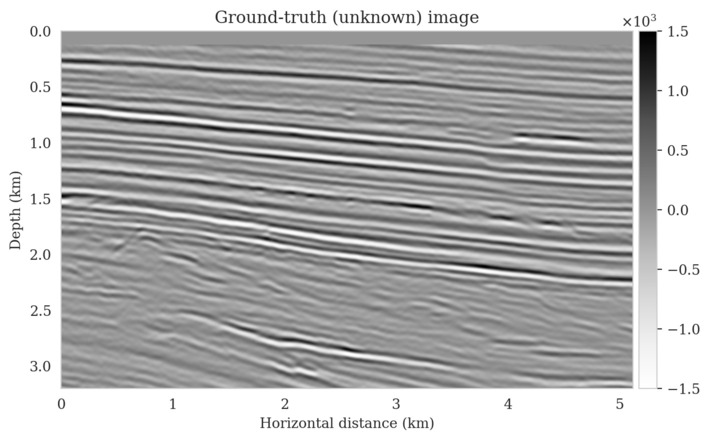}}
\\
\subfloat[\label{rtm}]{\includegraphics[width=0.650\hsize]{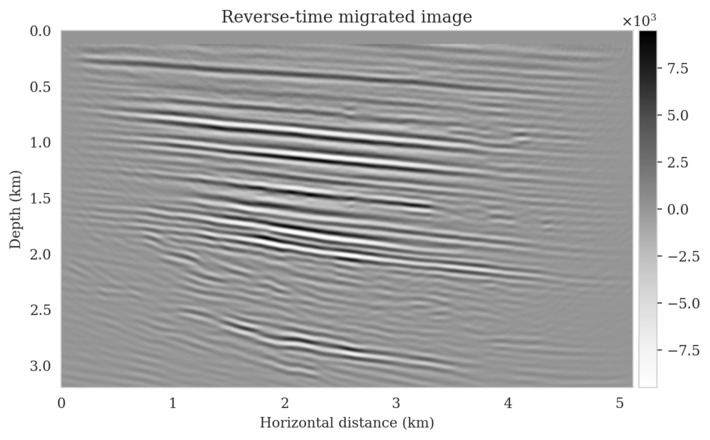}}
\caption{Amortized variational inference testing phase setup. (a)
High-fidelity ground-truth image. (b) Reverse-time migrated image with
SNR $- 12.17\,\mathrm{dB}$.}\label{seismic_avi_setup}
\end{figure}

We perform reverse-time migration to obtain the necessary input for the
conditional normalizing flow to obtain posterior samples. We show the
ground-truth seismic image (to be estimated) and the resulting
reverse-time migrated image in Figures~\ref{true_model} and~\ref{rtm},
respectively. Clearly, the reverse-time migrated image has grossly wrong
amplitudes, and more importantly, due to limited-aperture shot data, the
edges of the image are not well illuminated.

\begin{figure}
\centering
\captionsetup[subfigure]{labelformat=empty}
\subfloat[]{\includegraphics[width=0.500\hsize]{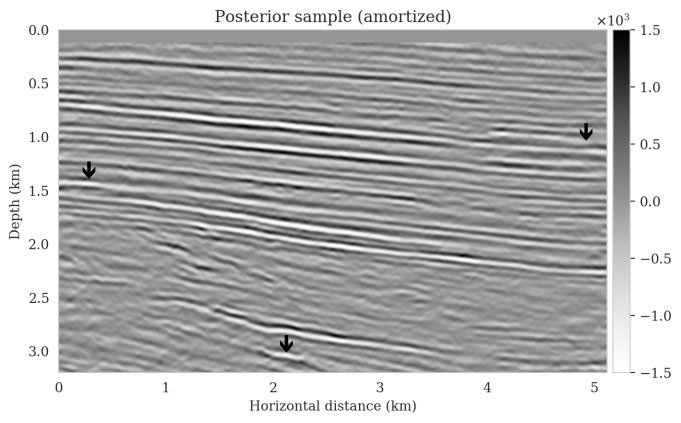}}
\subfloat[]{\includegraphics[width=0.500\hsize]{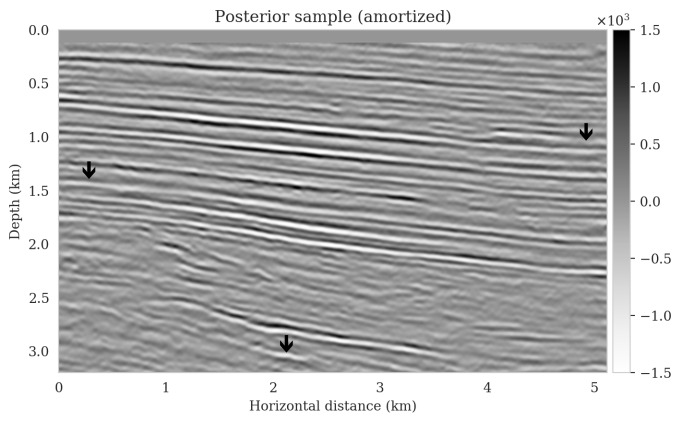}}
\\
\subfloat[]{\includegraphics[width=0.500\hsize]{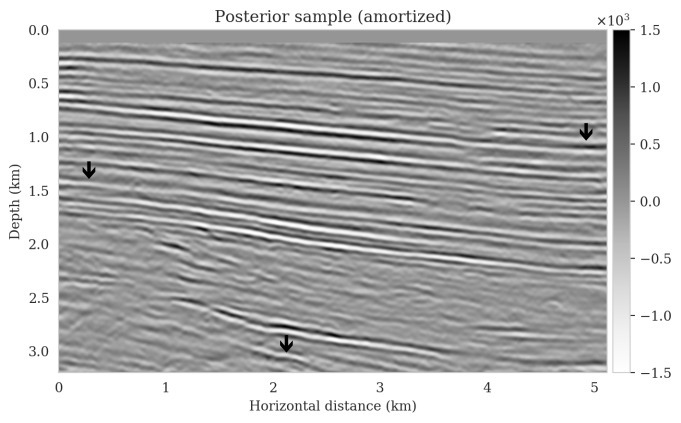}}
\subfloat[]{\includegraphics[width=0.500\hsize]{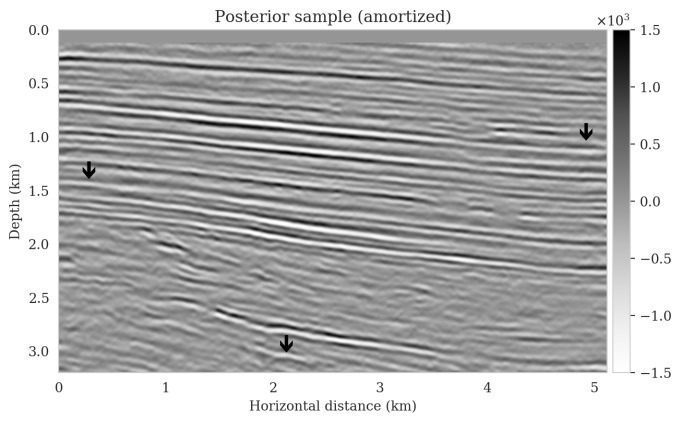}}
\\
\subfloat[]{\includegraphics[width=0.500\hsize]{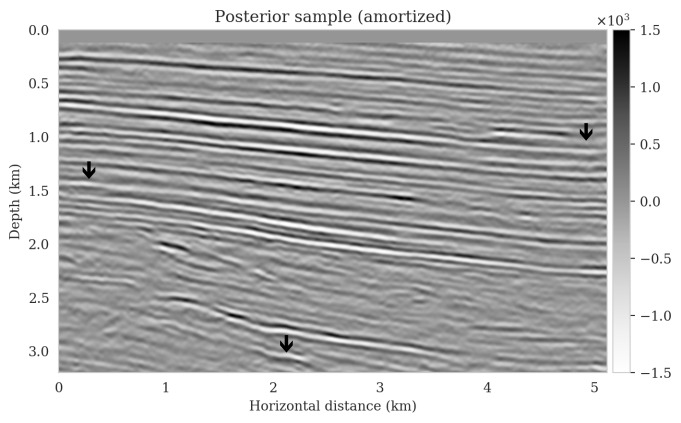}}
\subfloat[]{\includegraphics[width=0.500\hsize]{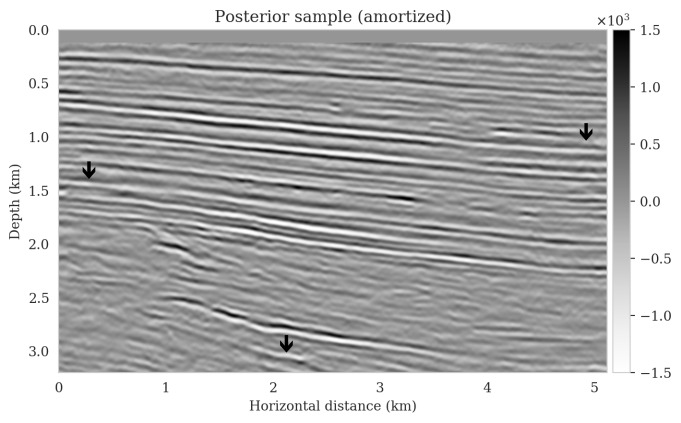}}
\caption{Samples drawn from posterior distribution using the pretrained
conditional normalizing flow via equation~\ref{amortization} with SNRs
ranging from $8.08\,\mathrm{dB}$ to
$8.92\,\mathrm{dB}$.}\label{seismic_avi_samples}
\end{figure}

We obtain one thousand posterior samples by providing the reverse-time
migrated image and latent samples drawn from the standard Gaussian
distribution to the pretrained conditional normalizing flow
(equation~\ref{amortization}). This process is fast as it does not
require any forward operator evaluations. To illustrate the variability
among the posterior samples, we show six of them in
Figure~\ref{seismic_avi_samples}. As shown in
Figure~\ref{seismic_avi_samples}, these image samples have amplitudes in
the same range as the ground-truth image and better predict the
reflectors at the edges of the image compared to the reverse-time
migration image (Figure~\ref{rtm}). In addition, the posterior samples
indicate improved imaging in deep regions, which is typically more
difficult due to the placement of the sources and receiver near the
surface.

\begin{figure}
\centering
\subfloat[\label{avi_cm}]{\includegraphics[width=0.650\hsize]{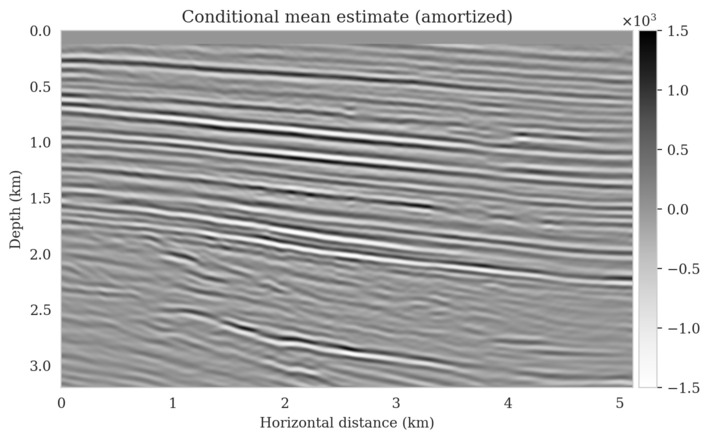}}
\\
\subfloat[\label{avi_std}]{\includegraphics[width=0.650\hsize]{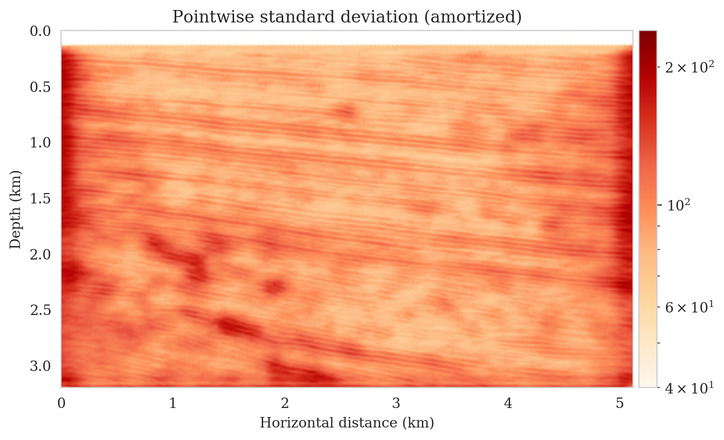}}
\\
\subfloat[\label{avi_std_normalized}]{\includegraphics[width=0.650\hsize]{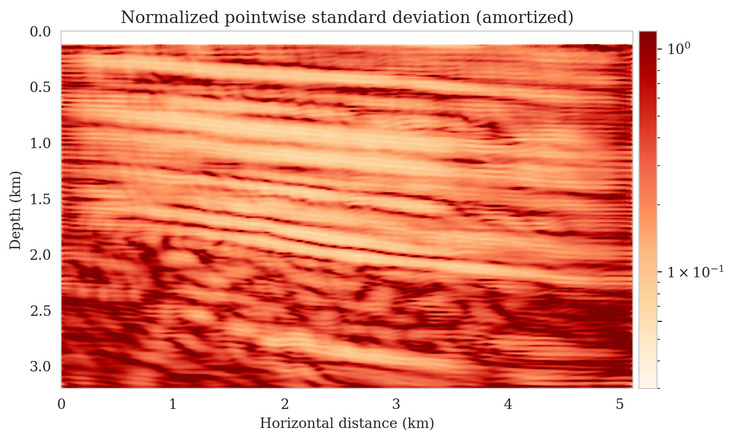}}
\caption{Amortized variational inference results. (a) The conditional
(posterior) mean estimate with SNR $9.44\,\mathrm{dB}$. (b) The
pointwise standard deviation among samples drawn from the posterior. (c)
Normalized pointwise standard deviation by the conditional mean estimate
(Figure~\ref{avi_cm}).}\label{seismic_avi_results}
\end{figure}

Samples from the posterior provide access to useful statistical
information including approximations to moments of the distribution such
as the mean and pointwise standard deviation
(Figure~\ref{seismic_avi_results}). We compute the mean of the posterior
samples to obtain the conditional mean estimate, i.e., the expected
value of the posterior distribution. This estimate is depicted in
Figure~\ref{avi_cm}. From Figure~\ref{avi_cm}, we observe that the
overall amplitudes are well recovered by the conditional mean estimate,
which includes partially recovered reflectors in badly illuminated areas
close to the boundaries. Although the reconstructions are not perfect,
they significantly improve upon the reverse-time migrations estimate. We
did not observe a significant increase in the signal-to-noise ratio
(SNR) of the conditional mean estimate when more than one thousand
samples from the posterior are drawn. We use the one thousand samples to
also estimate the pointwise standard deviation (Figure~\ref{avi_std}),
which serves as an assessment of the uncertainty. To avoid bias from
strong amplitudes in the estimated image, we also plot the stabilized
division of the standard deviation by the envelope of the conditional
mean in Figure~\ref{avi_std_normalized}. As expected, the pointwise
standard deviation in Figures~\ref{avi_std} and~\ref{avi_std_normalized}
indicate that we have the most uncertainty in areas of complex
geology---e.g., near channels and tortuous reflectors, and in areas with
a relatively poor illumination (deep and close to boundaries). The areas
with large uncertainty align well with difficult-to-image parts of the
model. The normalized pointwise standard deviation
(Figure~\ref{avi_std_normalized}) aims to visualize an
amplitude-independent assessment of uncertainty, which indicates high
uncertainty on the onset and offset of reflectors (both shallow and
deeper sections), while showing low uncertainty in the areas of the
image with no reflectors.

After incurring an upfront cost of training the conditional normalizing
flow, the computational cost of sampling the posterior distribution is
low as it does not involve any forward operator evaluations. However,
the accuracy of the presented results is directly linked to the
availability of high-quality training data that fully represent the
joint distribution for model and data. Due to our lack of access to the
subsurface of the Earth, obtaining high-quality training data is
challenging when dealing with geophysical inverse problems. To address
this issue, we propose to supplement amortized variational inference
with a physics-based latent distribution correction technique that
increases the reliability of this approach when dealing with moderate
shifts in the data distribution during inference.

\section{A physics-based treatment to data distribution shifts during
inference}\label{a-physics-based-treatment-to-data-distribution-shifts-during-inference}

For accurate Bayesian inference in the context of amortized variational
inference, the surrogate conditional distribution
$p_{\phi}( \B{x} \mid \B{y})$ must yield a zero amortized variational
inference objective value (equation~\ref{vi_forward_kl}). Achieving this
objective is challenging due to lack of access to model and data pairs
that sufficiently captures the underlying joint distribution in
equation~\ref{vi_forward_kl}. Additionally, due to potential shifts to
the joint distribution during inference, i.e., shifts in the prior
distribution or the forward (likelihood) model
(equation~\ref{forward_model}), the conditional normalizing flow can no
longer reliably provide samples from the posterior distribution due to
lack of generalization. Under such conditions feeding latent samples
drawn from a standard Gaussian distribution to the conditional
normalizing flow may lead to posterior sampling errors. To quantify the
posterior distribution approximation error and to propose our correction
method, we will use the invariance of the KL divergence to
differentiable and invertible mappings \citep{liu2014divergence}. This
property relates conditional normalizing flow's error in posterior
distribution approximation to its error in Gaussianizing the input model
and data pairs.

\subsection{KL divergence invariance
relation}\label{kl-divergence-invariance-relation}

The errors that the pretrained conditional normalizing flow makes in
approximating the posterior distribution can be formally quantified
using the invariance of the KL divergence under diffeomorphism mappings
\citep{liu2014divergence}. Using this relation, we relate the posterior
distribution approximation errors (KL divergence between true and
predicted posterior) to the errors that the conditional normalizing flow
makes in gaussianizing its inputs (KL divergence between the
distribution of ``gaussianized'' inputs and standard Gaussian
distribution). Specifically, for observed data $\B{y}_{\text{obs}}$
drawn from a shifted data distribution
$\widehat{p}_{\text{data}} (\B{y}) \neq p_{\text{data}} (\B{y})$, the
invariance relation states
\begin{equation}
\KL\,\left(p_{\phi} (\B{z} \mid \B{y}_{\text{obs}}) \mid\mid
\mathrm{N}(\B{z} \mid \B{0}, \B{I})\right) = \KL\,
\left(p_{\text{post}}(\B{x} \mid \B{y}_{\text{obs}})
\mid\mid p_{\phi}(\B{x} \mid \B{y}_{\text{obs}})\right) > 0.
\label{kl_divergence_invariance}
\end{equation}
 In this expression, $p_{\phi} (\B{z} \mid \B{y}_{\text{obs}})$
represents the distribution of conditional normalizing flow output
$\B{z} = f_{\phi}(\B{x}; \B{y}_{\text{obs}})$. That is, passing inputs
$\B{x} \sim p(\B{x} \mid \B{y}_{\text{obs}})$ for one instance of
observed data
$\B{y}_{\text{obs}} \sim \widehat{p}_{\text{data}} (\B{y})$ to the
conditional normalizing flow implicitly defines a (conditional)
distribution $p_{\phi} (\B{z} \mid \B{y}_{\text{obs}})$ in the
conditional normalizing flow output space. We refer to this distribution
as the shifted latent distribution as it is the result of a data
distribution shift translated through the conditional normalizing flow
to the latent space. The data distribution shifts can be caused by
changes in number of sources, noise distribution, wavelet source
frequency, and geological features to be imaged.
Equation~\ref{kl_divergence_invariance} states that the conditional
normalizing flow fails to accurately Gaussianize the input models
$\B{x} \sim p(\B{x} \mid \B{y}_{\text{obs}})$ for the given data
$\B{y}_{\text{obs}}$. Failure to take into account the mismatch between
the shifted latent distribution
$p_{\phi} (\B{z} \mid \B{y}_{\text{obs}})$ and
$\mathrm{N}(\B{z} \mid \B{0}, \B{I})$ leads to posterior sampling errors
as the KL divergence between the predicted and true posterior
distributions is nonzero (equation~\ref{kl_divergence_invariance}). In
other words, feeding latent samples drawn from a standard Gaussian
distribution to $f_{\phi}^{-1}(\B{z}; \B{y}_{\text{obs}})$ produces
samples from $p_{\phi}(\B{x} \mid \B{y}_{\text{obs}})$, which does not
accurately approximate the true posterior distribution under the
assumption of data distribution shift. On the other hand, with the same
reasoning via the KL divergence invariance relation, feeding samples
from the shifted latent distribution
$p_{\phi}(\B{z} \mid \B{y}_{\text{obs}})$ to the conditional normalizing
flow yields accurate posterior samples. However, obtaining samples from
$p_{\phi}(\B{z} \mid \B{y}_{\text{obs}})$ is not trivial as we do not
have a closed-form expression for its density. In the next section, we
introduce a physics-based approximation to the shifted-latent
distribution.

\subsection{Physics-based latent distribution
correction}\label{physics-based-latent-distribution-correction}

Ideally, performing accurate posterior sampling via the pretrained
conditional normalizing flow---in the presence of data distribution
shifts---requires passing samples from the shifted latent distribution
$p_{\phi} (\B{z} \mid \B{y}_{\text{obs}})$ to
$f_{\phi}^{-1}(\B{z}; \B{y}_{\text{obs}})$. Unfortunately, accurately
sampling $p_{\phi} (\B{z} \mid \B{y}_{\text{obs}})$ requires access to
the true posterior distribution, which we are ultimately after and do
not have access to. Alternatively, we propose to quantify
$p_{\phi} (\B{z} \mid \B{y}_{\text{obs}})$ using Bayes' rule,
\begin{equation}
p_{\phi} (\B{z} \mid \B{y}_{\text{obs}}) = \frac{p_{\text{like}}
(\B{y}_{\text{obs}} \mid \B{z})\, p_{\text{prior}}
(\B{z})}{\widehat{p}_{\text{data}} (\B{y}_{\text{obs}})},
\label{bayes_rule_physic_inf}
\end{equation}
 where the physics-informed likelihood function
$p_{\text{like}} (\B{y}_{\text{obs}} \mid \B{z})$ and the prior
distribution $p_{\text{prior}} (\B{z})$ over the latent variable are
defined as
\begin{equation}
\begin{aligned}
- \log p_{\phi}(\B{z} \mid \B{y}_{\text{obs}})  & = - \sum_{i=1}^{N}
\log p_{\text{like}} (\B{y}_{\text{obs}, i} \mid \B{z} )
- \log p_{\text{prior}} (\B{z}) + \log \widehat{p}_{\text{data}}
(\B{y}_{\text{obs}}) \\
& := \frac{1}{2 \sigma^2} \sum_{i=1}^{N}
\big \|\B{y}_{\text{obs}, i} - \mathcal{F}_i \circ
f_{\phi}^{-1}(\B{z}; \B{y}_{\text{obs}}) \big\|_2^2
+ \frac{1}{2 } \big \| \B{z} \big \|_2^2 + \text{const}.
\end{aligned}
\label{physics_informed_density}
\end{equation}
 In the above expression, the physics-informed likelihood function
$p_{\text{like}} (\B{y}_{\text{obs}} \mid \B{z})$ follows from the
forward model in equation~\ref{forward_model} with a Gaussian assumption
on the noise with mean zero and covariance matrix $\sigma^2 \B{I}$, and
the prior distribution $p_{\text{prior}} (\B{z})$ is chosen as a
standard Gaussian distribution with mean zero and covariance matrix
$\B{I}$. The choice of the likelihood function ensures physics and data
fidelity by giving more importance to latent variables that once passed
through the pretrained conditional normalizing flow and the forward
operator provide smaller data misfits while the prior distribution
$p_{\text{prior}} (\B{z})$ injects our prior beliefs about the latent
variable, which is by design chosen to be distributed according to a
standard Gaussian distribution.

Due to our choice of the likelihood function and prior distribution
above, the effective prior distribution over the unknown $\B{x}$ is in
fact a conditional prior characterized by the pretrained conditional
normalizing flow \citep{pmlr-v119-asim20a}. As observed by
\citet{yang2018conditional} and \citet{orozco2021photoacoustic}, using a
conditional prior may be more informative than its unconditional
counterpart because it is conditioned by the observed data
$\B{y}_{\text{obs}}$. Our approach can be also viewed as an instance of
online variational Bayes \citep{zeno2018task} where data arrives
sequentially and previous posterior approximates are used as priors for
subsequent approximations.

In the next section, we improve the available amortized approximation to
the posterior distribution by relaxing the standard Gaussian
distribution assumption of the conditional normalizing flow latent
distribution.

\subsubsection{Gaussian relaxation of the latent
distribution}\label{gaussian-relaxation-of-the-latent-distribution}

\begin{figure*}
\centering
\includegraphics[width=1.000\hsize]{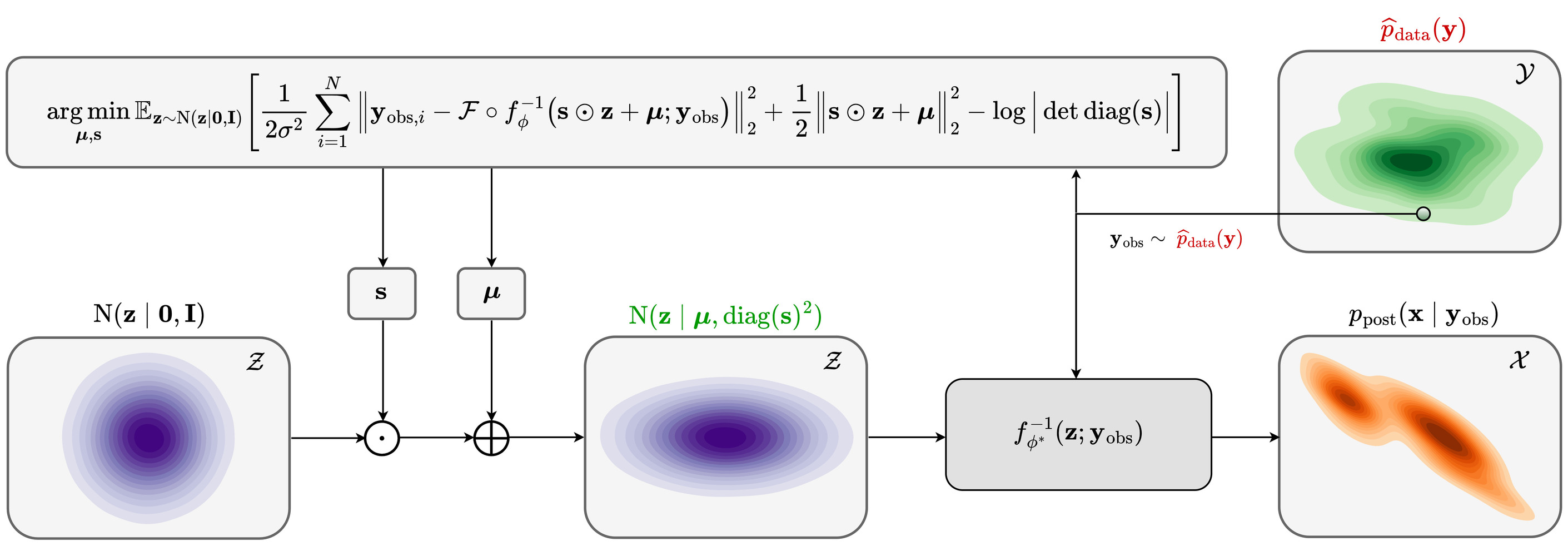}
\caption{A schematic representation of out proposed method. When dealing
with nonzero amortized variational inference objective value
(equation~\ref{average_forward_kl}) or in presence of data distribution
shifts during inference, we correct the latent distribution of the
pretrained conditional normalizing flow via a diagonal physics-based
correction. After the correction, the new latent samples result in
corrected posterior samples when fed to the pretrained normalizing
flow.}\label{schematic_correction}
\end{figure*}

By definition, feeding samples from
$p_{\phi}(\B{z} \mid \B{y}_{\text{obs}})$ to the pretrained amortized
conditional normalizing flows provides samples from the posterior
distribution (see discussion beneath
equation~\ref{kl_divergence_invariance}). To maintain the low
computational cost of sampling with amortized variational inference, it
is imperative that $p_{\phi}(\B{z} \mid \B{y}_{\text{obs}})$ is sampled
as cheaply as possible. To this end, we exploit the fact that
conditional normalizing flows in the context of amortized variational
inference are trained to Gaussianize the input model random variable
(equation~\ref{vi_forward_kl_nf}). This suggests that the shifted latent
distribution $p_{\phi}(\B{z} \mid \B{y}_{\text{obs}})$ will be close to
a standard Gaussian distribution for a certain class of data
distribution shifts. We exploit this property and approximate the
shifted latent distribution $p_{\phi} (\B{z} \mid \B{y}_{\text{obs}})$
via a Gaussian distribution with an unknown mean and diagonal covariance
matrix,
\begin{equation}
p_{\phi} (\B{z} \mid \B{y}_{\text{obs}}) \approx
\mathrm{N} \big(\B{z} \mid \Bs{\mu},
\operatorname{diag}(\B{s})^2\big), \quad \B{z}
\in \mathcal{Z}.
\label{gaussian_approximation}
\end{equation}
 In the above expression, the vector $\Bs{\mu}$ corresponds to the mean
and the vector $\operatorname{diag}(\B{s})^2$ represents a diagonal
covariance matrix with diagonal entries $\B{s} \odot \B{s}$ (with the
symbol $\odot$ denoting elementwise multiplication) that need to be
determined. We estimate these quantities by minimizing the reverse KL
divergence between the relaxed Gaussian latent distribution
$\mathrm{N} \big(\B{z} \mid \Bs{\mu}, \operatorname{diag}(\B{s})^2\big)$
and the shifted latent distribution
$p_{\phi}(\B{z} \mid \B{y}_{\text{obs}})$. According to the variational
inference objective function associated with the reverse KL divergence
in equation~\ref{vi_reverse_kl}, this correction can be achieved by
solving the following optimization problem (see derivation in Appendix
A),
\begin{equation}
\begin{aligned}
\Bs{\mu}^{\ast}, \B{s}^{\ast} & =
\argmin_{\Bs{\mu}, \B{s}} \KL\,\big(\mathrm{N}
\big(\B{z} \mid \Bs{\mu},
\operatorname{diag}(\B{s})^2\big) \mid\mid
p_{\phi} (\B{z} \mid \B{y}_{\text{obs}}) \big) \\
& = \argmin_{\Bs{\mu}, \B{s}}
\mathbb{E}_{\B{z} \sim \mathrm{N} (\B{z} \mid \B{0}, \B{I})}
\bigg [\frac{1}{2 \sigma^2} \sum_{i=1}^{N}
\big  \| \B{y}_{\text{obs}, i}-\mathcal{F}_i \circ
f_ {\phi} \big(\B{s} \odot\B{z}
+ \Bs{\mu}; \B{y}_{\text{obs}} \big) \big\|_2^2 \\
& \qquad \qquad \qquad \qquad \quad + \frac{1}{2}
\big\| \B{s} \odot \B{z}
+ \Bs{\mu} \big\|_2^2 - \log
\Big | \det \operatorname{diag}(\B{s}) \Big | \bigg ].
\end{aligned}
\label{reverse_kl_covariance_diagonal}
\end{equation}
 We solve optimization problem~\ref{reverse_kl_covariance_diagonal} with
the Adam optimizer where we select random batches of latent variable
variables $\B{z} \sim \mathrm{N} (\B{z} \mid \B{0}, \B{I})$ and data
indices. We initialize the optimization
problem~\ref{reverse_kl_covariance_diagonal} by $\Bs{\mu} = \B{0}$ and
$\operatorname{diag}(\B{s})^2 = \B{I}$. This initialization acts as a
warm-start and an implicit regularization \citep{pmlr-v119-asim20a}
since $f_{\phi}^{-1}(\B{z}; \B{y}_{\text{obs}})$ for standard Gaussian
distributed latent samples $\B{z}$ provides approximate samples from the
posterior distribution---thanks to amortization over different observed
data $\B{y}$. As a result, we expect the optimization
problem~\ref{reverse_kl_covariance_diagonal} to be solved relatively
cheaply. Additionally, the imposed standard Gaussian distribution prior
on $\B{s} \odot \B{z} + \Bs{\mu}$ regularizes inversion for the
corrections since
$\KL\,\left(p_{\phi} (\B{z} \mid \B{y}_{\text{obs}}) \mid\mid \mathrm{N}(\B{z} \mid \B{0}, \B{I})\right)$
is minimized during amortized variational inference
(equation~\ref{kl_divergence_invariance}). To relax the (conditional)
prior imposed by the pretrained conditional normalizing flow, instead of
a standard Gaussian prior, a Gaussian prior with a larger variance can
be imposed on the corrected latent variable. Conditional normalizing
flows' inherent invertibility allows the normalizing flow to represent
any solution $\B{x} \in \mathcal{X}$ in the solution space. This has the
additional benefit of limiting the adverse affects of imperfect
pretraining of $f_{\phi}$ in domains where access to high-fidelity
training data is limited, which is often the case in practice. The
output of the conditional normalizing flow can be further regularized by
including additional regularization terms in
equation~\ref{reverse_kl_covariance_diagonal} to prevent it from
producing out-of-range, non-physical results.
Figure~\ref{schematic_correction} summarizes our proposed method latent
distribution correction method.

\subsubsection{Inference with corrected latent
distribution}\label{inference-with-corrected-latent-distribution}

Once the optimization problem~\ref{reverse_kl_covariance_diagonal} is
solved with respect to $\Bs{\mu}$ and $\B{s}$, we obtain corrected
posterior samples by passing samples from the corrected latent
distribution
$\mathrm{N} \big(\B{z} \mid \Bs{\mu}^{\ast}, \operatorname{diag}(\B{s}^{\ast})^2\big) \approx p_{\phi} (\B{z} \mid \B{y}_{\text{obs}})$
to the conditional normalizing flow,
\begin{equation}
\B{x} = f_{\phi}^{-1}(\B{s}^{\ast} \odot\B{z}
+ \Bs{\mu}^{\ast}; \B{y}_{\text{obs}}), \quad \B{z} \sim \mathrm{N}
(\B{z} \mid \B{0}, \B{I}).
\label{inference_after_correction}
\end{equation}
 These corrected posterior samples are implicitly regularized by the
reparameterization with the pretrained conditional normalizing flow and
the standard Gaussian distribution prior on $\B{z}$
\citep{pmlr-v119-asim20a, orozco2021photoacoustic, siahkoohi2022EAGEweb}.
Next section applies this physics-based correction to a seismic imaging
example, in which we use the pretrained conditional normalizing flow
from the earlier example.

\section{Latent distribution correction applied to seismic
imaging}\label{latent-distribution-correction-applied-to-seismic-imaging}

The purpose of our proposed latent distribution correction approach is
to accelerate Bayesian inference while maintaining fidelity to a
specific observed dataset and physics. While this method is generic and
can be applied to a variety of inverse problems, it is particularly
relevant when solving geophysical inverse problems, where the unknown
quantity is high dimensional, the forward operator is computationally
costly to evaluate, and there is a lack of access to high-quality
training data that represents the true heterogeneity of the Earth's
subsurface. Therefore, we apply this approach to seismic imaging to
utilize the advantages of generative models for solving inverse
problems, including fast conditional sampling and learned prior
distributions, while limiting the negative bias induced by shifts in
data distributions.

The results are presented for two cases. The first case involves
introducing a series of changes in the distribution of observed data,
for example changing the number of sources and the noise levels. This is
followed by correcting for the error in predictions made by the
pretrained conditional normalizing flow using our proposed method. In
the second case, in addition to the shifts in the distribution of the
observed data (forward model), we also introduce a shift in the prior
distribution. We accomplish this by selecting a ground-truth image from
a deeper section of the Parihaka dataset that has different image
characteristics than the training images, such as tortuous reflectors
and more complex geological features. In both cases, we expect the
outcome of the Bayesian algorithm to improve following the correction of
latent distributions described above. We provide qualitative and
quantitative evaluations of the Bayesian inference results.

\subsection{Shift in the forward
model}\label{shift-in-the-forward-model}

In the following example, we introduce shifts in the distribution of
observed data---compared to the pretraining phase---by changing the
forward model. The shift involves reducing the number of sources ($N$ in
equation~\ref{forward_model}) by a factor of two to four, while adding
band-limited noise with 1.5 to three times larger standard deviation
($\sigma$ in equation~\ref{bayes_rule_log}). We will demonstrate the
potential pitfalls of relying solely on the pretrained conditional
normalizing flows in circumstances where the distribution of observed
data has shifted. With the use of our latent distribution correction, we
will demonstrate that we are able to correct for errors that are made by
the pretrained conditional normalizing flow as a result of changes to
data distribution.

Following the description of the problem setup, we will also provide
comparisons between the conditional mean estimation quality before and
after the latent distribution correction step. Before moving on to our
results relating to uncertainty quantification on the image, we
demonstrate the importance of the correction step by visualizing the
improvements in fitting the observed data. Lastly, we perform a series
of experiments to verify our Bayesian inference results.

\subsubsection{Problem setup}\label{problem-setup}

To induce shifts in the data distribution, we reduce the number of
sources and increase the standard deviation of the added band-limited
noise. Consequently, we have reduced the amount of data (due to having
fewer source experiments) and decreased the SNR of each shot record (due
to being contaminated with stronger noise). As a consequence, seismic
imaging becomes more challenging, i.e., more difficult to estimate the
ground truth image, and it is expected that the uncertainty associated
with the problem will also increase.

We use the same ground-truth image as in the previous example
(Figure~\ref{true_model}), while experimenting with 25, 51, 102 sources
and adding band-limited noise that has $1.5$, $2.0$, $2.5$, and $3.0$
times larger standard deviation than the pretraining setup. For each
combination of source number and noise level ($12$ combinations in
total), we compute the reverse-time migrated image corresponding to that
combination. Next, we perform latent distribution corrections for each
of the $12$ seismic imaging instances. All latent distribution
correction optimization problems
(equation~\ref{reverse_kl_covariance_diagonal}) are solved using the
Adam optimization algorithm \citep{kingma2014adam} for five passes over
the shot records (epochs). We did not observe a significant decrease in
the objective function after five epochs. The objective function is
evaluated each iteration by drawing a single latent sample from the
standard Gaussian distribution and randomly selecting (without
replacement) a data index $i \in \{1, \ldots, 25 \}$. We use a stepsize
of $10^{-1}$ and decrease it by a factor of $0.9$ at the end of every
two epochs.

After solving the optimization
problem~\ref{reverse_kl_covariance_diagonal} for the different seismic
imaging instances, we obtain corrected posterior samples for each
instance. The next section provides a detailed discussion of the latent
distribution correction that was applied to one such instance that had a
significant shift in data distribution.

\subsubsection{Improved Bayesian inference via latent distribution
correction}\label{improved-bayesian-inference-via-latent-distribution-correction}

\begin{figure*}
\centering
\subfloat[\label{d-noise-free_ldc_0}]{\includegraphics[width=0.350\hsize]{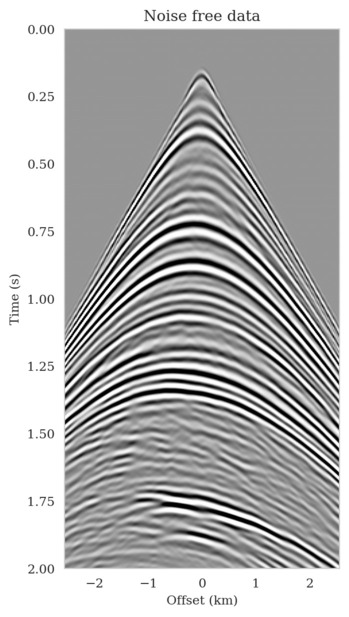}}
\subfloat[\label{d-obs_ldc_0}]{\includegraphics[width=0.350\hsize]{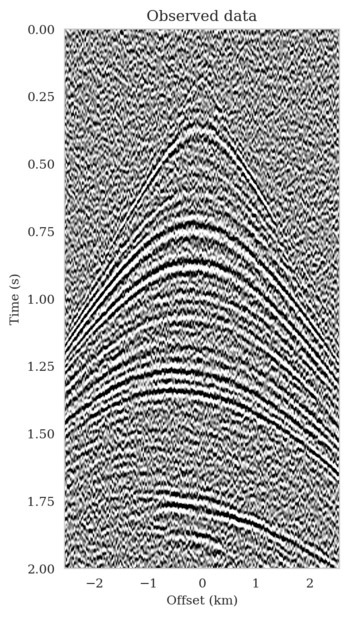}}
\caption{A shot record from the shifted data distribution. (a)
Noise-free linearized data (same as Figure~\ref{d-noise-free}). (b)
Noisy linearized data with $2.5$ larger band-limited noise standard
deviation (SNR $-2.78\,\mathrm{dB}$).}\label{data_ldc_0}
\end{figure*}

The aim of this section is to demonstrate how latent distribution
correction can be used to mitigate errors induced by data distribution
shifts. Specifically, we present the results for the case where the
number of sources is reduced by a factor of four ($N = 25$) as compared
to the pretrained data generation setup. The $25$ sources are spread
periodically over the survey area with a source sampling of
approximately $200$ meters. Moreover, we contaminate the resulting shot
records with band-limited noise with an increased standard deviation of
$2.5$ times when compared to the pretraining phase. The overall SNR for
the data thus becomes $-2.78\,\mathrm{dB}$, which is $7.95\,\mathrm{dB}$
lower that the SNR of the observed data during pretraining
(Figure~\ref{d-obs}). Figure~\ref{data_ldc_0} shows one of the
`\texttt{25} shot records.

\begin{figure}
\centering
\subfloat[\label{rtm_ldc_0}]{\includegraphics[width=0.650\hsize]{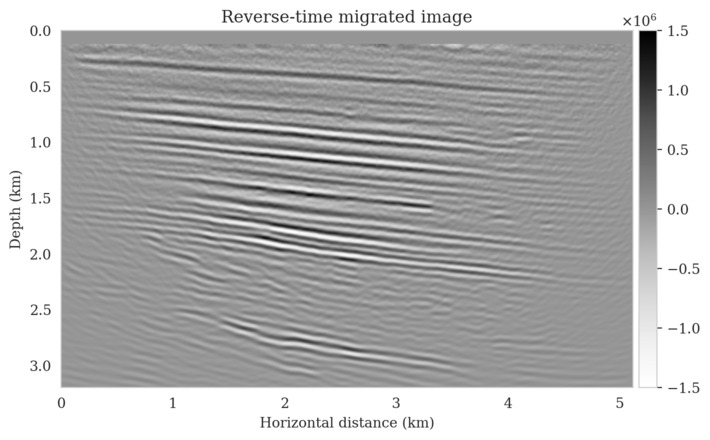}}
\\
\subfloat[\label{lsrtm_ldc_0}]{\includegraphics[width=0.650\hsize]{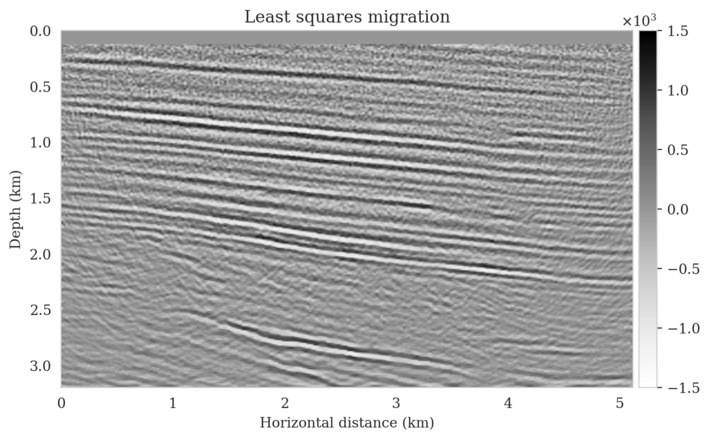}}
\caption{Latent distribution correction experiment setup. (a)
Reverse-time migrated image corresponding to the shifted forward model
with SNR $-8.22\,\mathrm{dB}$. (b) Least squares imaging, which is
equivalent to the minimizing
$\sum_{i=1}^{N} \big \| \B{d}_i-\B{J}(\B{m}_0, \B{q}_i)\delta \B{m} \big\|_2^2$
with respect to $\delta \B{m}$ with no regularization. The SNR for this
estimate is $6.90\,\mathrm{dB}$.}\label{seismic_ldc_0_setup}
\end{figure}

Utilizing the above observed dataset, we compute the reverse-time
migrated image as an input to our pretrained conditional normalizing
flow (Figure~\ref{rtm_ldc_0}). In contrast to the reverse-time migrated
image shown in Figure~\ref{d-obs}, this migrated image is, as expected,
noisier, and it displays visible near-source imaging artifacts as a
result of coarse source sampling. Additionally, we compute the
least-squares migrated seismic image that is obtained by minimizing the
negative-log likelihood (see the likelihood term in
equation~\ref{bayes_rule_log}). This image, shown in
Figure~\ref{lsrtm_ldc_0}, was constructed by fitting the data without
incorporating any prior information. It is evident from this image that
there are strong artifacts caused by noise in the data, underscoring the
importance of incorporating prior knowledge into solving seismic
imaging.

\paragraph{Improvements in posterior samples and conditional mean
estimate}\label{improvements-in-posterior-samples-and-conditional-mean-estimate}

\begin{figure}
\centering
\captionsetup[subfigure]{labelformat=empty}
\subfloat[]{\includegraphics[width=0.500\hsize]{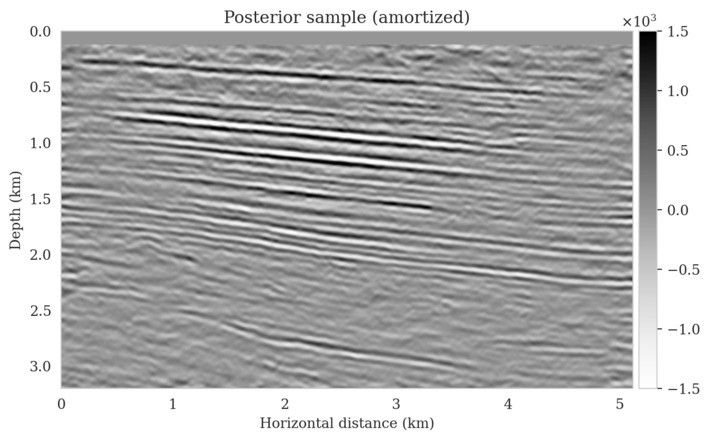}}
\subfloat[]{\includegraphics[width=0.500\hsize]{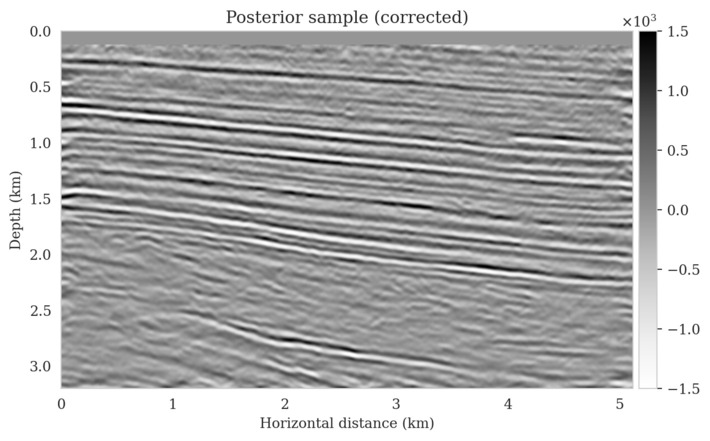}}
\\
\subfloat[]{\includegraphics[width=0.500\hsize]{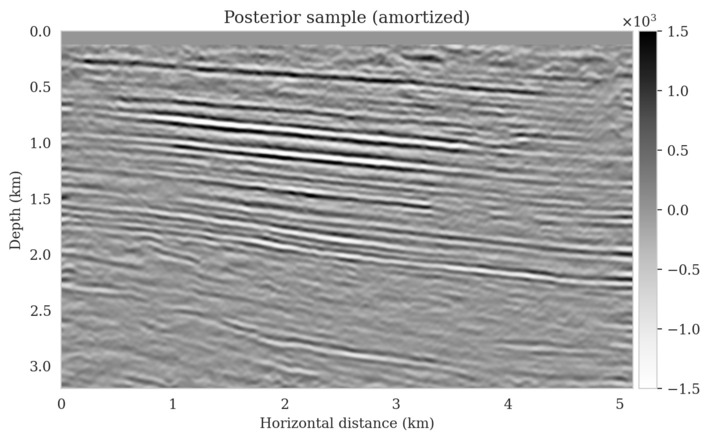}}
\subfloat[]{\includegraphics[width=0.500\hsize]{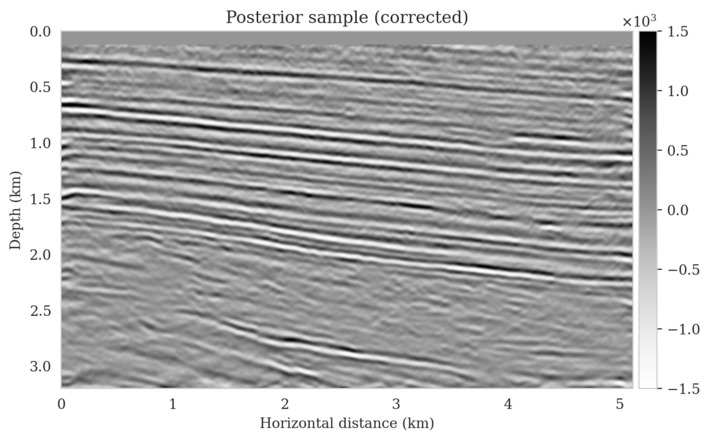}}
\\
\subfloat[]{\includegraphics[width=0.500\hsize]{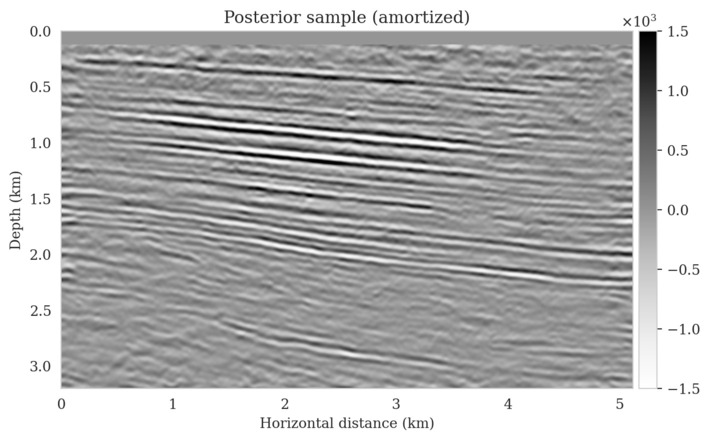}}
\subfloat[]{\includegraphics[width=0.500\hsize]{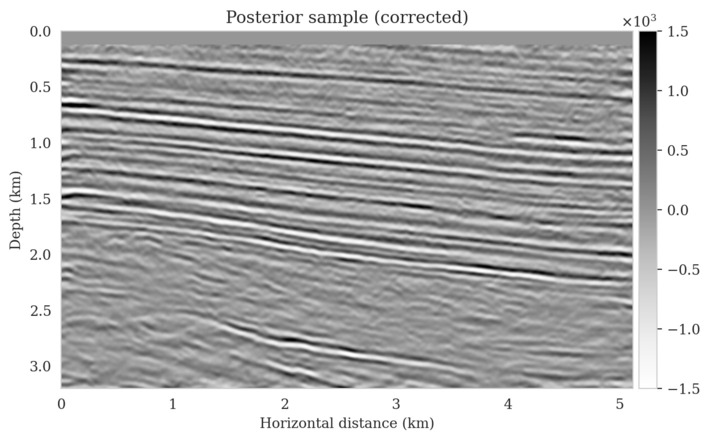}}
\caption{Samples from the posterior distribution (left) without latent
distribution correction with SNRs ranging from $4.57\,\mathrm{dB}$ to
$5.21\,\mathrm{dB}$; and (right) after latent distribution correction
with SNRs ranging from $7.80\,\mathrm{dB}$ to
$8.53\,\mathrm{dB}$.}\label{seismic_ldc_0_samples}
\end{figure}

To obtain amortized (uncorrected) posterior samples, we feed the
reverse-time migrated image (Figure~\ref{rtm_ldc_0}) and latent samples
drawn from the standard Gaussian distribution to the pretrained
normalizing flow. These samples, which are shown in the left column of
Figure~\ref{seismic_ldc_0_samples}, contain artifacts near the top of
the image. These artifacts are related to the near-source reverse-time
migrated image artifacts (Figure~\ref{rtm_ldc_0}). Since the
reverse-time migrated images used during pretraining do not contain
near-source imaging artifacts---due to fine source sampling---the
pretrained normalizing flow fails to eliminate them. Further, the
uncorrected posterior samples do not accurately predict reflectors as
they approach the boundaries and deeper sections of the image.

To illustrate the improved posterior sample quality following latent
distribution correction, we feed latent samples drawn from the corrected
latent distribution to the pretrained normalizing flow (right column of
Figure~\ref{seismic_ldc_0_samples}). Comparing the left and right
columns in Figure~\ref{seismic_ldc_0_samples} indicates an improvement
in the quality of samples from the posterior distribution, which can be
attributed to the attenuation of near-top artifacts and an improvement
in the image quality close to the boundary and deeper reflectors in the
image. Moreover, the SNR values of the posterior samples after
correction are approximately $3\,\mathrm{dB}$ higher, which represents a
significant improvement.

\begin{figure}
\centering
\subfloat[\label{ldc_0_avi_cm}]{\includegraphics[width=0.650\hsize]{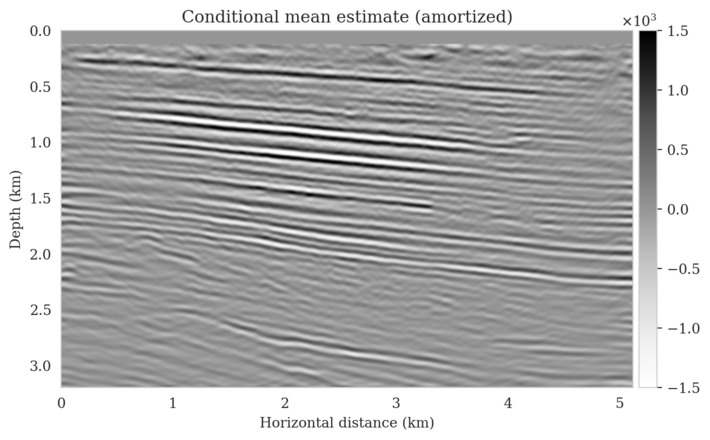}}
\\
\subfloat[\label{ldc_0_cm}]{\includegraphics[width=0.650\hsize]{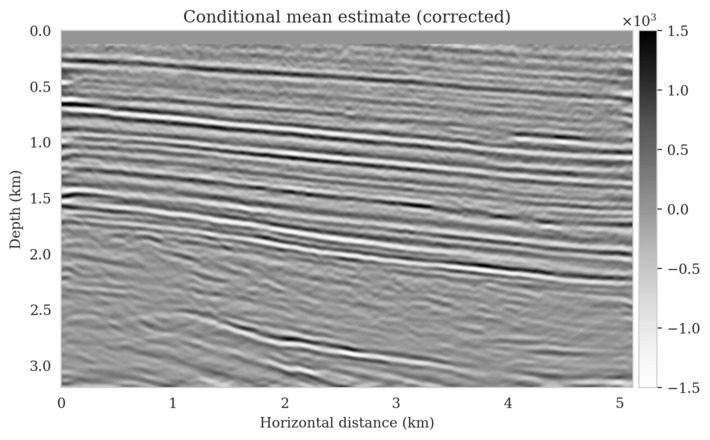}}
\caption{Improvements in conditional mean estimate due to latent
distribution correction. (a) The conditional (posterior) mean estimate
using the pretrained conditional normalizing flow without correction
(SNR $6.29\,\mathrm{dB}$). (b) The conditional mean estimate after
latent distribution correction (SNR
$10.36\,\mathrm{dB}$).}\label{seismic_ldc_0_results_cm}
\end{figure}

To compute the conditional mean estimate, we simulate one thousand
posterior samples before and after latent distribution correction. As
with the posterior samples before correction, drawing samples after
correction is very cheap once the correction is done as it only requires
evaluating the conditional normalizing flow over the corrected latent
samples. Figures~\ref{ldc_0_avi_cm} and~\ref{ldc_0_cm}~show conditional
mean estimates before and after latent distribution correction,
respectively. The conditional mean estimate before correction reveals
similar artifacts as the posterior samples before correction, in
particular, near-top imaging artifacts due to coarse sources sampling
and less illumination of reflectors located closer to the boundary and
deeper portions of the image. The importance of our proposed latent
distribution correction can be observed by juxtaposing the conditional
mean estimate before (Figure~\ref{ldc_0_avi_cm}) and after correction
(Figure~\ref{ldc_0_cm}). The conditional mean estimate obtained after
latent distribution correction eliminates the aforementioned
inaccuracies and enhances the quality of the image by approximately
$4\,\mathrm{dB}$. We gain similar improvements in SNR compared to the
least-squares migrated image (Figure~\ref{lsrtm_ldc_0}) with virtually
the same cost, i.e., five passes over the shot records. This is
significant improvement in SNR also is complimented by access to
information regarding the uncertainty of the image.

\paragraph{Data-space quality control}\label{data-space-quality-control}

\begin{figure*}
\centering
\subfloat[\label{d-pred_avi_ldc_0}]{\includegraphics[width=0.350\hsize]{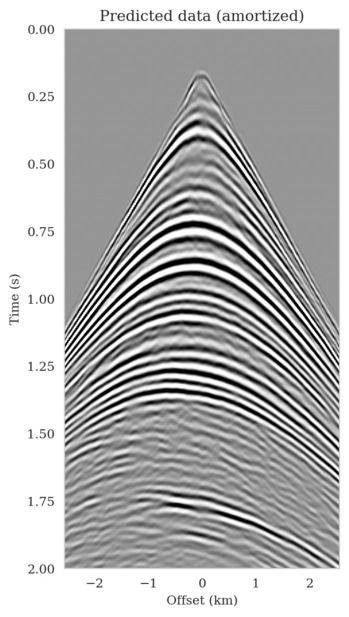}}
\subfloat[\label{d-pred_ldc_0}]{\includegraphics[width=0.350\hsize]{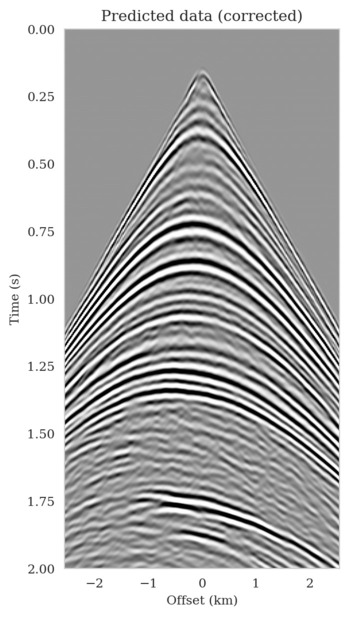}}
\\
\subfloat[\label{d-error_avi_ldc_0}]{\includegraphics[width=0.350\hsize]{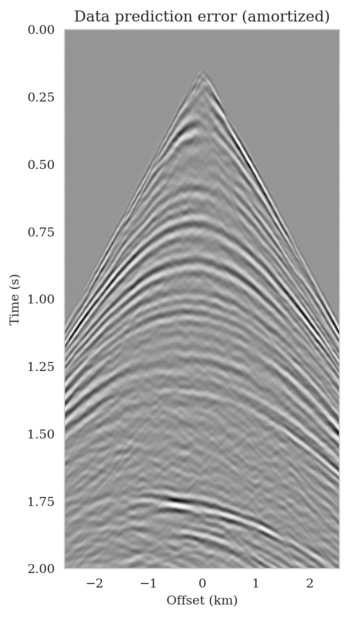}}
\subfloat[\label{d-error_ldc_0}]{\includegraphics[width=0.350\hsize]{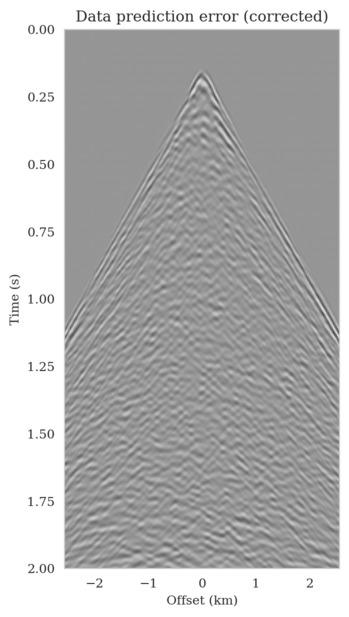}}
\caption{Quality control in data space. Data is simulated by applying
the forward operator to the conditional mean estimate (a) before (SNR
$11.62\,\mathrm{dB}$); and (b) after latent distribution correction (SNR
$16.57\,\mathrm{dB}$). (c) Prediction errors associated with
Figure~\ref{d-pred_avi_ldc_0}. (d) Prediction errors associated with
Figure~\ref{d-pred_ldc_0} (after latent distribution
correction).}\label{data_QC_ldc_0}
\end{figure*}

As the latent distribution correction step involves finding latent
samples that are better suited to fit the data
(equation~\ref{reverse_kl_covariance_diagonal}), we can expect an
improvement in fitting the observed data after correction. Predicted
data is obtained by applying the forward operator to the conditional
mean estimates, before and after latent distribution correction.
Figures~\ref{d-pred_avi_ldc_0} and~\ref{d-pred_ldc_0} show the predicted
shot records before and after correction, respectively. In spite of the
fact that both predicted data appear to be similar to ideal noise-free
data (Figure~\ref{d-noise-free_ldc_0}), the data residual associated
with the conditional mean without correction reveals several coherent
events that contain valuable information about the unknown seismic
image. The latent distribution correction allows us to fit these
coherent events as indicated by the data residual associated with the
corrected conditional mean estimate.

\paragraph{Uncertainty quantification---pointwise standard deviation and
histograms}\label{uncertainty-quantificationpointwise-standard-deviation-and-histograms}

\begin{figure}
\centering
\subfloat[\label{ldc_0_std}]{\includegraphics[width=0.650\hsize]{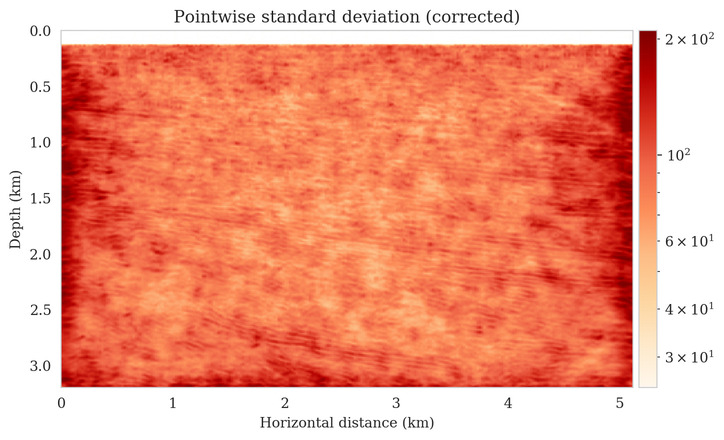}}
\\
\subfloat[\label{ldc_0_std_normalized}]{\includegraphics[width=0.650\hsize]{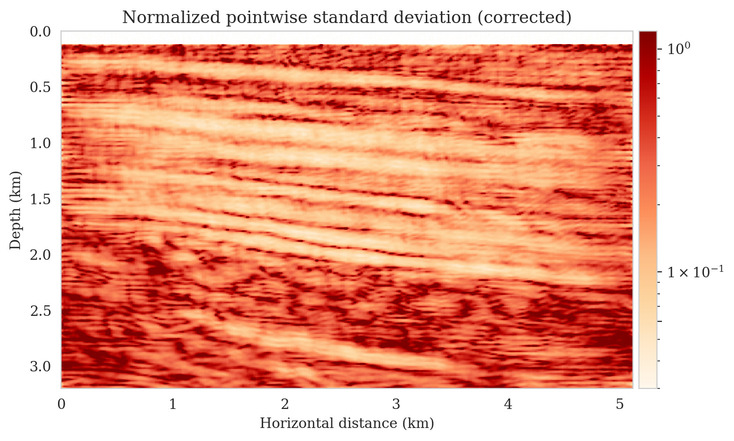}}
\caption{Uncertainty quantification with latent distribution correction.
(a) The pointwise standard deviation among samples drawn from the
posterior after latent distribution correction. (b) Normalized pointwise
standard deviation by the conditional mean estimate
(Figure~\ref{ldc_0_cm}).}\label{seismic_ldc_0_results_std}
\end{figure}

\begin{figure}
\centering
\subfloat[\label{avi_ldc_0_std_profile_64}]{\includegraphics[width=0.650\hsize]{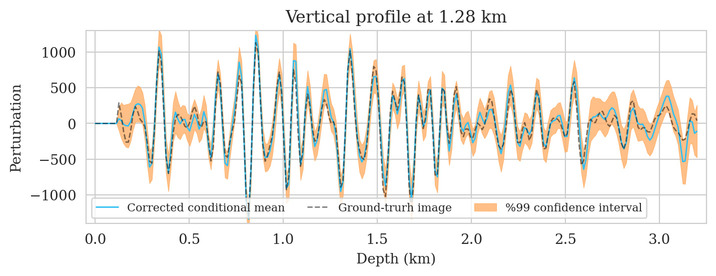}}
\\
\subfloat[\label{avi_ldc_0_std_profile_128}]{\includegraphics[width=0.650\hsize]{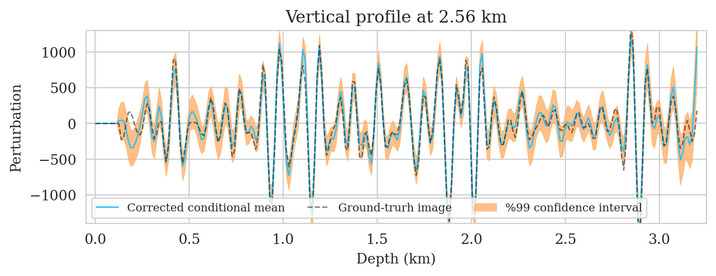}}
\\
\subfloat[\label{avi_ldc_0_std_profile_192}]{\includegraphics[width=0.650\hsize]{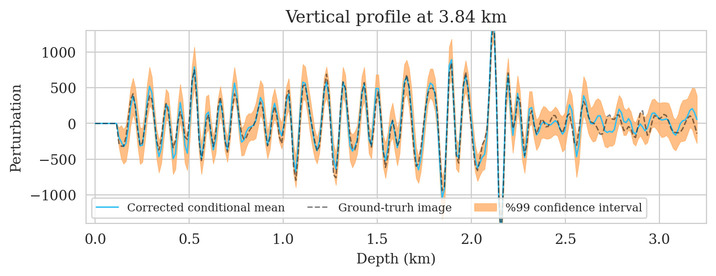}}
\caption{Confidence intervals for three vertical profiles. Traces of
$99 \%$ confidence interval (shaded orange color), corrected conditional
mean (solid blue), and ground truth (dashed black) at (a)
$1.28\,\mathrm{km}$, (b) $2.56\, \mathrm{km}$, and (c)
$3.84\,\mathrm{km}$~horizontal
location.}\label{seismic_ldc_0_results_profile}
\end{figure}

We exploit cheap access to corrected samples from the posterior in order
to extract information regarding uncertainty in the image estimates.
Figure~\ref{ldc_0_std} displays the pointwise standard deviation among
the one thousand corrected posterior samples. The overprint by the
strong reflectors can be reduced by normalizing the standard deviation
using a stabilized division by the conditional mean
(Figure~\ref{ldc_0_std_normalized}). The pointwise standard deviation
plots indicate high uncertainty in areas near the boundaries of the
image and in the deep parts of the image where illumination is
relatively poor. This observation is more evident in
Figure~\ref{seismic_ldc_0_results_profile}, which displays three
vertical profiles as $99\%$ confidence intervals (orange colored
shading) illustrating the expected increasing trend of uncertainty with
depth. We additionally observe that the ground truth (dashed black)
falls within the confidence intervals for most of the areas.

\begin{figure}
\centering
\subfloat[\label{avi_ldc_0_std_hist_60-70}]{\includegraphics[width=0.650\hsize]{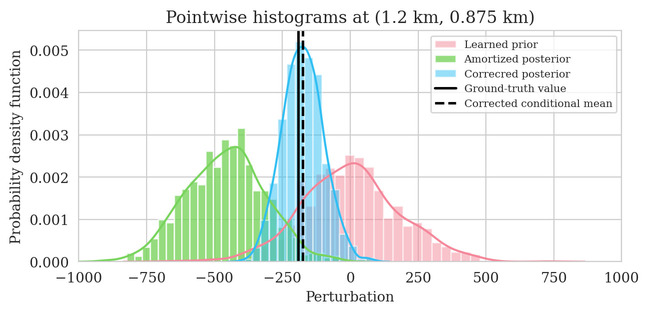}}
\\
\subfloat[\label{avi_ldc_0_std_hist_70-200}]{\includegraphics[width=0.650\hsize]{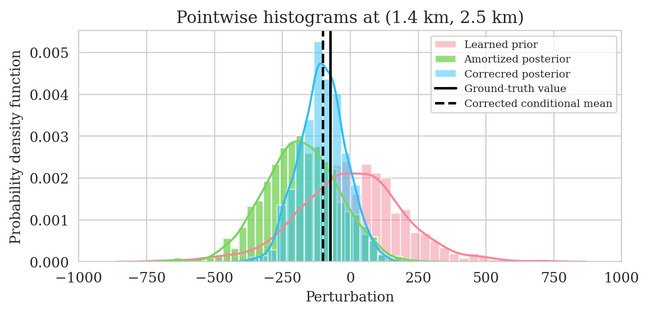}}
\\
\subfloat[\label{avi_ldc_0_std_hist_200-150}]{\includegraphics[width=0.650\hsize]{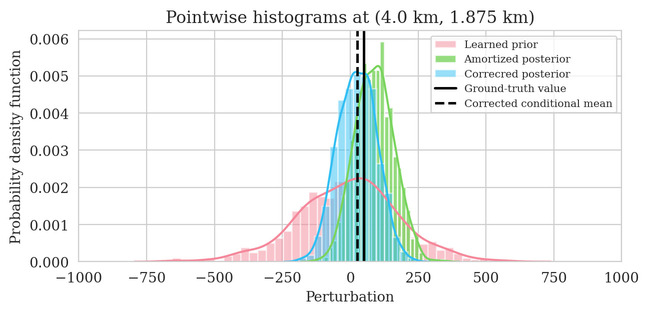}}
\caption{Pointwise prior (red), uncorrected amortized posterior (green),
and latent distribution corrected posterior (blue) histograms along with
the true perturbation values (solid black line) and the corrected
conditional mean (dashed black line) for points located at (a)
$(1.2\, \mathrm{km}, \ 0.875\, \mathrm{km})$, (b)
$(1.4\, \mathrm{km}, \ 2.5\, \mathrm{km})$, and (c)
$(4.0\, \mathrm{km},\
1.875\, \mathrm{km})$.}\label{seismic_ldc_0_results_hist}
\end{figure}

To demonstrate how the corrected posterior is informed by the observed
data, we calculated histograms at three locations in
Figure~\ref{ldc_0_std}. Prior histograms are calculated by feeding
latent samples drawn from the standard Gaussian distribution to the
pretrained conditional normalizing flow without using data conditioning
(see \citet{kruse2021hint} and \citet{siahkoohi2021Seglbe} for more
information). These samples in the image spaces are indicative of
samples from the prior distribution implicitly learned by the
conditional normalizing flow during pretraining. The resulting prior
histograms are shown in Figure~\ref{seismic_ldc_0_results_hist}.
Corresponding histograms are also obtained for the uncorrected amortized
posterior distribution (equation~\ref{sampling_amortized}). As mentioned
before, the uncorrected posterior distribution serves as an implicit
conditional prior for the subsequent step of correction of the latent
distribution. The green histograms in
Figure~\ref{seismic_ldc_0_results_hist} represent the uncorrected
amortized posterior distribution. A similar procedure is followed to
obtain histograms after latent distribution correction (blue
histograms). As expected, the histograms of the posterior distribution
are considerably narrower than those of the learned prior, which
indicates that the posterior is further informed by the specific
observed dataset and physics. As a means of evaluating the effect of
latent distribution correction, we provide a vertical solid line showing
the ground truth value's location. All three corrected posterior
histograms for each location are shifted towards the ground truth, and
their (conditional) mean plotted with the dashed vertical line indicates
improved recovery of the ground truth. Compared to the amortized
uncorrected histograms, the corrected histograms in
Figures~\ref{avi_ldc_0_std_hist_60-70}
--~\ref{avi_ldc_0_std_hist_70-200} are further contracted, suggesting
that the latent distribution correction step has further informed the
inference by the data.

\subsubsection{Bayesian inference
verification}\label{bayesian-inference-verification}

While we investigated the accuracy of the conditional mean estimate
after correction, we do not have access to the underlying true posterior
distribution to verify our proposed posterior sampling method. This is
partly due to our learned prior and the implicit conditional prior used
in latent variable correction, which make traditional MCMC-based
comparisons challenging. To further validate our Bayesian inference
procedure, we conduct a series of experiments in which we investigate
the effect of gradual increase in the number of sources ($N$) and
reduction of the noise level. As the number of sources increases and the
noise level decreases, we expect to see an increase in seismic image
quality and a decrease in uncertainty.

\paragraph{Estimation accuracy}\label{estimation-accuracy}

\begin{figure}
\centering
\includegraphics[width=0.500\hsize]{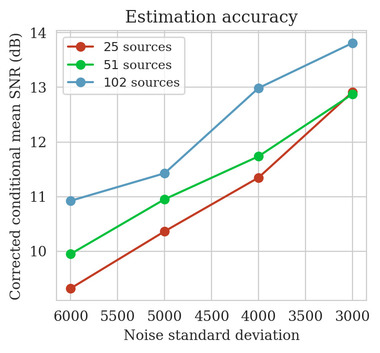}
\caption{Estimation accuracy as a function of number of sources and
noise levels. Colors correspond to different source
numbers.}\label{avi_ldc_post_accuracy_snr}
\end{figure}

The accuracy of the Bayesian parameter estimation method is directly
affected by the amount of data that has been collected
\citep{tarantola2005inverse}. That is to say with more observed data
(larger $N$), we should be able to obtain a more accurate seismic image
estimate. The same principle allows us to stack out noise when
increasing the fold in seismic data acquisition. In order to assess
whether our Bayesian inference approach has this property, we repeat the
latent distribution correction process while varying the number of
sources (using $N = 25, 51, 102$ sources) and the amount of band-limited
noise (standard deviations $1.5, 2.0, 2.5$, and $3.0$ times greater than
the noise standard deviation during pretraining). Each of the $12$
instances of latent distribution correction problems is treated
similarly with respect to the number of passes made over the shot
records and other optimization parameters. For each of the $12$
combinations of source numbers and noise levels, we calculate the
corrected conditional mean estimate and plot the SNRs as a function of
the noise's standard deviation in
Figure~\ref{avi_ldc_post_accuracy_snr}.

For a fixed number of sources ($25$, $51$, and $102$ sources shown with
red, green, and blue colors respectively), we plot the corrected
conditional means SNR as a function of the noise standard deviation.
There is a clear increase in SNR trend as we decrease the noise level.
In the same way, for each fixed noise level, the SNR increases with the
number of sources. This verifies the our Bayesian inference method
yields a more accurate estimate of the conditional mean for larger
number of sources and smaller noise levels.

\paragraph{Bayesian posterior
contraction}\label{bayesian-posterior-contraction}

\begin{figure}
\centering
\includegraphics[width=1.000\hsize]{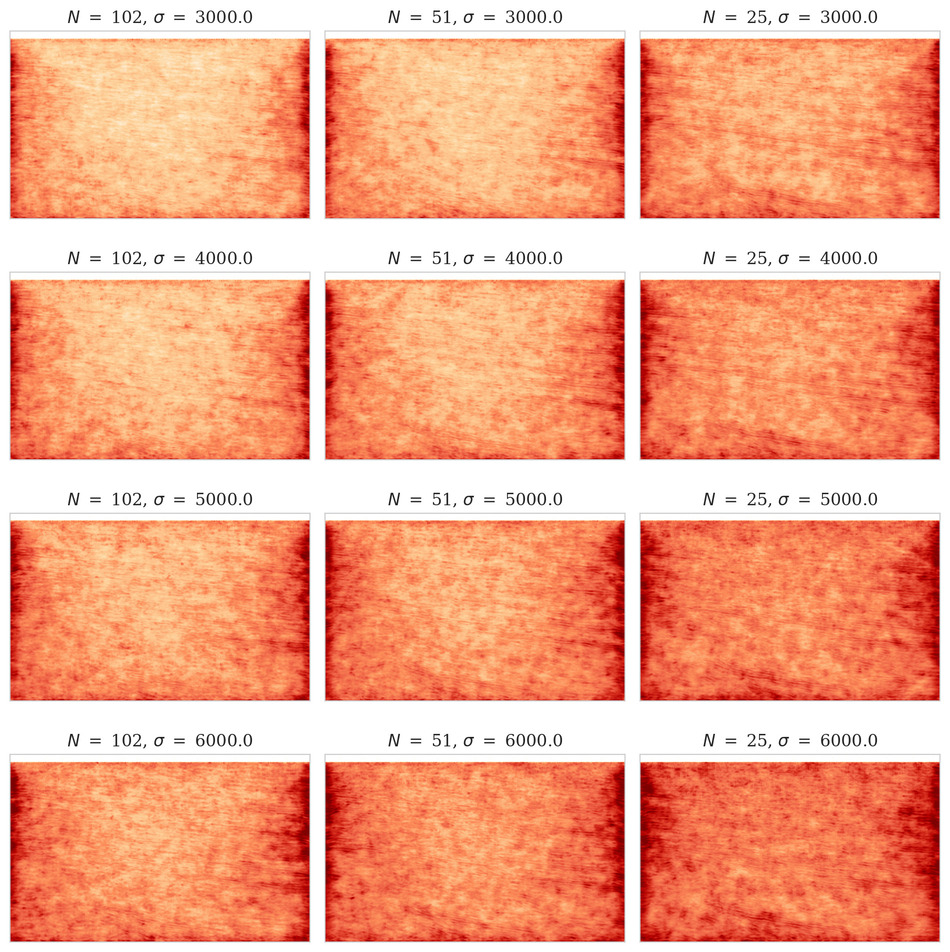}
\caption{Bayesian posterior contraction: visual inspection. Pointwise
standard deviations for varying number of sources (decreasing left to
right) and noise variances (increases top to bottom).}\label{all_stds}
\end{figure}

An alternative Bayesian inference verification method involves analyzing
the Bayesian posterior contraction, that is, the decrease of uncertainty
with more data. To examine whether or not our Bayesian inference method
possesses this property, we visually inspect the resulting pointwise
standard deviation plots in Figure~\ref{all_stds} for the $12$ possible
combinations of source numbers and noise levels. Each row corresponds to
the pointwise standard deviation plot for a fixed noise standard
deviation ($\sigma$), where the number of sources ($N$) decreases from
left to right. In each column, we maintain the number of sources and we
plot the pointwise standard deviation as we increase the noise standard
deviation from top to bottom. There is a consistent increase in standard
deviation values as we move from the top-left to the bottom-right
corner. In other words, the posterior contract (shrinks) when we have
more data (more numbers of sources, Figure~\ref{all_stds} from right to
left) and when we have less noise (Figure~\ref{all_stds} from bottom to
top), which effectively means more data.

\begin{figure}
\centering
\subfloat[\label{avi_ldc_0_nsrc25}]{\includegraphics[width=0.333\hsize]{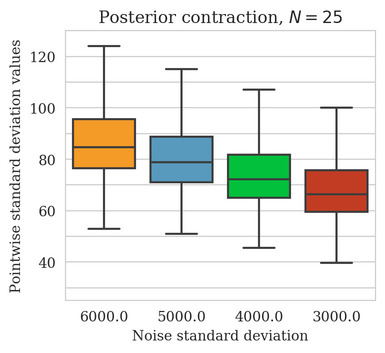}}
\subfloat[\label{avi_ldc_0_nsrc51}]{\includegraphics[width=0.333\hsize]{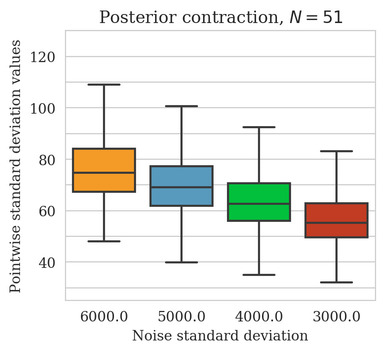}}
\subfloat[\label{avi_ldc_0_std_nsrc102}]{\includegraphics[width=0.333\hsize]{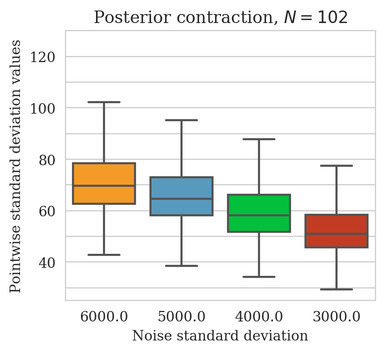}}
\caption{Box plots of pointwise standard deviation values as a function
of noise level for number of sources (a) $N=25$, (b) $N=51$, and (c)
$N=102$.}\label{all_stds_box}
\end{figure}

Figure~\ref{all_stds_box} offers an alternate method of visualizing
posterior contraction, displaying box plots of the standard deviation
values for each of the $12$ images in Figure~\ref{all_stds}. In each of
the three box plots, the vertical axis corresponds to the noise standard
deviation, and the horizontal axis represents the possible values in the
posterior pointwise standard deviation plots. The box indicates the
values that are between the first and third quartiles (where half of the
possible values fall) and the line in the middle indicates the median
value. Figures~\ref{avi_ldc_0_nsrc25} to~\ref{avi_ldc_0_std_nsrc102}
show box plots for experiments with $25$, $51$, and $102$ sources,
respectively, with each box plot color reflecting a particular noise
level. In each of the Figures~\ref{avi_ldc_0_nsrc25}
to~\ref{avi_ldc_0_std_nsrc102}, we observe a decrease in the range of
posterior standard deviation values, including median and quantiles, as
we lower the noise level from left to right. Similarly, for the same
noise levels, that is, box plots of the same color, the standard
deviations decrease from Figure~\ref{avi_ldc_0_nsrc25} to
Figure~\ref{avi_ldc_0_std_nsrc102} (increasing number of sources). The
observed trends in Figures~\ref{all_stds} and~\ref{all_stds_box} verify
that our Bayesian inference method exhibits the Bayesian posterior
contraction property.

\subsection{Shift in the forward model and prior
distribution}\label{shift-in-the-forward-model-and-prior-distribution}

This example is intended to demonstrate how our latent distribution
correction method can perform Bayesian inference when the unknown ground
truth image has properties that differ from the seismic images in the
pretraining dataset. This expectation is based on the invertible nature
of conditional normalizing flows, which allows them to represent any
image in the image space \citep{pmlr-v119-asim20a}.

\begin{figure}
\centering
\subfloat[\label{true_model_ldc_1}]{\includegraphics[width=0.500\hsize]{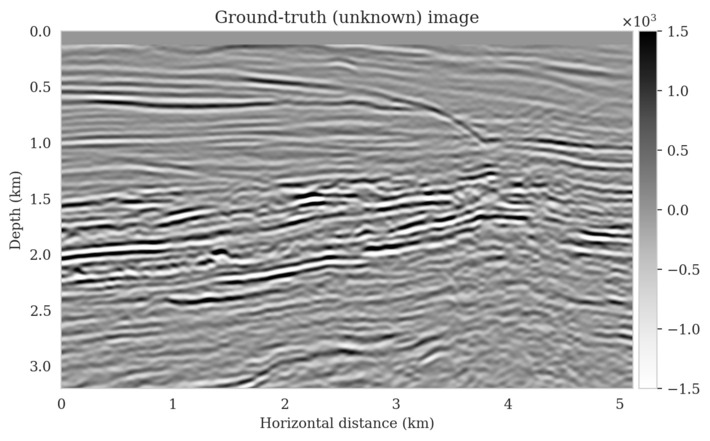}}
\subfloat[\label{true_model_ldc_2}]{\includegraphics[width=0.500\hsize]{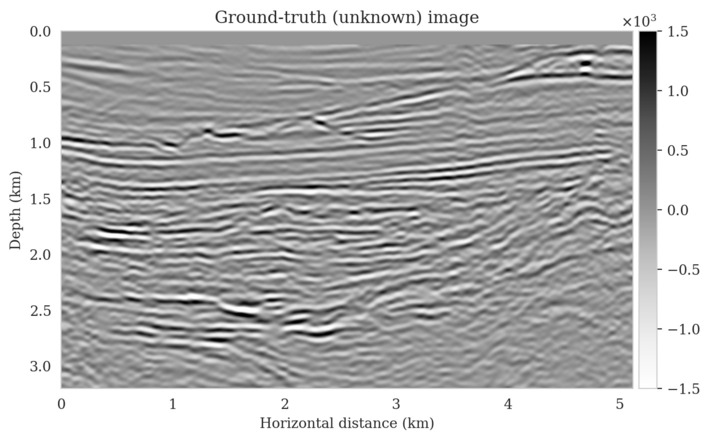}}
\\
\subfloat[\label{rtm_ldc_1}]{\includegraphics[width=0.500\hsize]{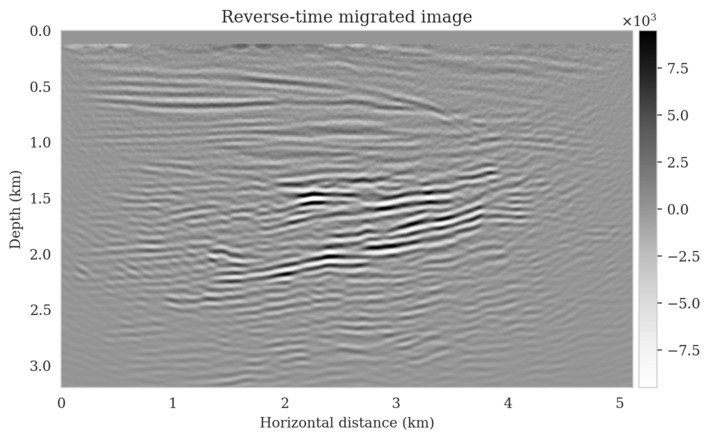}}
\subfloat[\label{rtm_ldc_2}]{\includegraphics[width=0.500\hsize]{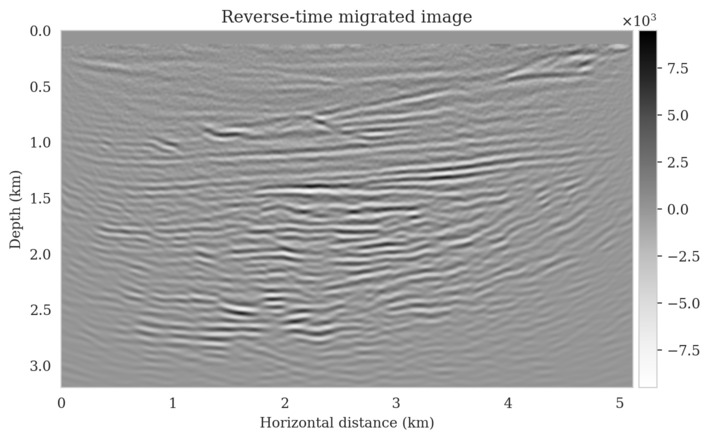}}
\\
\subfloat[\label{lsrtm_ldc_1}]{\includegraphics[width=0.500\hsize]{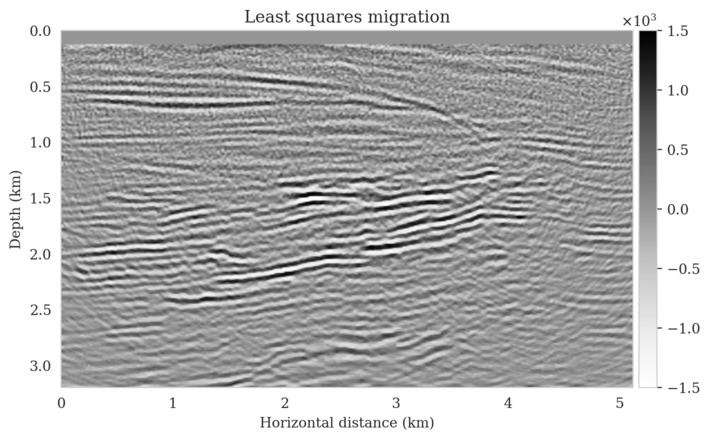}}
\subfloat[\label{lsrtm_ldc_2}]{\includegraphics[width=0.500\hsize]{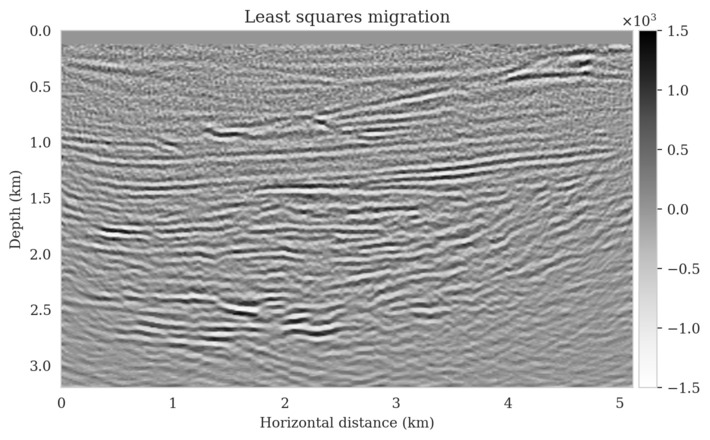}}
\caption{Setup for experiments involving shifts in the prior
distribution. (a) and (b) Two high-fidelity ground-truth images from
deeper sections of the Parihaka dataset. (c) and (d) Reverse-time
migrated image corresponding ground-truth images in
Figures~\ref{true_model_ldc_1} (SNR $-8.02\,\mathrm{dB}$)
and~\ref{true_model_ldc_2} (SNR $-8.49\,\mathrm{dB}$), respectively. (e)
and (f) Least squares imaging results (no regularization) corresponding
ground-truth images in Figures~\ref{true_model_ldc_1} (SNR
$4.94\,\mathrm{dB}$) and~\ref{true_model_ldc_2} (SNR
$5.59\,\mathrm{dB}$), respectively.}\label{seismic_ldc_12_setup}
\end{figure}

As a means to mimic this scenario, unlike images used in pretraining, we
have extracted two 2D seismic images from deeper sections of the
Parihaka dataset. In these sections, there are fewer continuous
reflectors and more complex geological features.
Figures~\ref{true_model_ldc_1} and~\ref{true_model_ldc_2} show two
images with drastically different geological features when compared to
Figure~\ref{true_model}. Following the pretraining data acquisition
setup, we add a water column on top of these images to reduce the
near-source artifacts. Similar to the previous experiment involving
forward model shifts, we use only 25 sources with 200 meters sampling
distance and add noise with a $2.5$ fold larger standard deviation than
during the pretraining phase. With this setup, the noisy observed data
associated with experiments involving seismic images in
Figures~\ref{true_model_ldc_1} and~\ref{true_model_ldc_2} have a SNR of
$-1.56\,\mathrm{dB}$ and $-2.41\,\mathrm{dB}$, respectively.

As part of our analysis, we compute the reverse-time (second row in
Figure~\ref{seismic_ldc_12_setup}) and least-squares (third row in
Figure~\ref{seismic_ldc_12_setup}) migrated images for these two
ground-truth images, where the former images serve as inputs to the
pretrained conditional normalizing flows. Similar to the previous
example, the reverse-time migrated images contain the near-source
artifacts and are contaminated by the input noise. Moreover, the
least-squares migrated images highlight the importance of including
prior information in this imaging problem, since this image contains
strong noise-related artifacts, which might impact downstream tasks,
such as horizon tracking.

\subsubsection{Conditional mean and pointwise standard
deviations}\label{conditional-mean-and-pointwise-standard-deviations}

\begin{figure}
\centering
\subfloat[\label{ldc_1_avi_cm}]{\includegraphics[width=0.500\hsize]{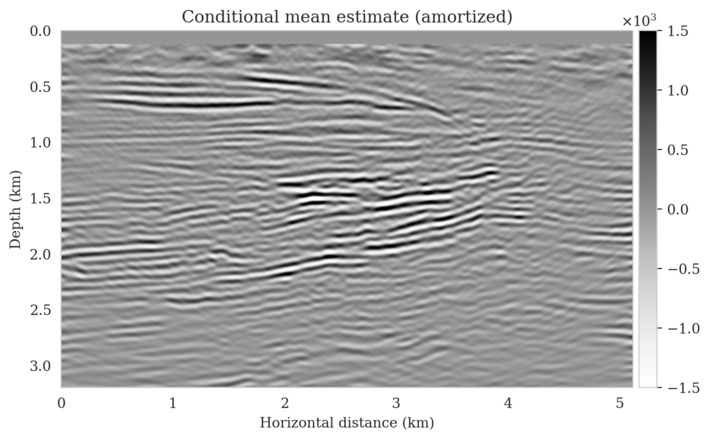}}
\subfloat[\label{ldc_2_avi_cm}]{\includegraphics[width=0.500\hsize]{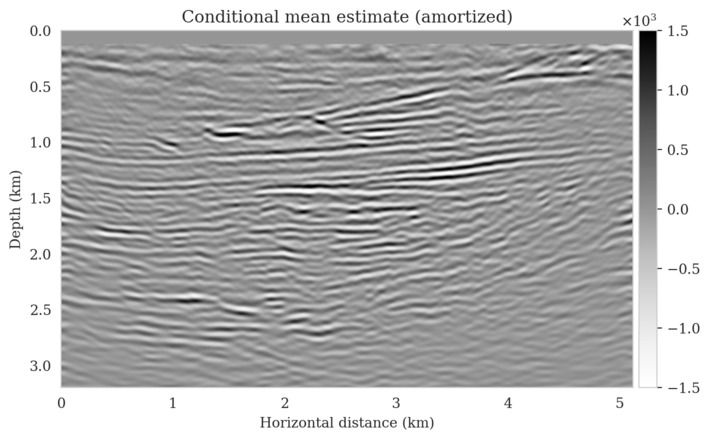}}
\\
\subfloat[\label{ldc_1_cm}]{\includegraphics[width=0.500\hsize]{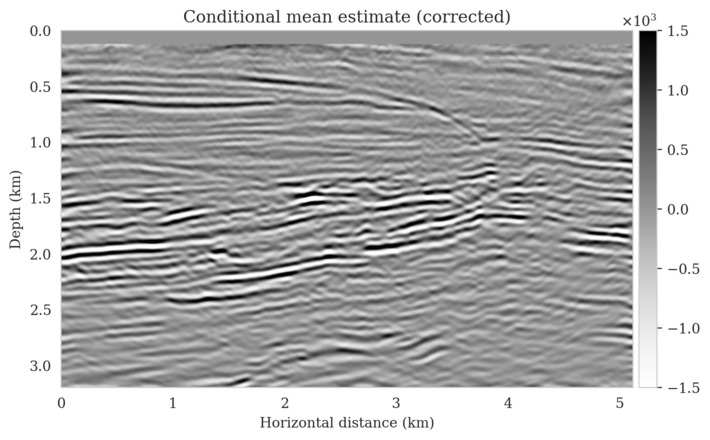}}
\subfloat[\label{ldc_2_cm}]{\includegraphics[width=0.500\hsize]{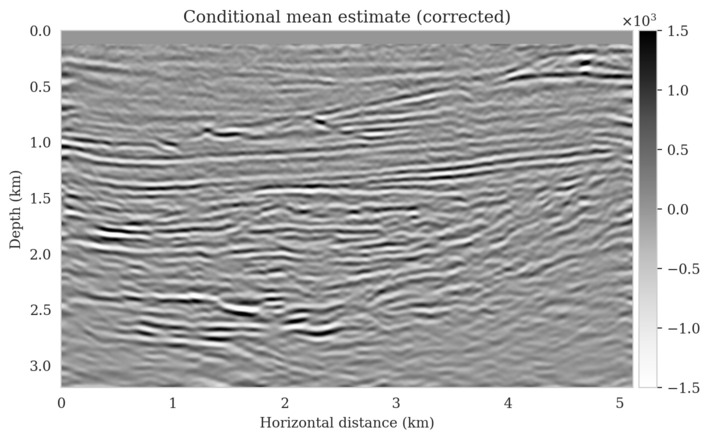}}
\caption{The improvement in conditional mean estimate due to latent
distribution correction. (a) and (b) Amortized uncorrected conditional
(posterior) mean estimates with SNRs $5.47\,\mathrm{dB}$ and
$6.17\,\mathrm{dB}$, respectively. (c) and (d) Conditional (posterior)
mean estimates after latent distribution correction with SNRs
$9.40\,\mathrm{dB}$ and $9.11\,\mathrm{dB}$,
respectively.}\label{seismic_ldc_12_results_cm}
\end{figure}

To obtain samples from the posterior distribution, we feed the
reverse-time migrated images (Figures~\ref{rtm_ldc_1}
and~\ref{rtm_ldc_2}) into the pretrained conditional normalizing flow.
The posterior is sampled using either standard Gaussian distributions or
corrected latent samples. It is apparent, once more, that the
uncorrected conditional mean estimates are significantly contaminated by
artifacts in the near-source region (Figures~\ref{ldc_1_avi_cm}
and~\ref{ldc_2_avi_cm}). Another type of noticeable error in these
predicted images includes lower amplitudes in deeper and closer to
boundary reflectors. A comparison of the conditional means estimates
before (Figures~\ref{ldc_1_avi_cm} and~\ref{ldc_2_avi_cm}) and after
(Figures~\ref{ldc_1_cm} and~\ref{ldc_2_cm}) latent distribution
correction indicates attenuation of near-source artifacts as well as an
improvement in reflector illumination near the boundary and deeper
sections where images are more difficult to capture. Following latent
distribution correction, the conditional mean estimate SNR is improved
by three to four decibels for both images. To provide more quantitative
results, we ran the experiments as set up in this section for eight
additional seismic images sampled from deeper sections of the Parihaka
dataset, sampled from deeper sections of the Parihaka dataset. The SNR
of the estimated seismic images before and after latent distribution are
$5.91 \pm 0.49\,\mathrm{dB}$ and $9.15 \pm 0.73\,\mathrm{dB}$,
respectively.

\begin{figure}
\centering
\subfloat[\label{ldc_1_std}]{\includegraphics[width=0.500\hsize]{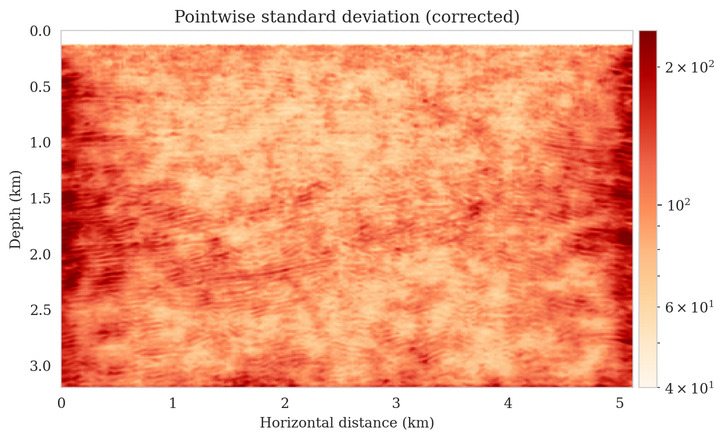}}
\subfloat[\label{ldc_2_std}]{\includegraphics[width=0.500\hsize]{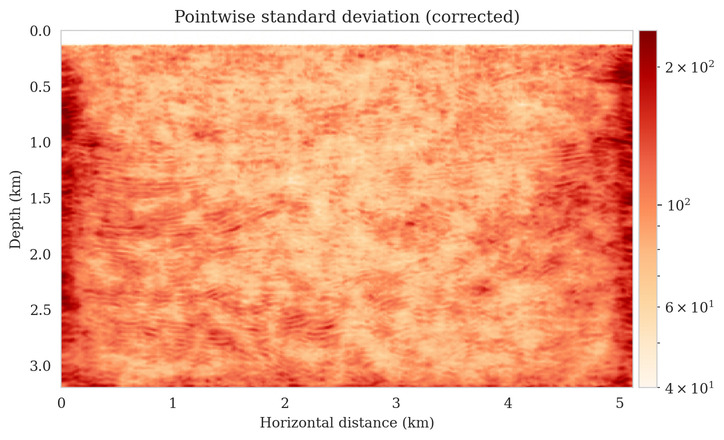}}
\\
\subfloat[\label{ldc_1_std_normalized}]{\includegraphics[width=0.500\hsize]{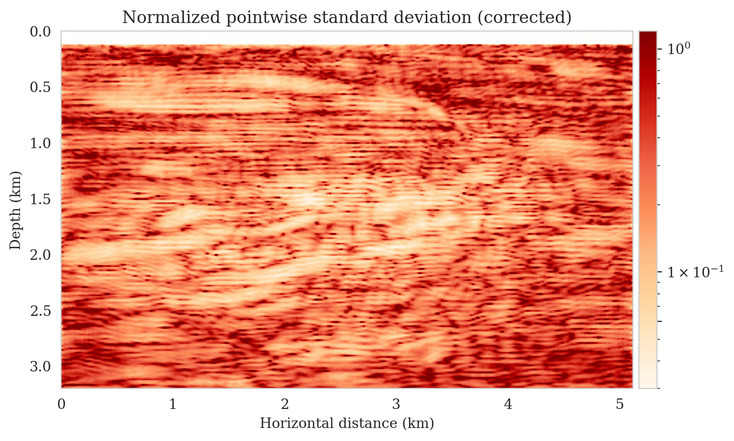}}
\subfloat[\label{ldc_2_std_normalized}]{\includegraphics[width=0.500\hsize]{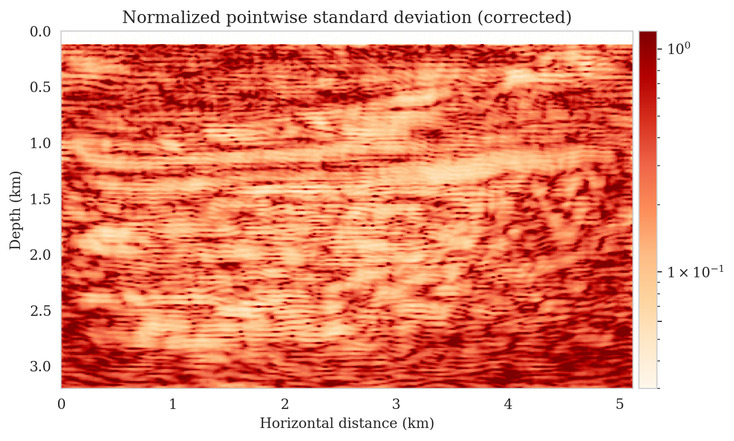}}
\caption{Uncertainty quantification with latent distribution correction.
(a) and (b) The pointwise standard deviation estimates among samples
drawn from the posterior after latent distribution correction. (b)
Pointwise standard deviations normalized by the envelop of conditional
mean estimates.}\label{seismic_ldc_12_results_std}
\end{figure}

Careful inspection at the boundaries in the corrected conditional mean
estimates reveals some nonrealistic events near the boundaries. The
plots of pointwise standard deviations (Figures~\ref{ldc_1_std}
and~\ref{ldc_2_std}) associated with corrected conditional means,
however, clearly indicate that there is uncertainty for these events.
This illustrates the importance of uncertainty quantification and not
relying on a single estimate when addressing ill-posed inverse problems.
To diminish the imprint of strong reflectors in the pointwise standard
deviations plots, we also display these images when normalized with
respect to the envelopes of the conditional mean estimates in
Figures~\ref{ldc_1_std_normalized} and~\ref{ldc_2_std_normalized}.

\subsubsection{Data residuals}\label{data-residuals}

\begin{figure*}
\centering
\subfloat[\label{d-pred_avi_ldc_1}]{\includegraphics[width=0.250\hsize]{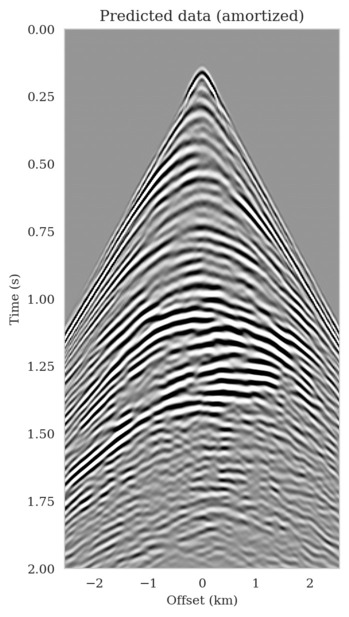}}
\subfloat[\label{d-pred_ldc_1}]{\includegraphics[width=0.250\hsize]{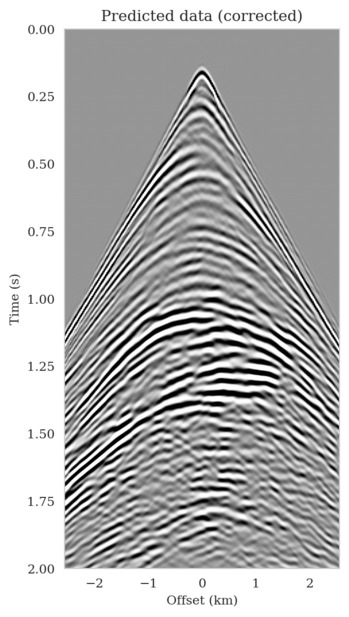}}
\subfloat[\label{d-error_avi_ldc_1}]{\includegraphics[width=0.250\hsize]{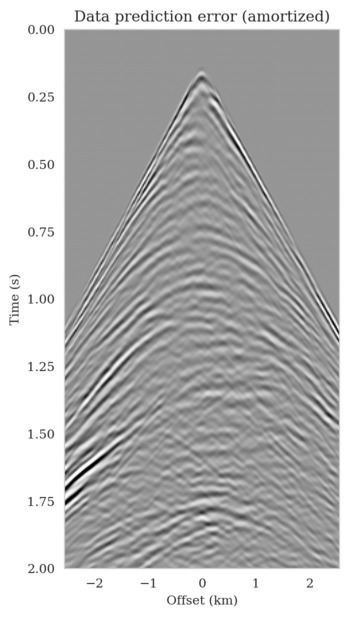}}
\subfloat[\label{d-error_ldc_1}]{\includegraphics[width=0.250\hsize]{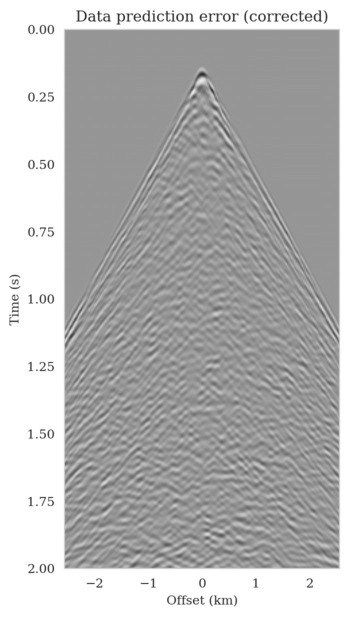}}
\\
\subfloat[\label{d-pred_avi_ldc_2}]{\includegraphics[width=0.250\hsize]{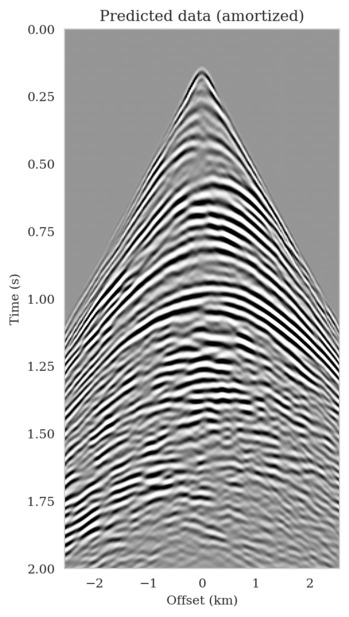}}
\subfloat[\label{d-pred_ldc_2}]{\includegraphics[width=0.250\hsize]{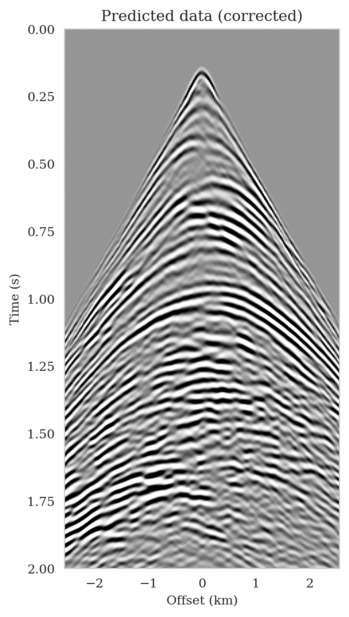}}
\subfloat[\label{d-error_avi_ldc_2}]{\includegraphics[width=0.250\hsize]{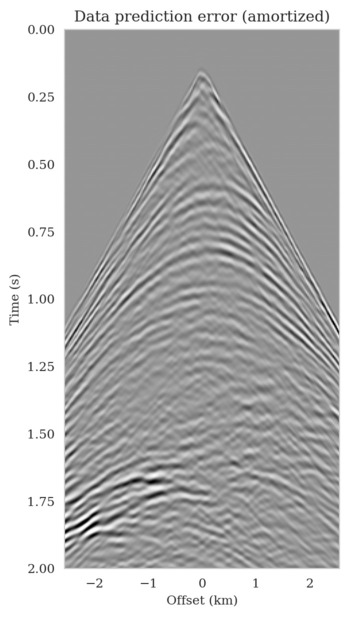}}
\subfloat[\label{d-error_ldc_2}]{\includegraphics[width=0.250\hsize]{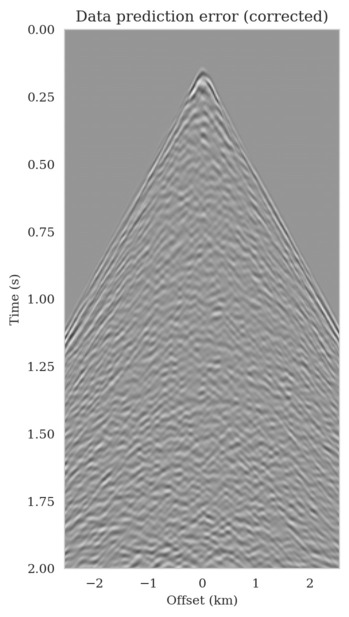}}
\caption{Quality control in data space. The first and second row
correspond to experiments involving Figures~\ref{true_model_ldc_1}
and~\ref{true_model_ldc_2}, respectively. (a) and (e) Predicted data
before correction with SNRs $9.13\,\mathrm{dB}$ and $9.02\,\mathrm{dB}$,
respectively. (b) and (f) Predicted data after latent distribution
correction with SNRs $15.27\,\mathrm{dB}$ and $14.52\,\mathrm{dB}$,
respectively. (c) and (g) Data residuals (before correction) associated
with Figures~\ref{d-pred_avi_ldc_1} and~\ref{d-pred_avi_ldc_2},
respectively. (d) and (h) Data residuals (after correction) associated
with Figures~\ref{d-pred_ldc_1} and~\ref{d-pred_ldc_2},
respectively.}\label{data_QC_ldc_12}
\end{figure*}

As before, we confirm that the latent distribution correction improves
the fit of the data. Figure~\ref{data_QC_ldc_12} shows the predicted
data as well as the data residuals for all conditional mean estimates.
The predicted data are obtained by applying the forward operator to the
conditional mean estimates, both before and after latent distribution
correction. The predicted data before and after correction are presented
in the first and second columns, respectively. The corresponding data
residuals before and after correction, which are computed by subtracting
the predicted data from ideal noise-free data, can be seen in the third
and last columns of Figure~\ref{data_QC_ldc_12}. Evidently, the latent
distribution correction stage has resulted in a better fit with the
observed data, as coherent data events show up in
Figures~\ref{d-error_avi_ldc_1} and~\ref{d-error_avi_ldc_2}, but are
attenuated in the corrected residual plots (Figures~\ref{d-error_ldc_1}
and~\ref{d-error_ldc_2}).

\section{Discussion}\label{discussion}

The examples presented demonstrate that deep neural networks trained in
the context of amortized variational inference can facilitate solving
inverse problems in two ways: (1) incorporating prior knowledge gained
through pretraining; and (2) accelerating Bayesian inference and
uncertainty quantification. Despite the fact that amortized variational
inference is capable of sampling the posterior distribution without
requiring forward operator evaluations during inference, the extent to
which it is considered reliable is dependent upon the availability of
high quality training data. We demonstrate this limitation of amortized
variational inference via a seismic imaging example where we alter the
number of sources, variance of noise, and the present geological
features to be imaged in comparison to the pretraining phase. These
variations shift the data distribution, and in certain cases, e.g.,
imaging slightly different geological features, this can be also thought
as evaluating the pretrained conditional normalizing flow over data
samples from the low probability regions of the training distribution.
Due to shifts in the data distribution, the obtained posterior samples
via amortized variational inference in our numerical experiments had
unusual near-source artifacts and less illuminated reflectors, which we
can be partly attributed to lack of these artifacts in the training
reverse-time migrated images.

As part of our efforts to extend the application of the supervised
amortized variational inference methods to domains with limited access
to high-quality training data, we developed an unsupervised,
physics-based variational inference formulation over the latent space of
a pretrained conditional normalizing flow that mitigates some of the
posterior sampling errors induced by data distribution shifts. In this
approach, a diagonal correction to the latent distribution is learned
that ensures that the conditional normalizing flow output distribution
better matches the desired posterior distribution. Based on observations
and other research
\citep{pmlr-v119-asim20a, orozco2021photoacoustic, siahkoohi2022EAGEweb},
we found that normalizing flows, because of their invertibility, are
capable of partially mitigating errors related to changes in data
distributions when they are used to solve inverse problems. We leave the
delineation of which data distribution shifts can be handled by our
diagonal correction approach to future work. Needless to say, by
adhering to more complex transformations in the latent space, e.g., via
a neural network, a wide range of data distribution shifts can be
handled but it would potentially require a more computationally costly
correction step.

Latent distribution correction requires several passes over the data
(five in our example), which includes evaluation of the forward operator
and the adjoint of its Jacobian. Due to the amortized nature of our
approach, these costs are significantly lower than those incurred by
other non-amortized variational inference methods
\citep{rizzuti2020SEGpub, andrle2021invertible, zhang2021bayesian2, zhao2022interrogating},
specifically tens of thousands of forward operator evaluations reported
by \citet{zhao2021bayesian} in the context of travel-time tomography. By
amortizing over the data, the pretrained conditional normalizing flow
provides posterior samples for previously unseen data (drawn from the
same distribution as training data) without the need for latent
distribution correction. In presence of moderate data distribution
shifts, the learned posterior is adjusted to new observed data via the
latent distribution correction step, which could be considered an
instance of transfer learning \citep{yosinski2014transferable}. In
contrast, existing methods that perform variational inference in the
latent space \citep{whang2021composing, kothari2021trumpets} require a
more substantial modification to the latent distribution, as the
pretrained model in these methods originally provides samples from the
prior distribution. For large-scale inverse problems, such as seismic
imaging, where solving a partial differential equation is required to
evaluate the forward operator, reduced computational costs of Bayesian
inference are particularly important. Our approach reduces the
computational costs associated with Bayesian inference in such problems
while also complementing the inversion with a learned prior. In order to
quantify the extent to which a diagonal correction to the latent
distribution mitigates errors resulting from data distribution shifts,
more research is required.

While the examples we presented in this paper were in regards to linear
inverse problem, variational inference can be also applied to nonlinear
problems
\citep{zhang2020seismic, zhang2021bayesian2, zhang2021introduction, li2021traversing, Curtis3d}.
However, in the context of amortized variational inference there are two
main considerations that need to be taken into account when dealing with
large-scale nonlinear inverse problems. First, the parametric surrogate
(conditional) distribution should be capable of approximating
multi-modal densities. Invertible neural networks
\citep{dinh2016density} are a suitable choice for this parameterization
as they are known to be universal approximators
\citep{NEURIPS2020_2290a738, ishikawa2022universal}, meaning that
invertible networks are capable of approximating a general class of
diffeomorphisms. In other words, as long as there is a smooth invertible
map between the latent space and the desired posterior distribution,
invertible neural networks can be used to approximate the posterior
distribution. The second consideration is regarding finding
low-dimensional summary statistics \citep{radev2020bayesflow} of
observed data to avoid training the conditional normalizing flow over
high dimensional data. In our linear seismic imaging example, we
``summarized'' seismic data (shot records) via the reverse-time migrated
image. The conditions for which this summarization does not negatively
bias the outcome of Bayesian inference (in the context of linear inverse
problems) is described by \citet{orozco2022SPIEadjoint}. Further work is
needed to design low-dimensional summary statistics for nonlinear
inverse problems in order to successfully apply our framework to
large-scale nonlinear inverse problems.

As far as seismic imaging is concerned, uncertainty can be attributed to
two main sources \citep{thore2002structural, osypov2013model, Ely2018}:
(1) data errors, including measurement noise; (2) modeling errors,
including linearization errors, which diminish with the accuracy of the
background velocity model. The scope of our research is focused on the
first source of uncertainty, and we will explore how variational
inference models can be used to capture errors in background models in
future work. In contrast to the problem highlighted in this paper,
capturing the uncertainty caused by errors in the background model would
require generating training data involving imaging experiments for a
variety of plausible background velocity models, which would be
computationally intensive. Recent developments in Fourier neural
operators \citep{li2021fourier, siahkoohi2022SEGvcw} may prove to be
useful in addressing this problem.

\section{Conclusions}\label{conclusions}

In high-dimensional inverse problems with computationally expensive
forward modeling operators, Bayesian inference is challenging due to the
cost of sampling the high-dimensional posterior distribution. The
computational costs associated with the forward operator often limit the
applicability of sampling the posterior distribution with traditional
Markov chain Monte Carlo and variational inference methods. Added to the
computational challenges are the difficulties associated with selecting
a prior distribution that encodes our prior knowledge while not
negatively biasing the outcome of Bayesian inference. Amortized
variational inference addresses these challenges by incurring an offline
initial training cost for a deep neural network that can approximate the
posterior distribution for previously unseen observed data, distributed
according to the training data distribution. When high-quality training
data is readily available, and as long as there are no shifts in the
data distribution, the pretrained network is capable of providing
samples from the posterior distribution for previously unknown data
virtually free of additional costs.

Unfortunately, in certain domains, such as geophysical inverse problems,
where the structure of the Earth's subsurface is unknown, it can be
challenging to obtain a training dataset, e.g., a collection of images
of the subsurface, which statistically captures the strong heterogeneity
exhibited by the Earth's subsurface. Furthermore, changes to the data
generation process could negatively influence the quality of Bayesian
inferences with amortized variational inferences due to generalization
errors associated with neural networks. To address these challenges
while exploiting the computational benefits of amortized variational
inference, we proposed a data-specific, physics-based, and
computationally cheap correction to the latent distribution of a
conditional normalizing flow, pretrained to via an amortized variational
inference objective. This correction involves solving a variational
inference problem in the latent space of the pretrained conditional
normalizing flow where we obtain a diagonal correction to the latent
distribution such that the predicted posterior distribution more closely
matches the desired posterior distribution.

Using a seismic imaging example, we demonstrate that the proposed latent
distribution correction, at a cost of five reverse-time migrations, can
be used to mitigate the effects of data distribution shifts, which
includes changes in the forward model as well as the prior distribution.
Our evaluation indicated improvements in seismic image quality,
comparable to least squares imaging, after the latent distribution
correction step, as well as estimate on the uncertainty of the image. We
presented the pointwise standard deviation as a measure of uncertainty
in the image, which indicated an increase in variability in complex
geological areas and poorly illuminated areas. This approach will enable
uncertainty quantification in large-scale inverse problems, which
otherwise would be computationally expensive to achieve.

\section{Related material}\label{related-material}

The latent distribution correction optimization problem
(equation~\ref{reverse_kl_covariance_diagonal}) involves computing
gradients of the composition of the forward operator and the pretrained
conditional normalizing flow with respect to the latent variable.
Computing this gradient requires actions of the forward operator and the
adjoint of its Jacobian. In our numerical experiments, these operations
involved solving wave equations. For maximal numerical performance, we
use \href{https://github.com/slimgroup/JUDI.jl}{JUDI}
\citep{witteJUDI2019} to construct wave-equation solvers, which utilizes
the just-in-time \href{https://www.devitoproject.org/}{Devito}
\citep{devito-compiler, devito-api} compiler for the wave-equation based
simulations. The invertible network architectures are implemented using
\href{https://github.com/slimgroup/InvertibleNetworks.jl}{InvertibleNetworks.jl}
\citep{invnet}, a memory-efficient framework for training invertible
nets in Julia programming language. For more details on our
implementation, please refer to our code on
\href{https://github.com/slimgroup/ReliableAVI.jl}{GitHub}.

\section{Acknowledgments}\label{acknowledgments}

This research was carried out with the support of Georgia Research
Alliance and partners of the ML4Seismic Center.

\section{Appendix A}\label{appendix-a}

\subsection{Derivation of latent distribution correction objective
function}\label{derivation-of-latent-distribution-correction-objective-function}

The optimization problem for correcting the latent distribution involves
minimizing the KL divergence between the Gaussian relaxation of the
latent distribution
$\mathrm{N} \big(\B{z} \mid \Bs{\mu}, \operatorname{diag}(\B{s})^2\big)$
and the shifted latent distribution
$p_{\phi} (\B{z} \mid \B{y}_{\text{obs}})$, which is conditioned on an
instance of out-of-distribution data
$\B{y}_{\text{obs}} \sim \widehat{p}_{\text{data}} (\B{y})$. This is an
instance of non-amortized variational inference (see
equations~\ref{reverse_kl_divergence} and~\ref{vi_reverse_kl}) defined
over the latent variable:
\begin{equation}
\begin{aligned}
& \argmin_{\Bs{\mu}, \B{s}}\, \KL\,\big(\mathrm{N} \big(\B{z} \mid \Bs{\mu},
\operatorname{diag}(\B{s})^2\big) \mid\mid
p_{\phi} (\B{z} \mid \B{y}_{\text{obs}}) \big) \\
& =
\argmin_{\Bs{\mu}, \B{s}}\, \mathbb{E}_{\B{z} \sim \mathrm{N} (\B{z} \mid \Bs{\mu},
\operatorname{diag}(\B{s})^2)}
\bigg [ -\log p_{\phi} (\B{z} \mid \B{y}_{\text{obs}}) + \log \mathrm{N} \big(\B{z} \mid \Bs{\mu},
\operatorname{diag}(\B{s})^2\big) \bigg ].
\end{aligned}
\label{reverse_kl_covariance_diagonal_derivation}
\end{equation}
 The above expression can be further simplified by rewriting the
expectation with respect to
$\mathrm{N} \big(\B{z} \mid \Bs{\mu}, \operatorname{diag}(\B{s})^2\big)$
as the expectation with respect to a standard Gaussian distribution,
followed by an elementwise scaling by $\B{s}$ and a shift by $\Bs{\mu}$
\citep[see reparameterization trick in][]{Kingma2014}, i.e.,
\begin{equation}
\argmin_{\Bs{\mu}, \B{s}}\, \mathbb{E}_{\B{z} \sim \mathrm{N} (\B{z} \mid \B{0}, \B{I})}
\bigg [ -\log p_{\phi} (\B{s} \odot\B{z}
+ \Bs{\mu} \mid \B{y}_{\text{obs}}) + \log \mathrm{N} \big(\B{s} \odot\B{z}
+ \Bs{\mu} \mid \Bs{\mu},
\operatorname{diag}(\B{s})^2\big) \bigg ].
\label{change-expectation}
\end{equation}
 The last term in the expectation in equation~\ref{change-expectation}
is the log-density of a Gaussian distribution, which is equal to:
\begin{equation}
\begin{aligned}
& \log \mathrm{N} \big(\B{s} \odot\B{z}
+ \Bs{\mu} \mid \Bs{\mu},
\operatorname{diag}(\B{s})^2\big) \\
& = -\frac{D}{2} \log (2 \pi)
-\frac{1}{2} \log
\Big | \det \operatorname{diag}(\B{s})^2 \Big | -\frac{1}{2}
\big(\B{s} \odot\B{z} + \Bs{\mu} -\Bs{\mu} \big)^{\top}
\operatorname{diag}(\B{s})^{-2}
\big(\B{s} \odot\B{z} + \Bs{\mu} -\Bs{\mu} \big) \\
& = - \log \Big | \det \operatorname{diag}(\B{s}) \Big |
-\frac{1}{2} \big\| \B{z} \big\|_2^2 +  \text{const} \\
& = - \log \Big | \det \operatorname{diag}(\B{s}) \Big | + \text{const}.
\end{aligned}
\label{gaussian_density}
\end{equation}
 In the above equation, $D$ is the dimension of $\B{z}$, and the
constants represent terms that are not a function of $\Bs{\mu}$ or
$\B{s}$. By inserting equation~\ref{gaussian_density} into
equation~\ref{change-expectation} we arrive at the following objective
function for the latent distribution correction:
\begin{equation}
\begin{split}
\argmin_{\Bs{\mu}, \B{s}}\, \mathbb{E}_{\B{z} \sim \mathrm{N} (\B{z} \mid \B{0}, \B{I} )}
\bigg [ -\log p_{\phi} (\B{s} \odot\B{z} + \Bs{\mu}  \mid \B{y}_{\text{obs}}) - \log \Big | \det \operatorname{diag}(\B{s}) \Big | \bigg ] \\
\end{split}
\label{reverse_kl_covariance_diagonal_final}
\end{equation}
 Finally, we use use Bayes' rule and inserting the shifted latent
density function from equation~\ref{physics_informed_density} to arrive
at the objective function for latent distribution correction
(equation~\ref{reverse_kl_covariance_diagonal}):
\begin{equation}
\begin{aligned}
\argmin_{\Bs{\mu}, \B{s}}
\mathbb{E}_{\B{z} \sim \mathrm{N} (\B{z} \mid \B{0}, \B{I})}
\bigg [&  \frac{1}{2 \sigma^2} \sum_{i=1}^{N}
\big  \| \B{y}_{\text{obs}, i}-\mathcal{F}_i \circ
f_ {\phi} \big(\B{s} \odot\B{z}
+ \Bs{\mu}; \B{y}_{\text{obs}} \big) \big\|_2^2 \\
&  + \frac{1}{2}
\big\| \B{s} \odot \B{z}
+ \Bs{\mu} \big\|_2^2 - \log
\Big | \det \operatorname{diag}(\B{s}) \Big | \bigg ].
\end{aligned}
\label{reverse_kl_covariance_diagonal-derived}
\end{equation}

\bibliography{paper}

\begin{thebibliography}{90}
\providecommand{\natexlab}[1]{#1}
\providecommand{\url}[1]{\texttt{#1}}
\expandafter\ifx\csname urlstyle\endcsname\relax
  \providecommand{\doi}[1]{doi: #1}\else
  \providecommand{\doi}{doi: \begingroup \urlstyle{rm}\Url}\fi

\bibitem[Aster et~al.(2018)Aster, Borchers, and Thurber]{aster2018parameter}
Richard~C Aster, Brian Borchers, and Clifford~H Thurber.
\newblock \emph{Parameter Estimation and Inverse Problems}.
\newblock Elsevier, 2018.
\newblock \doi{10.1016/C2015-0-02458-3}.

\bibitem[Tarantola(2005)]{tarantola2005inverse}
Albert Tarantola.
\newblock \emph{Inverse problem theory and methods for model parameter
  estimation}.
\newblock SIAM, 2005.
\newblock ISBN 978-0-89871-572-9.
\newblock \doi{10.1137/1.9780898717921}.

\bibitem[Robert and Casella(2004)]{robert2004monte}
C.~Robert and G.~Casella.
\newblock \emph{Monte {C}arlo statistical methods}.
\newblock Springer-Verlag, 2004.

\bibitem[Martin et~al.(2012)Martin, Wilcox, Burstedde, and
  Ghattas]{MartinMcMC2012}
James Martin, Lucas~C. Wilcox, Carsten Burstedde, and OMAR Ghattas.
\newblock A {S}tochastic {N}ewton {MCMC} {M}ethod for {L}arge-scale
  {S}tatistical {I}nverse {P}roblems with {A}pplication to {S}eismic
  {I}nversion.
\newblock \emph{SIAM Journal on Scientific Computing}, 34\penalty0
  (3):\penalty0 A1460--A1487, 2012.
\newblock URL \url{http://epubs.siam.org/doi/abs/10.1137/110845598}.

\bibitem[Fang et~al.(2018)Fang, Silva, Kuske, and Herrmann]{fang2018uqfip}
Zhilong Fang, Curt~Da Silva, Rachel Kuske, and Felix~J. Herrmann.
\newblock Uncertainty quantification for inverse problems with weak
  partial-differential-equation constraints.
\newblock \emph{GEOPHYSICS}, 83\penalty0 (6):\penalty0 R629--R647, 2018.
\newblock \doi{10.1190/geo2017-0824.1}.

\bibitem[Siahkoohi et~al.(2022{\natexlab{a}})Siahkoohi, Rizzuti, and
  Herrmann]{siahkoohiGEOdbif}
Ali Siahkoohi, Gabrio Rizzuti, and Felix~J. Herrmann.
\newblock {Deep Bayesian inference for seismic imaging with tasks}.
\newblock \emph{Geophysics}, 87\penalty0 (5), 6 2022{\natexlab{a}}.
\newblock \doi{10.1190/geo2021-0666.1}.
\newblock URL \url{https://arxiv.org/abs/2110.04825}.

\bibitem[Gelman et~al.(2013)Gelman, Carlin, Stern, Dunson, Vehtari, and
  Rubin]{gelman2013bayesian}
Andrew Gelman, John~B Carlin, Hal~S Stern, David~B Dunson, Aki Vehtari, and
  Donald~B Rubin.
\newblock \emph{Bayesian Data Analysis}.
\newblock CRC press, 2013.
\newblock \doi{10.1201/9780429258480}.
\newblock URL \url{http://www.stat.columbia.edu/~gelman/book/BDA3.pdf}.

\bibitem[Curtis and Lomax(2001)]{curtis2001prior}
Andrew Curtis and Anthony Lomax.
\newblock Prior information, sampling distributions, and the curse of
  dimensionality.
\newblock \emph{Geophysics}, 66\penalty0 (2):\penalty0 372--378, 2001.

\bibitem[Welling and Teh(2011)]{welling2011bayesian}
Max Welling and Yee~Whye Teh.
\newblock {Bayesian Learning via Stochastic Gradient Langevin Dynamics}.
\newblock In \emph{{Proceedings of the 28th International Conference on Machine
  Learning}}, {ICML’11}, pages 681--688, Madison, WI, USA, 2011. Omnipress.
\newblock ISBN 9781450306195.
\newblock \doi{10.5555/3104482.3104568}.
\newblock URL \url{https://dl.acm.org/doi/abs/10.5555/3104482.3104568}.

\bibitem[Herrmann et~al.(2019)Herrmann, Siahkoohi, and
  Rizzuti]{herrmann2019NIPSliwcuc}
Felix~J. Herrmann, Ali Siahkoohi, and Gabrio Rizzuti.
\newblock Learned imaging with constraints and uncertainty quantification.
\newblock In \emph{{Neural Information Processing Systems (NeurIPS) 2019 Deep
  Inverse Workshop}}, 12 2019.
\newblock URL \url{https://arxiv.org/pdf/1909.06473.pdf}.

\bibitem[Zhao and Sen(2019)]{zhao2019gradient}
Zeyu Zhao and Mrinal~K Sen.
\newblock A gradient based {MCMC} method for {FWI} and uncertainty analysis.
\newblock In \emph{89th Annual International Meeting, SEG}, pages 1465--1469.
  Expanded Abstracts, 2019.
\newblock \doi{10.1190/segam2019-3216560.1}.

\bibitem[Kotsi et~al.(2020)Kotsi, Malcolm, and Ely]{kotsi2020}
M~Kotsi, A~Malcolm, and G~Ely.
\newblock {Uncertainty quantification in time-lapse seismic imaging: a
  full-waveform approach}.
\newblock \emph{{Geophysical Journal International}}, 222\penalty0
  (2):\penalty0 1245--1263, 05 2020.
\newblock ISSN 0956-540X.
\newblock \doi{10.1093/gji/ggaa245}.

\bibitem[Siahkoohi et~al.(2020{\natexlab{a}})Siahkoohi, Rizzuti, and
  Herrmann]{siahkoohi2020EAGEdlb}
Ali Siahkoohi, Gabrio Rizzuti, and Felix~J. Herrmann.
\newblock A deep-learning based bayesian approach to seismic imaging and
  uncertainty quantification.
\newblock In \emph{82nd EAGE Conference and Exhibition}. Extended Abstracts,
  2020{\natexlab{a}}.
\newblock \doi{10.3997/2214-4609.202010770}.

\bibitem[Siahkoohi et~al.(2020{\natexlab{b}})Siahkoohi, Rizzuti, and
  Herrmann]{siahkoohi2020SEGuqi}
Ali Siahkoohi, Gabrio Rizzuti, and Felix~J. Herrmann.
\newblock {Uncertainty quantification in imaging and automatic horizon
  tracking---a Bayesian deep-prior based approach}.
\newblock In \emph{90th Annual International Meeting, SEG}, pages 1636--1640.
  Expanded Abstracts, 9 2020{\natexlab{b}}.
\newblock \doi{10.1190/segam2020-3417560.1}.

\bibitem[Jordan et~al.(1999)Jordan, Ghahramani, Jaakkola, and
  Saul]{jordan1999introduction}
Michael~I Jordan, Zoubin Ghahramani, Tommi~S Jaakkola, and Lawrence~K Saul.
\newblock {An Introduction to Variational Methods for Graphical Models}.
\newblock \emph{{Machine Learning}}, 37\penalty0 (2):\penalty0 183--233, 1999.
\newblock \doi{10.1023/A:1007665907178}.

\bibitem[Wainwright and Jordan(2008)]{wainwright2008graphical}
Martin~J Wainwright and Michael~Irwin Jordan.
\newblock \emph{Graphical models, exponential families, and variational
  inference}.
\newblock Now Publishers Inc, 2008.

\bibitem[Rezende and Mohamed(2015)]{rezende2015variational}
Danilo Rezende and Shakir Mohamed.
\newblock Variational inference with normalizing flows.
\newblock volume~37 of \emph{Proceedings of Machine Learning Research}, pages
  1530--1538. PMLR, 07--09 Jul 2015.
\newblock URL \url{http://proceedings.mlr.press/v37/rezende15.html}.

\bibitem[Liu and Wang(2016)]{liu2016stein}
Qiang Liu and Dilin Wang.
\newblock {Stein Variational Gradient Descent: A General Purpose Bayesian
  Inference Algorithm}.
\newblock In \emph{{Advances in Neural Information Processing Systems}},
  volume~29, pages 2378--2386. {Curran Associates, Inc.}, 2016.
\newblock URL
  \url{https://proceedings.neurips.cc/paper/2016/file/b3ba8f1bee1238a2f37603d90b58898d-Paper.pdf}.

\bibitem[Rizzuti et~al.(2020)Rizzuti, Siahkoohi, Witte, and
  Herrmann]{rizzuti2020SEGpub}
Gabrio Rizzuti, Ali Siahkoohi, Philipp~A. Witte, and Felix~J. Herrmann.
\newblock Parameterizing uncertainty by deep invertible networks, an
  application to reservoir characterization.
\newblock In \emph{90th Annual International Meeting, SEG}, pages 1541--1545,
  09 2020.
\newblock \doi{10.1190/segam2020-3428150.1}.

\bibitem[Zhang and Curtis(2020)]{zhang2020seismic}
Xin Zhang and Andrew Curtis.
\newblock Seismic tomography using variational inference methods.
\newblock \emph{Journal of Geophysical Research: Solid Earth}, 125\penalty0
  (4):\penalty0 e2019JB018589, 2020.

\bibitem[T{\"o}lle et~al.(2021)T{\"o}lle, Laves, and Schlaefer]{tolle2021mean}
Malte T{\"o}lle, Max-Heinrich Laves, and Alexander Schlaefer.
\newblock {A Mean-Field Variational Inference Approach to Deep Image Prior for
  Inverse Problems in Medical Imaging}.
\newblock In \emph{{Medical Imaging with Deep Learning}}, 2021.

\bibitem[Li et~al.(2021{\natexlab{a}})Li, Denli, MacDonald, Basler-Reeder,
  Baumstein, and Daves]{li2021multiparameter}
Dongzhuo Li, Huseyin Denli, Cody MacDonald, Kyle Basler-Reeder, Anatoly
  Baumstein, and Jacquelyn Daves.
\newblock Multiparameter geophysical reservoir characterization augmented by
  generative networks.
\newblock In \emph{First International Meeting for Applied Geoscience \&
  Energy}, pages 1364--1368. Society of Exploration Geophysicists,
  2021{\natexlab{a}}.

\bibitem[Li(2022)]{li2021traversing}
Dongzhuo Li.
\newblock Differentiable gaussianization layers for inverse problems
  regularized by deep generative models.
\newblock \emph{arXiv preprint arXiv:2112.03860}, 2022.

\bibitem[Blei et~al.(2017)Blei, Kucukelbir, and McAuliffe]{blei2017variational}
David~M Blei, Alp Kucukelbir, and Jon~D McAuliffe.
\newblock {Variational inference: A review for statisticians}.
\newblock \emph{{Journal of the American statistical Association}},
  112\penalty0 (518):\penalty0 859--877, 2017.

\bibitem[Zhang et~al.(2021)Zhang, Nawaz, Zhao, and
  Curtis]{zhang2021introduction}
Xin Zhang, Muhammad~Atif Nawaz, Xuebin Zhao, and Andrew Curtis.
\newblock An introduction to variational inference in geophysical inverse
  problems.
\newblock In \emph{Inversion of Geophysical Data}, pages 73--140. Elsevier,
  2021.
\newblock \doi{10.1016/bs.agph.2021.06.003}.
\newblock URL \url{https://doi.org/10.1016%2Fbs.agph.2021.06.003}.

\bibitem[Kim et~al.(2018)Kim, Wiseman, Miller, Sontag, and Rush]{pmlrv80kim18e}
Yoon Kim, Sam Wiseman, Andrew Miller, David Sontag, and Alexander Rush.
\newblock Semi-amortized variational autoencoders.
\newblock In Jennifer Dy and Andreas Krause, editors, \emph{Proceedings of the
  35th International Conference on Machine Learning}, volume~80 of
  \emph{Proceedings of Machine Learning Research}, pages 2678--2687. PMLR,
  10--15 Jul 2018.
\newblock URL \url{http://proceedings.mlr.press/v80/kim18e.html}.

\bibitem[Baptista et~al.(2020)Baptista, Zahm, and
  Marzouk]{baptista2020adaptive}
Ricardo Baptista, Olivier Zahm, and Youssef Marzouk.
\newblock An adaptive transport framework for joint and conditional density
  estimation.
\newblock \emph{arXiv preprint arXiv:2009.10303}, 2020.
\newblock URL \url{https://arxiv.org/abs/2009.10303}.

\bibitem[Kruse et~al.(2021)Kruse, Detommaso, Scheichl, and
  K{\"o}the]{kruse2021hint}
Jakob Kruse, Gianluca Detommaso, Robert Scheichl, and Ullrich K{\"o}the.
\newblock {HINT}: {H}ierarchical {I}nvertible {N}eural {T}ransport for
  {D}ensity {E}stimation and {B}ayesian {I}nference.
\newblock \emph{Proceedings of AAAI-2021}, 2021.
\newblock URL \url{https://arxiv.org/pdf/1905.10687.pdf}.

\bibitem[Kovachki et~al.(2021)Kovachki, Baptista, Hosseini, and
  Marzouk]{kovachki2021conditional}
Nikola Kovachki, Ricardo Baptista, Bamdad Hosseini, and Youssef Marzouk.
\newblock {Conditional Sampling With Monotone GANs}, 2021.

\bibitem[Radev et~al.(2022)Radev, Mertens, Voss, Ardizzone, and
  Köthe]{radev2020bayesflow}
Stefan~T. Radev, Ulf~K. Mertens, Andreas Voss, Lynton Ardizzone, and Ullrich
  Köthe.
\newblock {BayesFlow}: Learning complex stochastic models with invertible
  neural networks.
\newblock \emph{IEEE Transactions on Neural Networks and Learning Systems},
  33\penalty0 (4):\penalty0 1452--1466, 2022.
\newblock \doi{10.1109/TNNLS.2020.3042395}.

\bibitem[Siahkoohi et~al.(2021)Siahkoohi, Rizzuti, Louboutin, Witte, and
  Herrmann]{siahkoohi2020ABIpto}
Ali Siahkoohi, Gabrio Rizzuti, Mathias Louboutin, Philipp Witte, and Felix~J.
  Herrmann.
\newblock Preconditioned training of normalizing flows for variational
  inference in inverse problems.
\newblock In \emph{{3rd Symposium on Advances in Approximate Bayesian
  Inference}}, 1 2021.
\newblock URL \url{https://openreview.net/pdf?id=P9m1sMaNQ8T}.

\bibitem[Ren et~al.(2021)Ren, Witte, Siahkoohi, Louboutin, and
  Herrmann]{ren2021uq}
Yuxiao Ren, Philipp~A. Witte, Ali Siahkoohi, Mathias Louboutin, and Felix~J.
  Herrmann.
\newblock {Seismic Velocity Inversion and Uncertainty Quantification Using
  Conditional Normalizing Flows}.
\newblock In \emph{American Geophysical Union (AGU) Fall Meeting}, 12 2021.
\newblock URL
  \url{https://agu.confex.com/agu/fm21/meetingapp.cgi/Paper/815883}.

\bibitem[Siahkoohi and Herrmann(2021)]{siahkoohi2021Seglbe}
Ali Siahkoohi and Felix~J Herrmann.
\newblock {Learning by example: fast reliability-aware seismic imaging with
  normalizing flows}.
\newblock In \emph{First International Meeting for Applied Geoscience \&
  Energy}, pages 1580--1585. Expanded Abstracts, 2021.
\newblock \doi{10.1190/segam2021-3581836.1}.

\bibitem[Khorashadizadeh et~al.(2022)Khorashadizadeh, Kothari, Salsi, Harandi,
  de~Hoop, and Dokmani'c]{khorashadizadeh2022conditional}
AmirEhsan Khorashadizadeh, Konik Kothari, Leonardo Salsi, Ali~Aghababaei
  Harandi, Maarten de~Hoop, and Ivan Dokmani'c.
\newblock {Conditional Injective Flows for Bayesian Imaging}.
\newblock \emph{arXiv preprint arXiv:2204.07664}, 2022.

\bibitem[Orozco et~al.(2021)Orozco, Siahkoohi, Rizzuti, van Leeuwen, and
  Herrmann]{orozco2021photoacoustic}
Rafael Orozco, Ali Siahkoohi, Gabrio Rizzuti, Tristan van Leeuwen, and
  Felix~Johan Herrmann.
\newblock Photoacoustic imaging with conditional priors from normalizing flows.
\newblock In \emph{NeurIPS 2021 Workshop on Deep Learning and Inverse
  Problems}, 2021.
\newblock URL \url{https://openreview.net/forum?id=woi1OTvROO1}.

\bibitem[Siahkoohi et~al.(2022{\natexlab{b}})Siahkoohi, Orozco, Rizzuti, and
  Herrmann]{siahkoohi2022EAGEweb}
Ali Siahkoohi, Rafael Orozco, Gabrio Rizzuti, and Felix~J. Herrmann.
\newblock Wave-equation based inversion with amortized variational bayesian
  inference.
\newblock In \emph{EAGE Deep learning for seismic processing: Investigating the
  foundations workshop}, 6 2022{\natexlab{b}}.
\newblock URL \url{https://arxiv.org/abs/2203.15881}.

\bibitem[Taghvaei and Hosseini(2022)]{taghvaei2022optimal}
Amirhossein Taghvaei and Bamdad Hosseini.
\newblock An optimal transport formulation of bayes' law for nonlinear
  filtering algorithms.
\newblock \emph{arXiv preprint arXiv:2203.11869}, 2022.

\bibitem[Sun et~al.(2021)Sun, Innanen, and Huang]{sun2021physics}
Jian Sun, Kristopher~A Innanen, and Chao Huang.
\newblock Physics-guided deep learning for seismic inversion with hybrid
  training and uncertainty analysis.
\newblock \emph{Geophysics}, 86\penalty0 (3):\penalty0 R303--R317, 2021.

\bibitem[Jin et~al.(2022)Jin, Zhang, Chen, Huang, Liu, and
  Lin]{jin2022unsupervised}
Peng Jin, Xitong Zhang, Yinpeng Chen, Sharon~X Huang, Zicheng Liu, and Youzuo
  Lin.
\newblock Unsupervised learning of full-waveform inversion: Connecting {CNN}
  and partial differential equation in a loop.
\newblock In \emph{International Conference on Learning Representations}, 2022.
\newblock URL \url{https://openreview.net/forum?id=izvwgBic9q}.

\bibitem[Schmitt et~al.(2021)Schmitt, B{\"u}rkner, K{\"o}the, and
  Radev]{schmitt2021bayesflow}
Marvin Schmitt, Paul-Christian B{\"u}rkner, Ullrich K{\"o}the, and Stefan~T
  Radev.
\newblock Detecting model misspecification in amortized bayesian inference with
  neural networks.
\newblock \emph{arXiv preprint arXiv:2112.08866}, 2021.

\bibitem[Dinh et~al.(2016)Dinh, Sohl-Dickstein, and Bengio]{dinh2016density}
Laurent Dinh, Jascha Sohl-Dickstein, and Samy Bengio.
\newblock {Density estimation using Real NVP}.
\newblock In \emph{{International Conference on Learning Representations,
  {ICLR}}}, 2016.
\newblock URL \url{http://arxiv.org/abs/1605.08803}.

\bibitem[Asim et~al.(2020)Asim, Daniels, Leong, Ahmed, and
  Hand]{pmlr-v119-asim20a}
Muhammad Asim, Max Daniels, Oscar Leong, Ali Ahmed, and Paul Hand.
\newblock Invertible generative models for inverse problems: mitigating
  representation error and dataset bias.
\newblock In \emph{Proceedings of the 37th International Conference on Machine
  Learning}, volume 119 of \emph{Proceedings of Machine Learning Research},
  pages 399--409. PMLR, 07 2020.
\newblock URL \url{http://proceedings.mlr.press/v119/asim20a.html}.

\bibitem[Parno and Marzouk(2018)]{marzuk2018}
Matthew~D Parno and Youssef~M Marzouk.
\newblock Transport {M}ap {A}ccelerated {M}arkov {C}hain {M}onte {C}arlo.
\newblock \emph{SIAM/ASA Journal on Uncertainty Quantification}, 6\penalty0
  (2):\penalty0 645--682, 2018.
\newblock \doi{10.1137/17M1134640}.

\bibitem[Peherstorfer and Marzouk(2019)]{marzouk2018multifidelity}
Benjamin Peherstorfer and Youssef Marzouk.
\newblock A transport-based multifidelity preconditioner for {M}arkov chain
  {M}onte {C}arlo.
\newblock \emph{Advances in Computational Mathematics}, 45\penalty0
  (5-6):\penalty0 2321--2348, 2019.

\bibitem[Bora et~al.(2017)Bora, Jalal, Price, and Dimakis]{pmlr-v70-bora17a}
Ashish Bora, Ajil Jalal, Eric Price, and Alexandros~G. Dimakis.
\newblock Compressed sensing using generative models.
\newblock In Doina Precup and Yee~Whye Teh, editors, \emph{Proceedings of the
  34th International Conference on Machine Learning}, volume~70 of
  \emph{Proceedings of Machine Learning Research}, pages 537--546. PMLR, 06--11
  Aug 2017.
\newblock URL \url{https://proceedings.mlr.press/v70/bora17a.html}.

\bibitem[Andrle et~al.(2021)Andrle, Farchmin, Hagemann, Heidenreich, Soltwisch,
  and Steidl]{andrle2021invertible}
Anna Andrle, Nando Farchmin, Paul Hagemann, Sebastian Heidenreich, Victor
  Soltwisch, and Gabriele Steidl.
\newblock Invertible neural networks versus mcmc for posterior reconstruction
  in grazing incidence x-ray fluorescence.
\newblock In \emph{International Conference on Scale Space and Variational
  Methods in Computer Vision}, pages 528--539. Springer, 2021.

\bibitem[Zhao et~al.(2021)Zhao, Curtis, and Zhang]{zhao2021bayesian}
Xuebin Zhao, Andrew Curtis, and Xin Zhang.
\newblock {Bayesian seismic tomography using normalizing flows}.
\newblock \emph{Geophysical Journal International}, 228\penalty0 (1):\penalty0
  213--239, 07 2021.
\newblock ISSN 0956-540X.
\newblock \doi{10.1093/gji/ggab298}.
\newblock URL \url{https://doi.org/10.1093/gji/ggab298}.

\bibitem[Zhang and Curtis(2021)]{zhang2021bayesian2}
Xin Zhang and Andrew Curtis.
\newblock Bayesian geophysical inversion using invertible neural networks.
\newblock \emph{Journal of Geophysical Research: Solid Earth}, 126\penalty0
  (7):\penalty0 e2021JB022320, 2021.

\bibitem[Zhao et~al.(2022)Zhao, Curtis, and Zhang]{zhao2022interrogating}
Xuebin Zhao, Andrew Curtis, and Xin Zhang.
\newblock Interrogating subsurface structures using probabilistic tomography:
  an example assessing the volume of irish sea basins.
\newblock \emph{Journal of Geophysical Research: Solid Earth}, 127\penalty0
  (4):\penalty0 e2022JB024098, 2022.

\bibitem[Kothari et~al.(2021)Kothari, Khorashadizadeh, de~Hoop, and
  Dokmanic]{kothari2021trumpets}
Konik Kothari, AmirEhsan Khorashadizadeh, Maarten de~Hoop, and Ivan Dokmanic.
\newblock {Trumpets: Injective Flows for Inference and Inverse Problems}, 2021.

\bibitem[Adler and {\"O}ktem(2018)]{adler2018deep}
Jonas Adler and Ozan {\"O}ktem.
\newblock {Deep Bayesian Inversion}.
\newblock \emph{arXiv preprint arXiv:1811.05910}, 2018.
\newblock URL \url{https://arxiv.org/abs/1811.05910}.

\bibitem[Whang et~al.(2021)Whang, Lindgren, and Dimakis]{whang2021composing}
Jay Whang, Erik Lindgren, and Alex Dimakis.
\newblock Composing normalizing flows for inverse problems.
\newblock In \emph{International Conference on Machine Learning}, pages
  11158--11169. PMLR, 2021.

\bibitem[Malinverno and Briggs(2004)]{malinverno2004expanded}
Alberto Malinverno and Victoria~A Briggs.
\newblock Expanded uncertainty quantification in inverse problems: Hierarchical
  bayes and empirical bayes.
\newblock \emph{{GEOPHYSICS}}, 69\penalty0 (4):\penalty0 1005--1016, 2004.
\newblock \doi{10.1190/1.1778243}.

\bibitem[Malinverno and Parker(2006)]{malinverno2006two}
Alberto Malinverno and Robert~L Parker.
\newblock Two ways to quantify uncertainty in geophysical inverse problems.
\newblock \emph{{GEOPHYSICS}}, 71\penalty0 (3):\penalty0 W15--W27, 2006.
\newblock \doi{10.1190/1.2194516}.

\bibitem[Ray et~al.(2017)Ray, Kaplan, Washbourne, and Albertin]{chevron2017}
Anandaroop Ray, Sam Kaplan, John Washbourne, and Uwe Albertin.
\newblock {Low frequency full waveform seismic inversion within a tree based
  Bayesian framework}.
\newblock \emph{Geophysical Journal International}, 212\penalty0 (1):\penalty0
  522--542, 10 2017.
\newblock ISSN 0956-540X.
\newblock \doi{10.1093/gji/ggx428}.
\newblock URL \url{https://doi.org/10.1093/gji/ggx428}.

\bibitem[Stuart et~al.(2019)Stuart, Minkoff, and Pereira]{stuart2019two}
Georgia~K Stuart, Susan~E Minkoff, and Felipe Pereira.
\newblock A two-stage {Markov chain Monte Carlo} method for seismic inversion
  and uncertainty quantification.
\newblock \emph{{GEOPHYSICS}}, 84\penalty0 (6):\penalty0 R1003--R1020, 11 2019.
\newblock \doi{10.1190/geo2018-0893.1}.

\bibitem[Li et~al.(2016)Li, Chen, Carlson, and Carin]{li2016preconditioned}
Chunyuan Li, Changyou Chen, David Carlson, and Lawrence Carin.
\newblock {Preconditioned Stochastic Gradient Langevin Dynamics for Deep Neural
  Networks}.
\newblock In \emph{{Proceedings of the Thirtieth AAAI Conference on Artificial
  Intelligence}}, {AAAI’16}, pages 1788--1794. {AAAI Press}, 2016.
\newblock \doi{10.5555/3016100.3016149}.
\newblock URL \url{https://dl.acm.org/doi/abs/10.5555/3016100.3016149}.

\bibitem[Bishop(2006)]{bishop2006pattern}
Christopher~M Bishop.
\newblock \emph{{Pattern Recognition and Machine Learning}}.
\newblock {Springer-Verlag New York}, 2006.

\bibitem[Robbins and Monro(1951)]{robbins1951stochastic}
Herbert Robbins and Sutton Monro.
\newblock A stochastic approximation method.
\newblock \emph{The annals of mathematical statistics}, pages 400--407, 1951.

\bibitem[Nemirovski et~al.(2009)Nemirovski, Juditsky, Lan, and
  Shapiro]{nemirovski2009robust}
Arkadi Nemirovski, Anatoli Juditsky, Guanghui Lan, and Alexander Shapiro.
\newblock Robust stochastic approximation approach to stochastic programming.
\newblock \emph{SIAM Journal on optimization}, 19\penalty0 (4):\penalty0
  1574--1609, 2009.

\bibitem[Tieleman and Hinton(2012)]{rmsprop}
Tijmen Tieleman and Geoffrey Hinton.
\newblock Lecture 6.5-{RMS}prop: {D}ivide the gradient by a running average of
  its recent magnitude.
\newblock
  \url{https://www.cs.toronto.edu/~tijmen/csc321/slides/lecture_slides_lec6.pdf},
  2012.

\bibitem[Kingma and Ba(2014)]{kingma2014adam}
Diederik~P Kingma and Jimmy Ba.
\newblock {Adam: A method for stochastic optimization}.
\newblock \emph{arXiv preprint arXiv:1412.6980}, 2014.
\newblock URL \url{https://arxiv.org/pdf/1412.6980.pdf}.

\bibitem[Villani(2009)]{villani2009optimal}
C{\'e}dric Villani.
\newblock \emph{{Optimal transport: old and new}}.
\newblock {Springer-Verlag Berlin Heidelberg}, 2009.
\newblock \doi{10.1007/978-3-540-71050-9}.

\bibitem[Kingma and Welling(2014)]{Kingma2014}
Diederik~P. Kingma and Max Welling.
\newblock {Auto-Encoding Variational Bayes}.
\newblock In \emph{2nd International Conference on Learning Representations,
  {ICLR} 2014, Banff, AB, Canada, April 14-16, 2014, Conference Track
  Proceedings}, 2014.

\bibitem[Cranmer et~al.(2020)Cranmer, Brehmer, and Louppe]{cranmer2020frontier}
Kyle Cranmer, Johann Brehmer, and Gilles Louppe.
\newblock The frontier of simulation-based inference.
\newblock \emph{Proceedings of the National Academy of Sciences}, 117\penalty0
  (48):\penalty0 30055--30062, 2020.

\bibitem[Lavin et~al.(2021)Lavin, Zenil, Paige, Krakauer, Gottschlich, Mattson,
  Anandkumar, Choudry, Rocki, Baydin, et~al.]{lavin2021simulation}
Alexander Lavin, Hector Zenil, Brooks Paige, David Krakauer, Justin
  Gottschlich, Tim Mattson, Anima Anandkumar, Sanjay Choudry, Kamil Rocki,
  At{\i}l{\i}m~G{\"u}ne{\c{s}} Baydin, et~al.
\newblock {Simulation intelligence: Towards a new generation of scientific
  methods}.
\newblock \emph{arXiv preprint arXiv:2112.03235}, 2021.

\bibitem[Teshima et~al.(2020)Teshima, Ishikawa, Tojo, Oono, Ikeda, and
  Sugiyama]{NEURIPS2020_2290a738}
Takeshi Teshima, Isao Ishikawa, Koichi Tojo, Kenta Oono, Masahiro Ikeda, and
  Masashi Sugiyama.
\newblock Coupling-based invertible neural networks are universal
  diffeomorphism approximators.
\newblock In H.~Larochelle, M.~Ranzato, R.~Hadsell, M.F. Balcan, and H.~Lin,
  editors, \emph{Advances in Neural Information Processing Systems}, volume~33,
  pages 3362--3373. Curran Associates, Inc., 2020.
\newblock URL
  \url{https://proceedings.neurips.cc/paper/2020/file/2290a7385ed77cc5592dc2153229f082-Paper.pdf}.

\bibitem[Ishikawa et~al.(2022)Ishikawa, Teshima, Tojo, Oono, Ikeda, and
  Sugiyama]{ishikawa2022universal}
Isao Ishikawa, Takeshi Teshima, Koichi Tojo, Kenta Oono, Masahiro Ikeda, and
  Masashi Sugiyama.
\newblock Universal approximation property of invertible neural networks.
\newblock \emph{arXiv preprint arXiv:2204.07415}, 2022.
\newblock URL \url{https://arxiv.org/abs/2204.07415}.

\bibitem[Gubernatis et~al.(1977)Gubernatis, Domany, Krumhansl, and
  Huberman]{gubernatis77}
J.~E. Gubernatis, E.~Domany, J.~A. Krumhansl, and M.~Huberman.
\newblock {The Born approximation in the theory of the scattering of elastic
  waves by flaws}.
\newblock \emph{Journal of Applied Physics}, 48\penalty0 (7):\penalty0
  2812--2819, 1977.
\newblock \doi{10.1063/1.324142}.

\bibitem[Lambar{\'e} et~al.(1992)Lambar{\'e}, Virieux, Madariaga, and
  Jin]{lambare1992iterative}
Gilles Lambar{\'e}, Jean Virieux, Raul Madariaga, and Side Jin.
\newblock Iterative asymptotic inversion in the acoustic approximation.
\newblock \emph{Geophysics}, 57\penalty0 (9):\penalty0 1138--1154, 1992.

\bibitem[Schuster(1993)]{schuster1993least}
Gerard~T Schuster.
\newblock Least-squares cross-well migration.
\newblock In \emph{SEG Technical Program Expanded Abstracts 1993}, pages
  110--113. Society of Exploration Geophysicists, 1993.

\bibitem[Nemeth et~al.(1999)Nemeth, Wu, and Schuster]{nemeth1999least}
Tamas Nemeth, Chengjun Wu, and Gerard~T Schuster.
\newblock Least-squares migration of incomplete reflection data.
\newblock \emph{GEOPHYSICS}, 64\penalty0 (1):\penalty0 208--221, 1999.
\newblock \doi{10.1190/1.1444517}.

\bibitem[Veritas(2005)]{Veritas2005}
Veritas.
\newblock {Parihaka 3D Marine Seismic Survey - Acquisition and Processing
  Report}.
\newblock Technical Report New Zealand Petroleum Report 3460, New Zealand
  Petroleum \& Minerals, Wellington, 2005.

\bibitem[WesternGeco.(2012)]{WesternGeco2012}
WesternGeco.
\newblock {Parihaka 3D PSTM Final Processing Report}.
\newblock Technical Report New Zealand Petroleum Report 4582, New Zealand
  Petroleum \& Minerals, Wellington, 2012.

\bibitem[Adler et~al.(2022)Adler, Lunz, Verdier, Schönlieb, and
  Öktem]{Adler_2022}
Jonas Adler, Sebastian Lunz, Olivier Verdier, Carola-Bibiane Schönlieb, and
  Ozan Öktem.
\newblock Task adapted reconstruction for inverse problems.
\newblock \emph{Inverse Problems}, 38\penalty0 (7):\penalty0 075006, may 2022.
\newblock \doi{10.1088/1361-6420/ac28ec}.
\newblock URL \url{https://doi.org/10.1088/1361-6420/ac28ec}.

\bibitem[Liu et~al.(2014)Liu, Chakrabarti, Samanta, Ghosh, and
  Ghosh]{liu2014divergence}
Ruitao Liu, Arijit Chakrabarti, Tapas Samanta, Jayanta~K Ghosh, and Malay
  Ghosh.
\newblock On divergence measures leading to jeffreys and other reference
  priors.
\newblock \emph{Bayesian Analysis}, 9\penalty0 (2):\penalty0 331--370, 2014.

\bibitem[Yang and Soatto(2018)]{yang2018conditional}
Yanchao Yang and Stefano Soatto.
\newblock {Conditional Prior Networks for Optical Flow}.
\newblock In \emph{{Proceedings of the European Conference on Computer Vision
  (ECCV)}}, pages 271--287, 2018.

\bibitem[Zeno et~al.(2018)Zeno, Golan, Hoffer, and Soudry]{zeno2018task}
Chen Zeno, Itay Golan, Elad Hoffer, and Daniel Soudry.
\newblock {Task Agnostic Continual Learning Using Online Variational Bayes}.
\newblock \emph{arXiv preprint arXiv:1803.10123}, 2018.

\bibitem[Yosinski et~al.(2014)Yosinski, Clune, Bengio, and
  Lipson]{yosinski2014transferable}
Jason Yosinski, Jeff Clune, Yoshua Bengio, and Hod Lipson.
\newblock How transferable are features in deep neural networks?
\newblock In \emph{Proceedings of the 27th International Conference on Neural
  Information Processing Systems}, NIPS'14, pages 3320--3328, 2014.
\newblock URL \url{http://dl.acm.org/citation.cfm?id=2969033.2969197}.

\bibitem[Zhang et~al.(2022)Zhang, Lomas, Zhou, Zheng, and Curtis]{Curtis3d}
Xin Zhang, Angus Lomas, Muhong Zhou, York Zheng, and Andrew Curtis.
\newblock 3d bayesian variational full waveform inversion, 2022.
\newblock URL \url{https://arxiv.org/abs/2210.03613}.

\bibitem[Orozco et~al.(2023)Orozco, Siahkoohi, Rizzuti, van Leeuwen, and
  Herrmann]{orozco2022SPIEadjoint}
Rafael Orozco, Ali Siahkoohi, Gabrio Rizzuti, Tristan van Leeuwen, and Felix~J.
  Herrmann.
\newblock Adjoint operators enable fast and amortized machine learning based
  bayesian uncertainty quantification.
\newblock In \emph{SPIE Medical Imaging Conference}, 02 2023.
\newblock URL
  \url{https://slim.gatech.edu/Publications/Public/Conferences/SPIE/2023/orozco2022SPIEadjoint/SPIE_2022_adjoint.html}.
\newblock (SPIE, San Diego).

\bibitem[Thore et~al.(2002)Thore, Shtuka, Lecour, Ait-Ettajer, and
  Cognot]{thore2002structural}
Pierre Thore, Arben Shtuka, Magali Lecour, Taoufik Ait-Ettajer, and Richard
  Cognot.
\newblock {Structural uncertainties: Determination, management, and
  applications }.
\newblock \emph{Geophysics}, 67\penalty0 (3):\penalty0 840--852, 2002.

\bibitem[Osypov et~al.(2013)Osypov, Yang, Fournier, Ivanova, Bachrach, Yarman,
  You, Nichols, and Woodward]{osypov2013model}
Konstantin Osypov, Yi~Yang, Aim{\'e} Fournier, Natalia Ivanova, Ran Bachrach,
  Can~Evren Yarman, Yu~You, Dave Nichols, and Marta Woodward.
\newblock Model-uncertainty quantification in seismic tomography: method and
  applications.
\newblock \emph{Geophysical Prospecting}, 61\penalty0 (6-Challenges of Seismic
  Imaging and Inversion Devoted to Goldin):\penalty0 1114--1134, 2013.

\bibitem[Ely et~al.(2018)Ely, Malcolm, and Poliannikov]{Ely2018}
Gregory Ely, Alison Malcolm, and Oleg~V. Poliannikov.
\newblock Assessing uncertainties in velocity models and images with a fast
  nonlinear uncertainty quantification method.
\newblock \emph{GEOPHYSICS}, 83\penalty0 (2):\penalty0 R63--R75, 2018.
\newblock \doi{10.1190/geo2017-0321.1}.
\newblock URL \url{https://doi.org/10.1190/geo2017-0321.1}.

\bibitem[Li et~al.(2021{\natexlab{b}})Li, Kovachki, Azizzadenesheli, Liu,
  Bhattacharya, Stuart, and Anandkumar]{li2021fourier}
Zongyi Li, Nikola~Borislavov Kovachki, Kamyar Azizzadenesheli, Burigede Liu,
  Kaushik Bhattacharya, Andrew~M. Stuart, and Anima Anandkumar.
\newblock {Fourier Neural Operator for Parametric Partial Differential
  Equations}.
\newblock In \emph{9th International Conference on Learning Representations}.
  OpenReview.net, 2021{\natexlab{b}}.
\newblock URL \url{https://openreview.net/forum?id=c8P9NQVtmnO}.

\bibitem[Siahkoohi et~al.(2022{\natexlab{c}})Siahkoohi, Louboutin, and
  Herrmann]{siahkoohi2022SEGvcw}
Ali Siahkoohi, Mathias Louboutin, and Felix~J. Herrmann.
\newblock {Velocity continuation with Fourier neural operators for accelerated
  uncertainty quantification}.
\newblock In \emph{2nd International Meeting for Applied Geoscience \& Energy},
  2022{\natexlab{c}}.

\bibitem[Witte et~al.(2019)Witte, Louboutin, Kukreja, Luporini, Lange, Gorman,
  and Herrmann]{witteJUDI2019}
Philipp~A. Witte, Mathias Louboutin, Navjot Kukreja, Fabio Luporini, Michael
  Lange, Gerard~J. Gorman, and Felix~J. Herrmann.
\newblock A large-scale framework for symbolic implementations of seismic
  inversion algorithms in julia.
\newblock \emph{GEOPHYSICS}, 84\penalty0 (3):\penalty0 F57--F71, 2019.
\newblock \doi{10.1190/geo2018-0174.1}.
\newblock URL \url{https://doi.org/10.1190/geo2018-0174.1}.

\bibitem[{Luporini} et~al.(2018){Luporini}, {Lange}, {Louboutin}, {Kukreja},
  {H{\"u}ckelheim}, {Yount}, {Witte}, {Kelly}, {Herrmann}, and
  {Gorman}]{devito-compiler}
F.~{Luporini}, M.~{Lange}, M.~{Louboutin}, N.~{Kukreja}, J.~{H{\"u}ckelheim},
  C.~{Yount}, P.~{Witte}, P.~H.~J. {Kelly}, F.~J. {Herrmann}, and G.~J.
  {Gorman}.
\newblock Architecture and performance of devito, a system for automated
  stencil computation.
\newblock \emph{CoRR}, abs/1807.03032, jul 2018.
\newblock URL \url{http://arxiv.org/abs/1807.03032}.

\bibitem[Louboutin et~al.(2019)Louboutin, Lange, Luporini, Kukreja, Witte,
  Herrmann, Velesko, and Gorman]{devito-api}
M.~Louboutin, M.~Lange, F.~Luporini, N.~Kukreja, P.~A. Witte, F.~J. Herrmann,
  P.~Velesko, and G.~J. Gorman.
\newblock Devito (v3.1.0): an embedded domain-specific language for finite
  differences and geophysical exploration.
\newblock \emph{Geoscientific Model Development}, 12\penalty0 (3):\penalty0
  1165--1187, 2019.
\newblock \doi{10.5194/gmd-12-1165-2019}.
\newblock URL \url{https://www.geosci-model-dev.net/12/1165/2019/}.

\bibitem[Witte et~al.(2021)Witte, Rizzuti, Louboutin, Siahkoohi, Herrmann, and
  Peters]{invnet}
Philipp Witte, Gabrio Rizzuti, Mathias Louboutin, Ali Siahkoohi, Felix
  Herrmann, and Bas Peters.
\newblock {InvertibleNetworks.jl: A Julia framework for invertible neural
  networks}, March 2021.
\newblock URL \url{https://doi.org/10.5281/zenodo.4610118}.

\end{thebibliography}

\end{document}